\documentclass{article}
\usepackage{lipsum}
\usepackage{graphicx,wrapfig} 
\usepackage{textcomp}
\usepackage{amssymb}
\usepackage{authblk}
\usepackage{gensymb}

\usepackage{amsmath}
\usepackage{algorithm}
\usepackage{algpseudocode}
\usepackage{adjustbox}
\usepackage{subcaption}
\usepackage{tikz}
\usepackage{nicematrix}
\usepackage[a4paper, margin=1.8cm]{geometry}
\usepackage{color,soul}
\usepackage{booktabs}
\usepackage[colorlinks=true, allcolors=blue]{hyperref}
\usepackage{cleveref,enumitem}

\title{Graph-Convolutional-Beta-VAE for Synthetic Abdominal Aorta Aneurysm Generation}

\author[1]{Francesco Fabbri}
\author[1,4]{Martino Andrea Scarpolini}
\author[3]{Angelo Iollo}
\author[1]{Francesco Viola}
\author[1,2]{Francesco Tudisco}

\affil[1]{Gran Sasso Science Institute, Italy}
\affil[2]{University of Edinburgh, UK}
\affil[3]{University of Bordeaux, France}
\affil[4]{INFN Tor Vergata, Rome, Italy.}
\date{}

\begin{document}

\maketitle

\begin{abstract}
Synthetic data generation plays a crucial role in medical research by mitigating privacy concerns and enabling large-scale patient data analysis. This study presents a $\beta$-Variational Autoencoder Graph Convolutional Neural Network framework for generating synthetic Abdominal Aorta Aneurysms (AAA). Using a small real-world dataset, our approach extracts key anatomical features and captures complex statistical relationships within a compact disentangled latent space. To address data limitations, low-impact data augmentation based on Procrustes analysis was employed, preserving anatomical integrity. The generation strategies both deterministic and stochastic manage to enhance data diversity while ensuring realism. Compared to PCA-based approaches, our model performs more robustly on unseen data by capturing complex, nonlinear anatomical variations. This enables more comprehensive clinical and statistical analyses than the original dataset alone. The resulting synthetic AAA dataset preserves patient privacy while providing a scalable foundation for medical research, device testing, and computational modeling.
\end{abstract}

\section{Introduction}
In clinical practice, synthetic data play an increasingly important role in tailoring medical treatments and therapeutic strategies to individual patients. They enable deeper understanding of disease progression and diagnostic patterns, while maintaining strict patient confidentiality and protecting sensitive health information \cite{PEZOULAS20242892,28f13de42e764597ac43c4db6b8e611e,Tucker2020,Lucini2021}.
Simulated or “virtual” patients are used to evaluate medical procedures and treatment strategies before they are applied in real-world clinical settings. By leveraging computational models capable of predicting clinical outcomes, researchers and clinicians can reduce uncertainty and support more informed, data-driven decisions. These tools may also enable the development of personalized treatment plans, such as custom stent-graft designs tailored to individual patient anatomies, helping to lower the risk of post-procedural complications \cite{WANG,WHITAKER,Czermak2001,Becker2022}. Despite their potential, the validation of new interventions typically requires large and diverse patient datasets, which are often challenging to obtain due to privacy restrictions and logistical constraints. One promising solution is the generation of synthetic patient cohorts by capturing and modeling key anatomical characteristics from a limited set of real cases. These virtual populations can then be used to simulate a broader spectrum of anatomical and physiological variations, facilitating comprehensive statistical assessments and in silico clinical trials. However, generating accurate synthetic datasets from limited patient populations presents notable challenges. 

Traditional linear methods, such as Principal Component Analysis (PCA), are advantageous for their simplicity and interpretability with small datasets; however, they often fail to adequately generalize beyond the limited variations captured in the original samples. On the other hand, neural network-based approaches, such as Variational Autoencoders (VAEs), are promising candidates due to their inherent nonlinear modeling capabilities \cite{Bourlard1988,8953837,chen2019isolatingsourcesdisentanglementvariational,dai2019hiddentalentsvariationalautoencoder,BAUR2021101952}. These methods effectively capture complex anatomical variations \cite{Jiang_2019_CVPR, Ge_2019_CVPR,  Zhao_2019_CVPR,9320455} and potentially generalize beyond the original range of anatomical variations while still being particularly effective in representing complex features within highly compact latent spaces. As a natural evolution of the VAE framework, deep probabilistic neural networks leveraging $\beta$-VAE architectures have successfully been applied in fluid dynamics, specifically for the extraction of interpretable non-linear modes from turbulent flows \cite{EIVAZI2022117038}, demonstrating superior performance compared to traditional linear methods like Proper Orthogonal Decomposition (POD) due to its ability to provide disentangled latent representations.  Graph Convolutional Networks have recently been applied in diverse areas of medical research \cite{Ranjan_2018_ECCV}. For example, Graph-Autoencoders have been employed for automatic sex determination from hip bones, significantly enhancing interpretability by explicitly encoding categorical class differences within a compact latent space \cite{Zou_2023}. Similarly, unsupervised deep learning ensemble-based approaches have been utilized for phenotyping cardiac morphology from large datasets such as the UK Biobank, improving gene discoverability related to complex heart shape variations \cite{bonazzola2023unsupervisedensemblebasedphenotypinghelps}.

The objective of this work is to generate realistic, synthetic abdominal aorta aneurysm (AAA) datasets that accurately capture the nonlinear anatomical variability encountered in real clinical scenarios. This is accomplished through a Graph Convolutional Network $\beta$-Variational Autoencoder (GCN-$\beta$-VAE) architecture, which leverages graph-based representations to effectively model complex anatomical shapes. By employing the $\beta$-VAE framework, the latent disentanglement is enhanced, allowing targeted exploration and variation of specific morphological attributes within a compact latent space while preserving essential anatomical features.

The paper first presents a description of the dataset, followed by an in-depth explanation of the proposed GCN-$\beta$-VAE architecture, highlighting key components such as the reconstruction loss functions, spectral graph convolution, and spatial pooling sampling. A Procrustes-based data augmentation method is then introduced to enrich and diversify the training set while maintaining anatomical fidelity. The proposed methodology is evaluated through systematic benchmarking against conventional PCA-based techniques. The analysis includes the assessment of the augmentation strategy’s impact, exploration of latent space properties—such as generalization capability, disentanglement, and hierarchical mode contributions—and a demonstration of the generative capabilities of the best-performing model, highlighting its potential for clinical research and in silico experimentation.

\section{Methods}\label{sec:Methods}

In \Cref{sec:dataset}, a description of the dataset used in this study is presented, consisting of patient-specific 3D aneurysm meshes derived from CT scans through segmentation and surface reconstruction. Next, in \Cref{sec:vae}, the GCN-$\beta$-VAE Architecture (\Cref{fig:vae}) is described, outlining how 3D aneurysm meshes are encoded into a latent multivariate Gaussian space and then reconstructed. In particular, the theoretical background of the VAE is addressed, along with the role of the $\beta$-hyperparameter in promoting latent disentanglement. Then, it is presented in \Cref{sec:rec-loss}, the reconstruction loss used to recover high-fidelity meshes, including vertex coordinate, Chamfer, normal, and edge losses (each addressing a particular geometric constraint). Subsequently, \Cref{sec:graph-convolution} presents the Spectral Graph Convolution approach, which leverages the eigen-decomposition of the graph Laplacian and applies Chebyshev polynomial-based filters directly on mesh vertices, while \Cref{sec:spatial-pooling} describes the Spatial Pooling layer, highlighting how pair-contraction-based down-sampling and corresponding up-sampling matrices are used to iteratively simplify the mesh while enabling multi-resolution reconstruction. Finally, in \Cref{sec:procau}, a Procrustes-Based RandAugment protocol (ProcAug) is proposed, leveraging transformations computed via Procrustes alignment, specifically scaling and rotation, to design an online augmentation procedure.

\begin{figure}[t]
\centering
\begin{subfigure}{0.85\textwidth}
    \centering
    {\includegraphics[trim={11.75cm 1cm 11.75cm 1cm}, clip,width=0.12\textwidth]{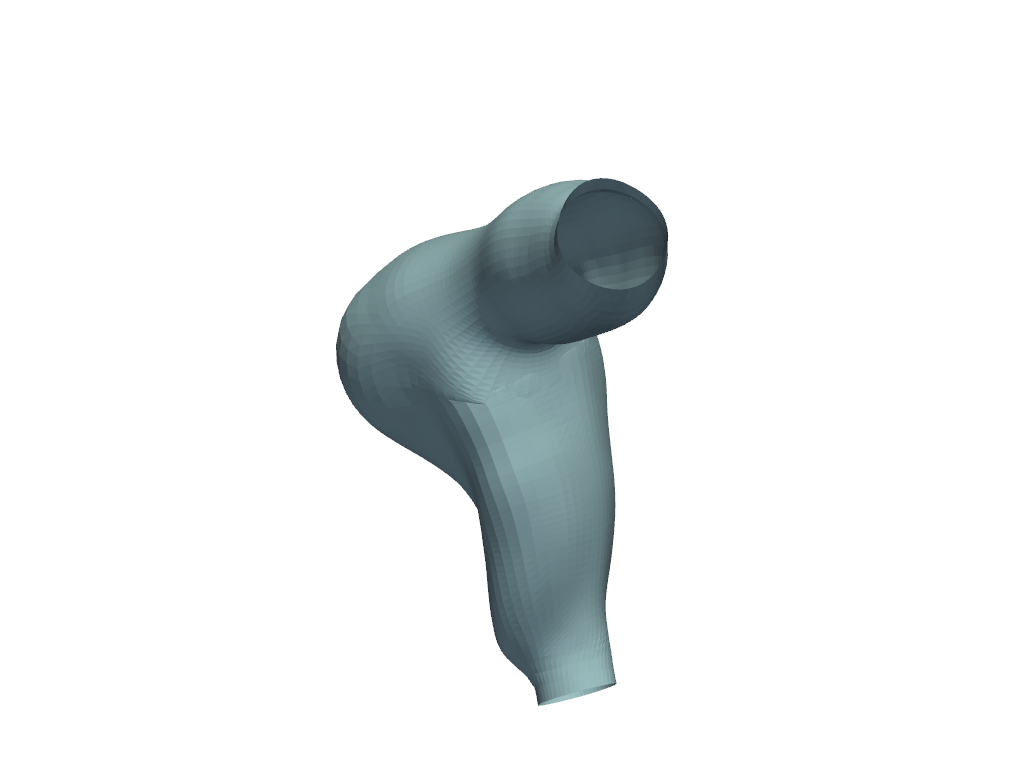}} \quad
    \subfloat[][]
    {\includegraphics[trim={11.75cm 1cm 11.75cm 1cm}, clip,width=0.12\textwidth]{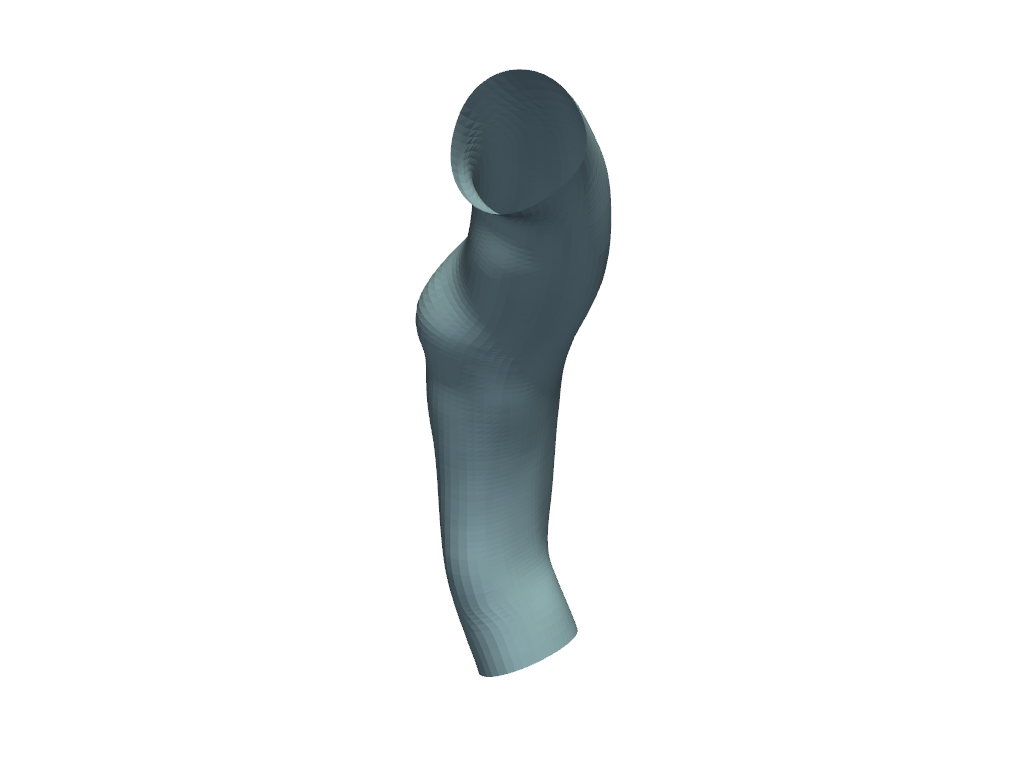}} \quad
    \subfloat[][]
    {\includegraphics[trim={11.75cm 1cm 11.75cm 1cm}, clip,width=0.12\textwidth]{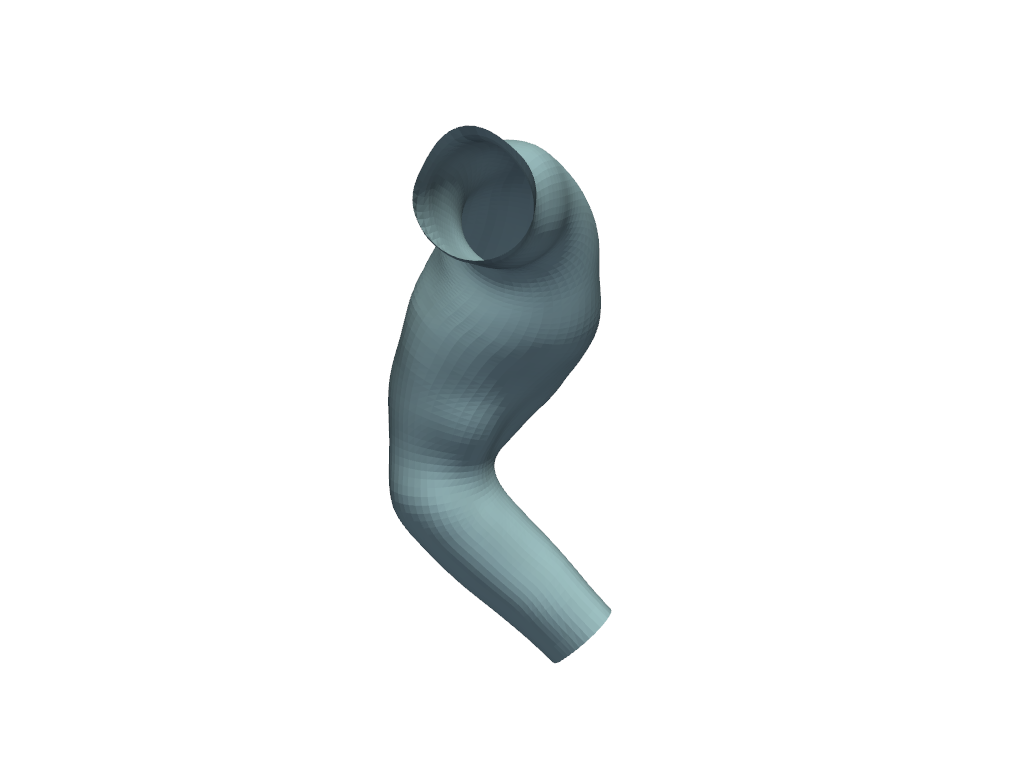}} \quad
    \subfloat[][]
    {\includegraphics[trim={11.75cm 1cm 11.75cm 1cm}, clip,width=0.12\textwidth]{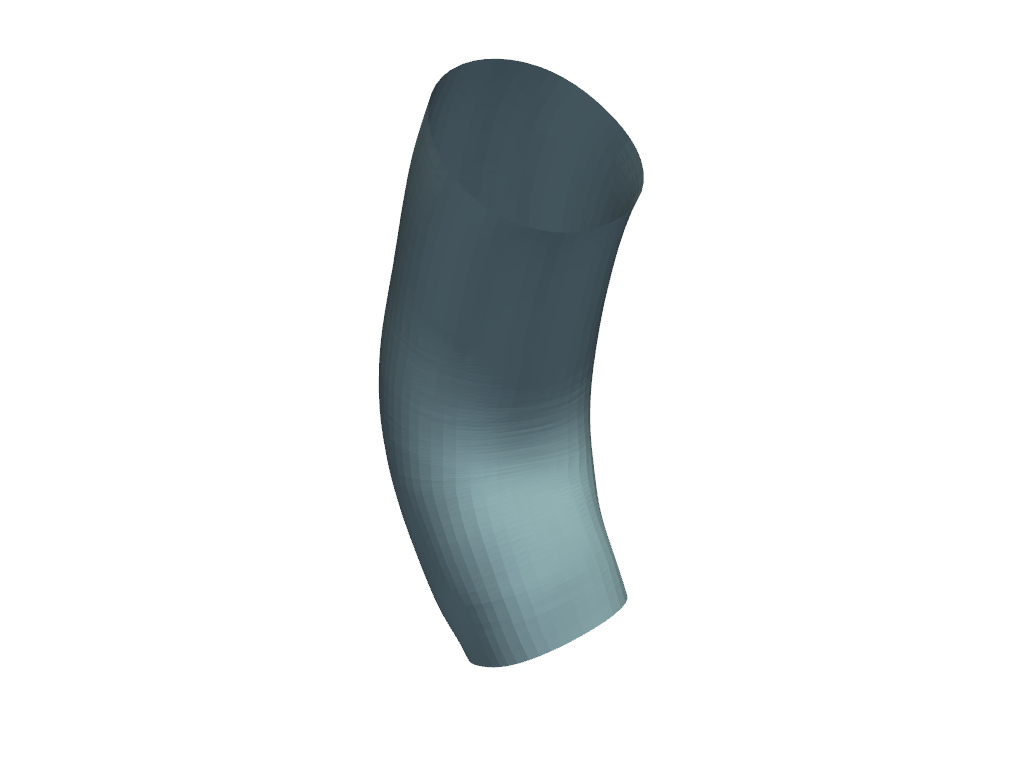}} \quad
    \subfloat[][]
    {\includegraphics[trim={11.75cm 1cm 11.75cm 1cm}, clip,width=0.12\textwidth]{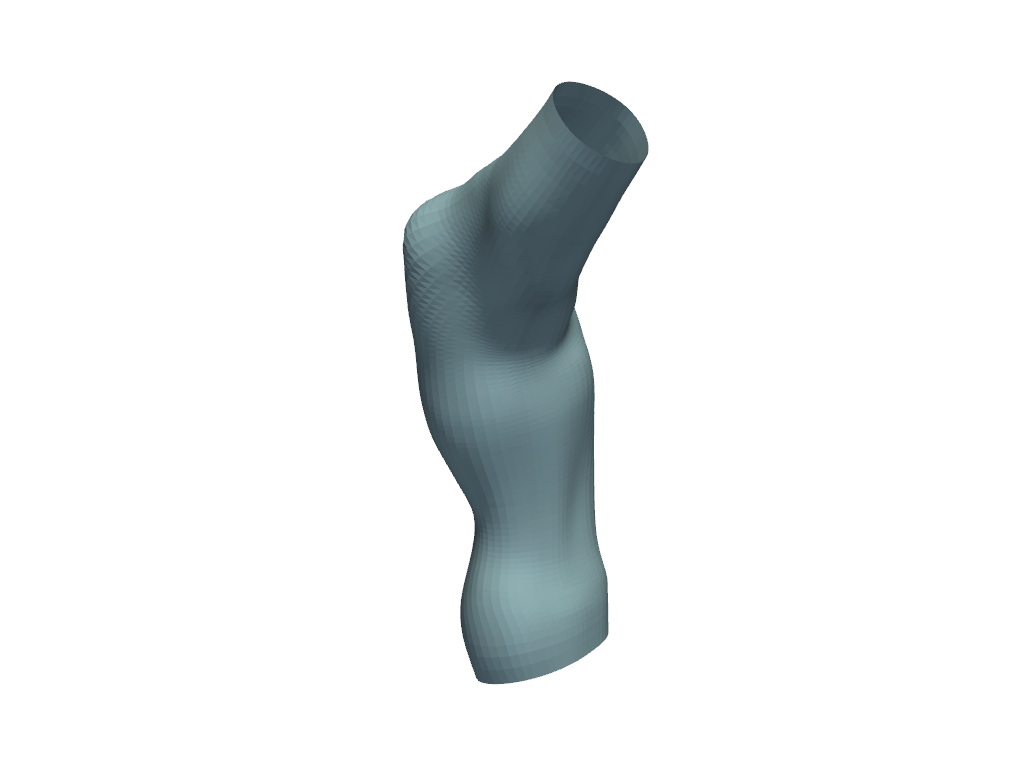}} \\
    \subfloat[][]
    {\includegraphics[trim={11.75cm 1cm 11.75cm 1cm}, clip,width=0.12\textwidth]{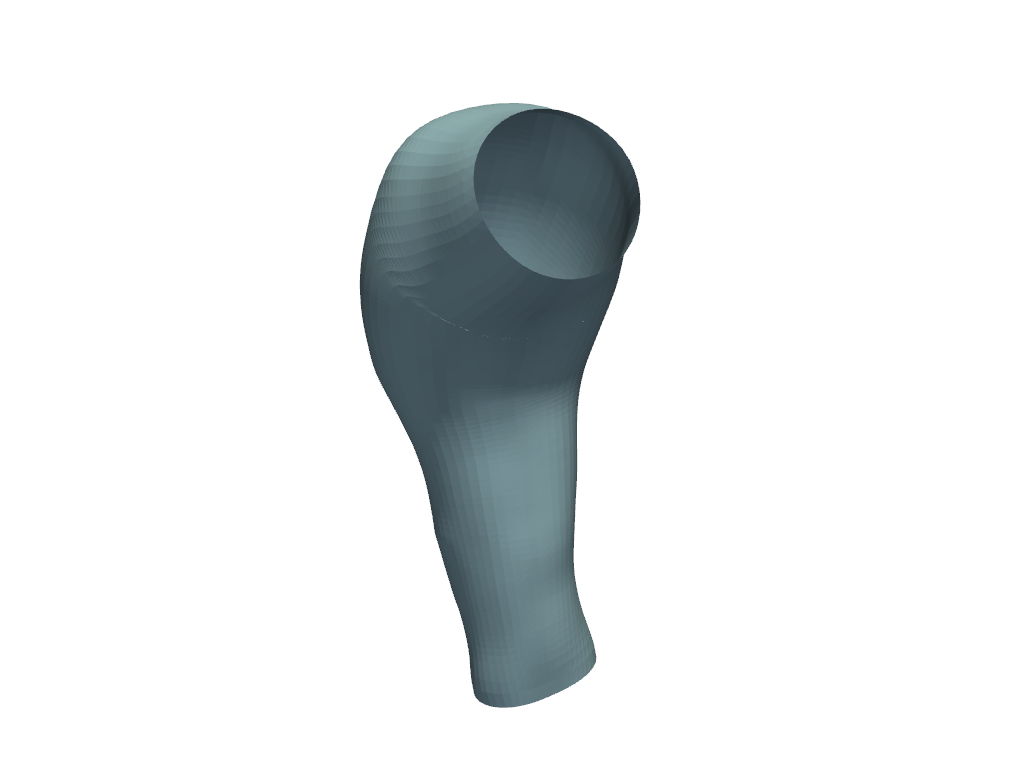}} \quad
    \subfloat[][]
    {\includegraphics[trim={11.75cm 1cm 11.75cm 1cm}, clip,width=0.12\textwidth]{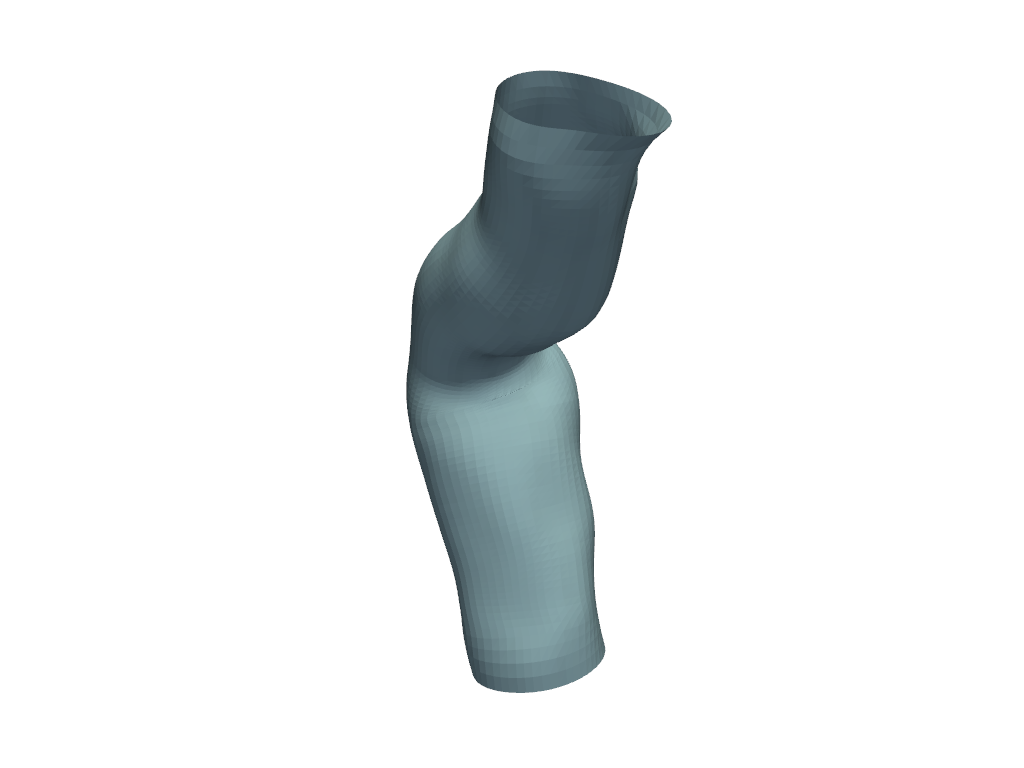}} \quad
    \subfloat[][]
    {\includegraphics[trim={11.75cm 1cm 11.75cm 1cm}, clip,width=0.12\textwidth]{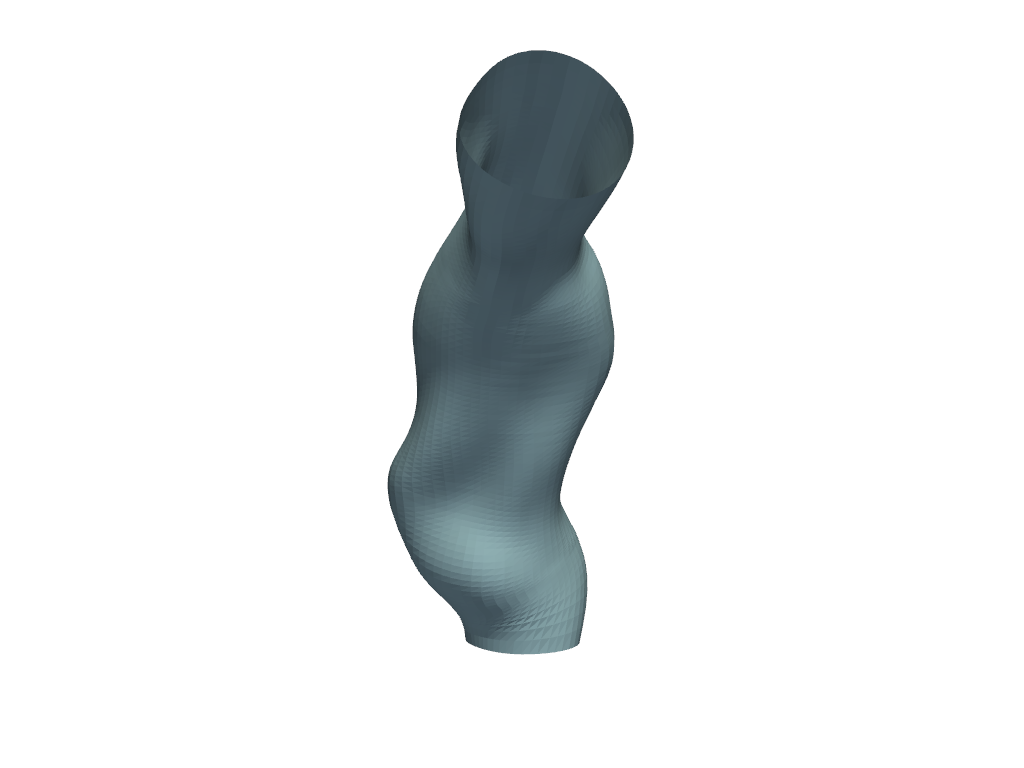}} \quad
    \subfloat[][]
    {\includegraphics[trim={11.75cm 1cm 11.75cm 1cm}, clip,width=0.12\textwidth]{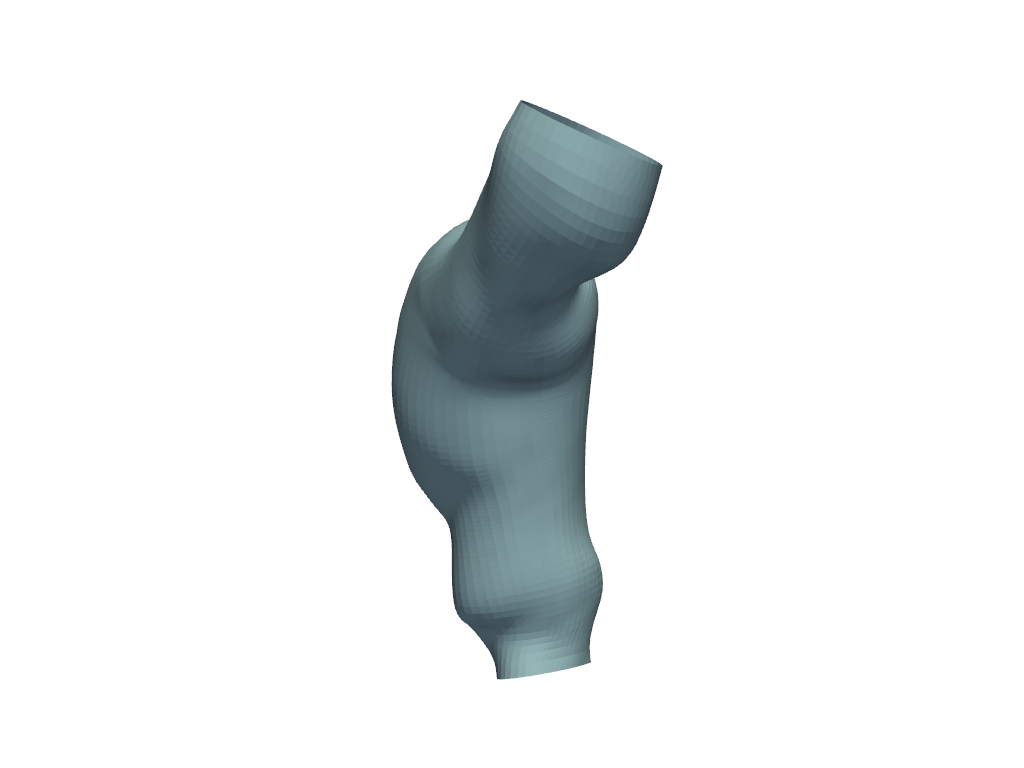}} \quad
    \subfloat[][]
    {\includegraphics[trim={11.75cm 1cm 11.75cm 1cm}, clip,width=0.12\textwidth]{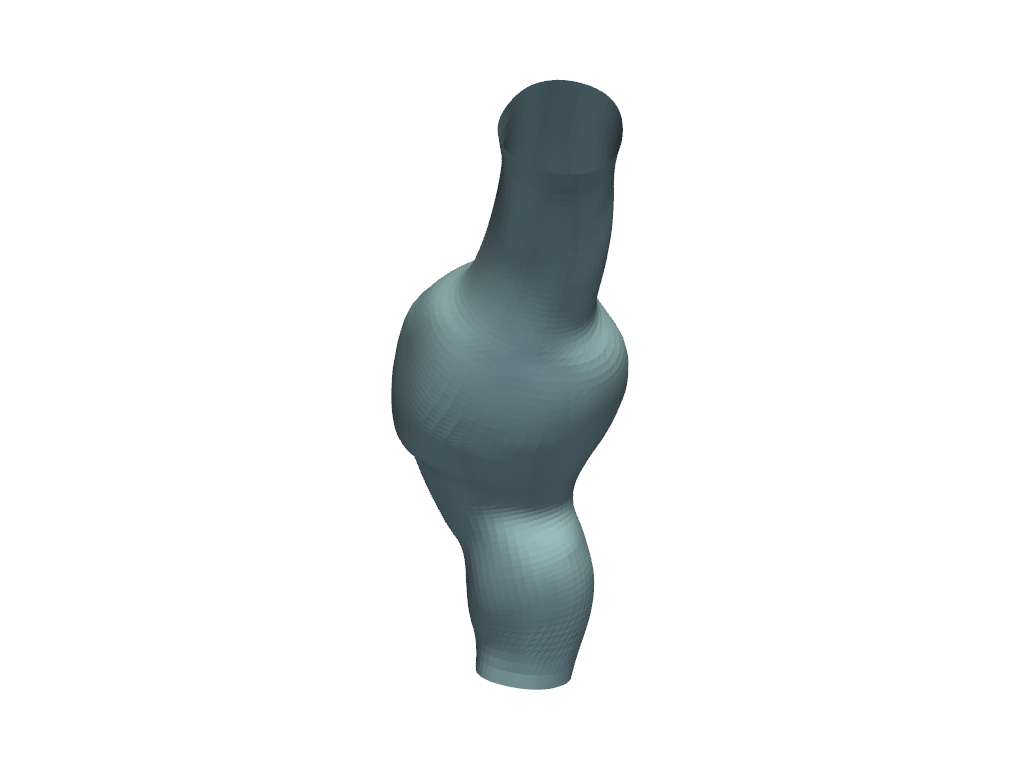}} \\
    {\includegraphics[trim={11.75cm 1cm 11.75cm 1cm}, clip,width=0.12\textwidth]{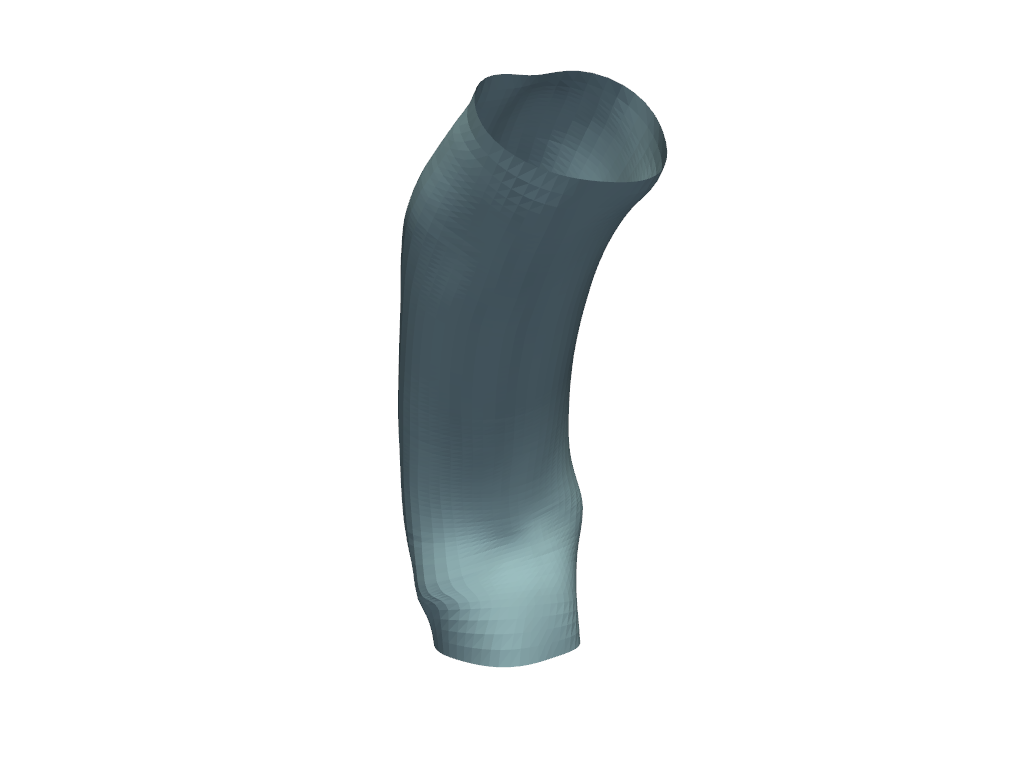}} \quad
    \subfloat[][]
    {\includegraphics[trim={11.75cm 1cm 11.75cm 1cm}, clip,width=0.12\textwidth]{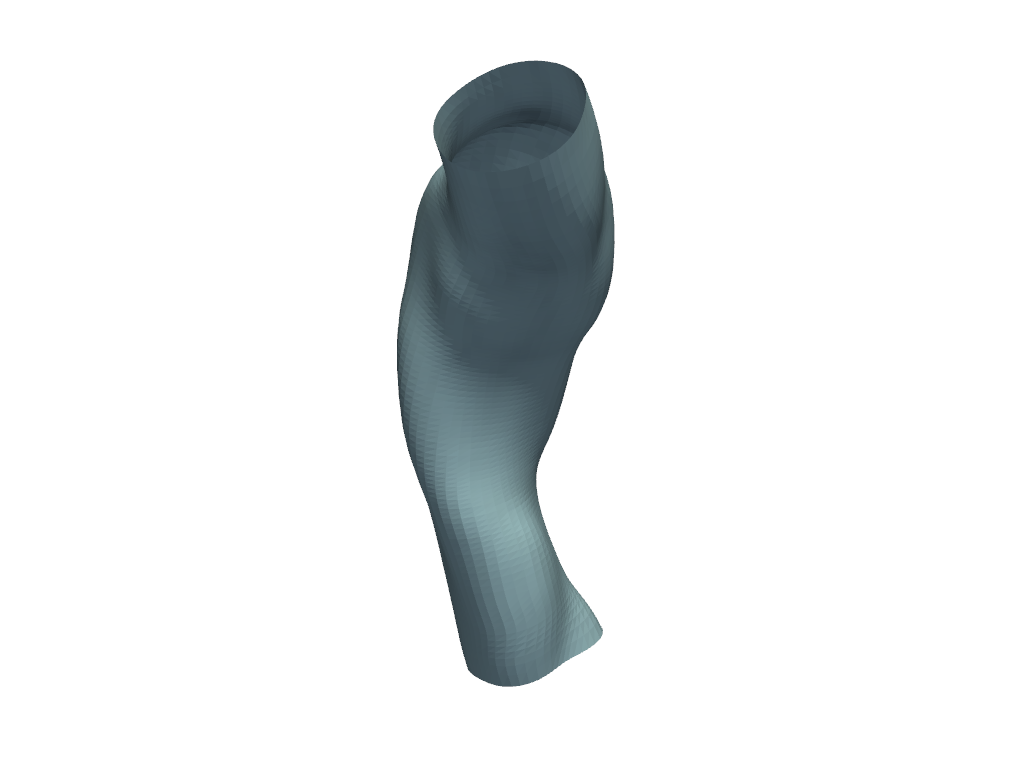}} \quad
    \subfloat[][]
    {\includegraphics[trim={11.75cm 1cm 11.75cm 1cm}, clip,width=0.12\textwidth]{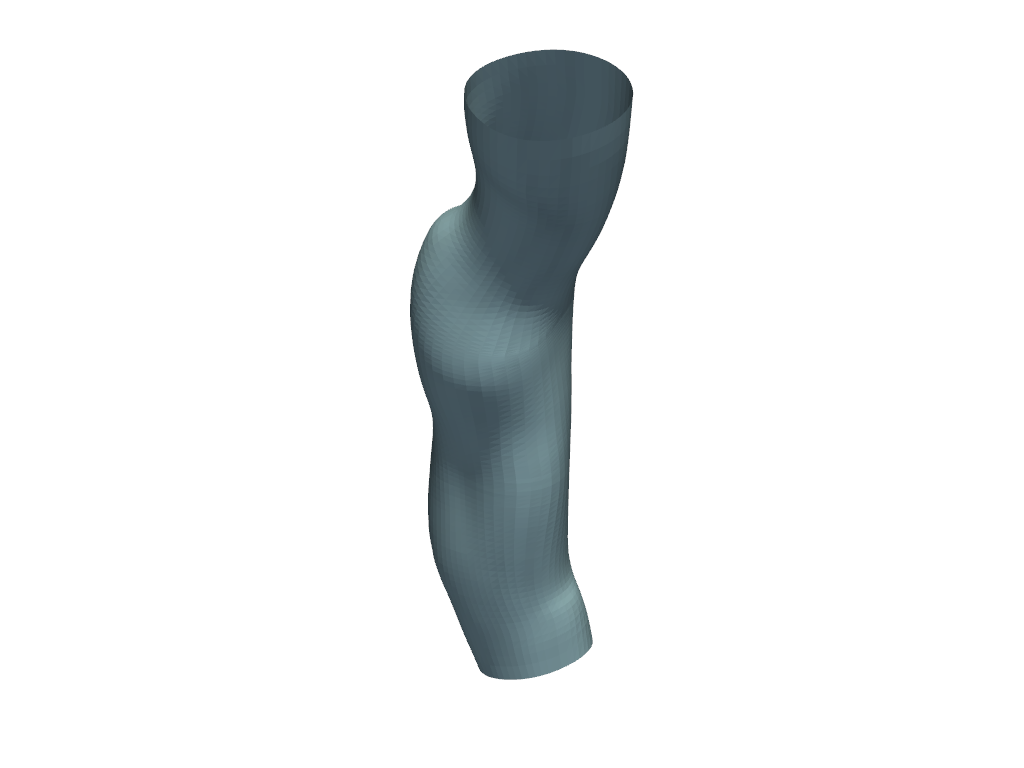}} \quad
    \subfloat[][]
    {\includegraphics[trim={11.75cm 1cm 11.75cm 1cm}, clip,width=0.12\textwidth]{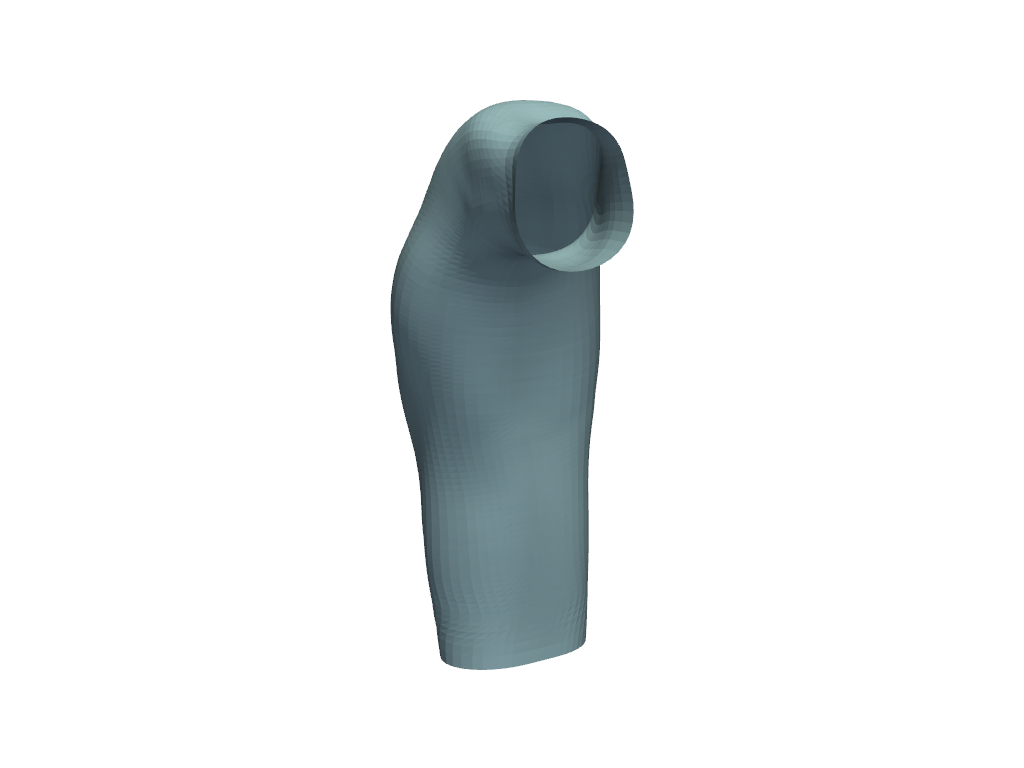}} \quad
    \subfloat[][]
    {\includegraphics[trim={11.75cm 1cm 11.75cm 1cm}, clip,width=0.12\textwidth]{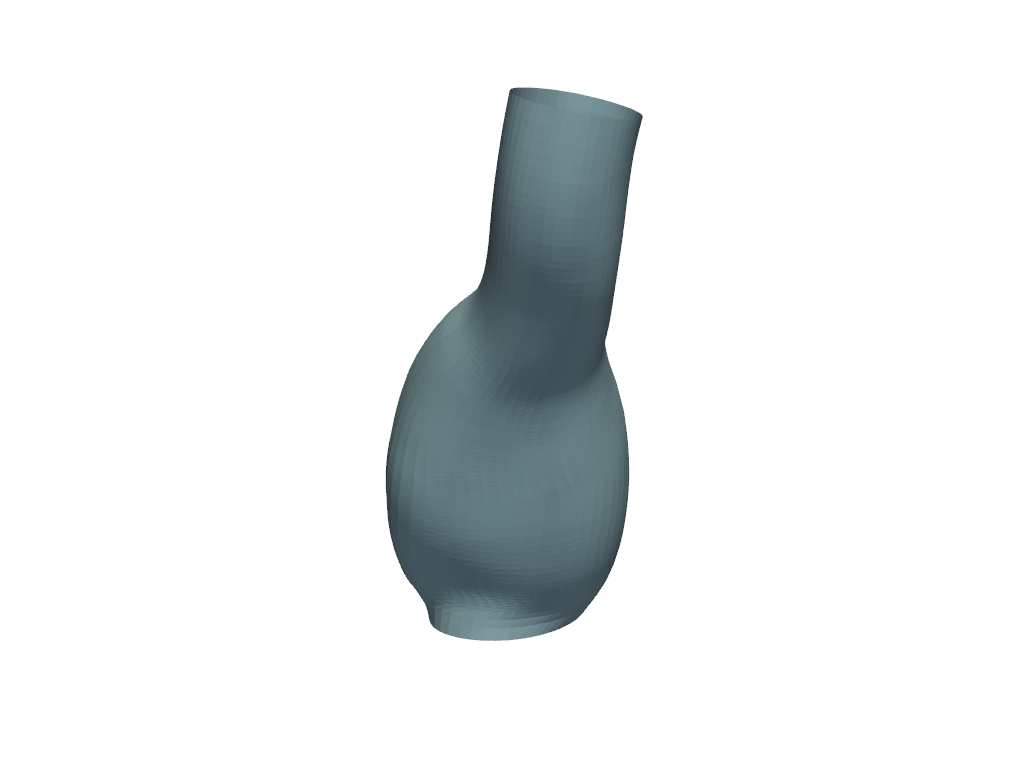}}
\end{subfigure}
\caption{Fifteen representative AAAs meshes selected from the complete dataset described in \Cref{sec:dataset}.}
\label{fig:meshes}
\end{figure}
\subsection{Dataset}\label{sec:dataset}
The dataset used is composed of 60 patient-specific three-dimensional (3D) geometric reconstructions of AAAs, which were initially obtained from pre-operative CT scans of patients scheduled for abdominal aortic aneurysm treatment. These original data consist of two-dimensional grayscale image slices from abdominal CT scans, capturing cross-sectional views of the patient's vascular anatomy \cite{MOLL2011S1, jin2023aibasedaorticvesseltree}. To construct the dataset, these 2D CT images are processed using specialized medical imaging software, \emph{PRAEVAorta} \cite{Caradu2021}, developed by \href{https://www.nurea-soft.com}{NUREA}, which performs automated segmentation of the vascular tree, isolating the aortic geometry from surrounding anatomical structures \cite{Riffaud2022, ravon:hal-03600571}. After initial segmentation, the specific anatomical region containing the aneurysm, located below the renal arteries and extending downwards approximately 30 mm beyond the iliac bifurcation, is automatically extracted using an algorithm designed to detect and separate these anatomical landmarks \cite{Riffaud2022,ravon:hal-03600571}. With the anatomical landmarks identified, the segmented aortic geometry is transformed into 3D surface meshes via the marching-cube algorithm, which converts segmented image volumes into polygonal surface representations \cite{10.1145/37402.37422}. The resulting surface meshes are then subjected to a smoothing procedure to remove artifacts and irregularities introduced during segmentation, utilizing the Vascular Modeling Toolkit (\texttt{vmtk}) Python library, enhancing their geometric accuracy and usability \cite{Piccinelli2009, Antiga2004}. By slicing it below the renal arteries and right before the iliac bifurcation, the final processed dataset thus contains detailed, patient-specific 3D surface models accurately representing the geometry of aneurysmal segments of the abdominal aorta, including features such as aneurysm sacs and associated vessel neck regions. \Cref{fig:meshes} visually illustrates fifteen examples of these final processed meshes, clearly depicting the anatomical variability captured by the dataset. The complete and detailed description of the entire pipeline, including image segmentation, geometry reconstruction, and surface processing steps outlined above, is described in detail in \cite{10.1007/978-3-031-55315-8_19}.

\subsection{\texorpdfstring{$\beta$-VAE Architecture}{Beta-VAE Architecture}}\label{sec:vae}
\begin{figure}[t]
    \centering
    \includegraphics[trim={6cm 0cm 6cm 0cm},clip,width=\textwidth]{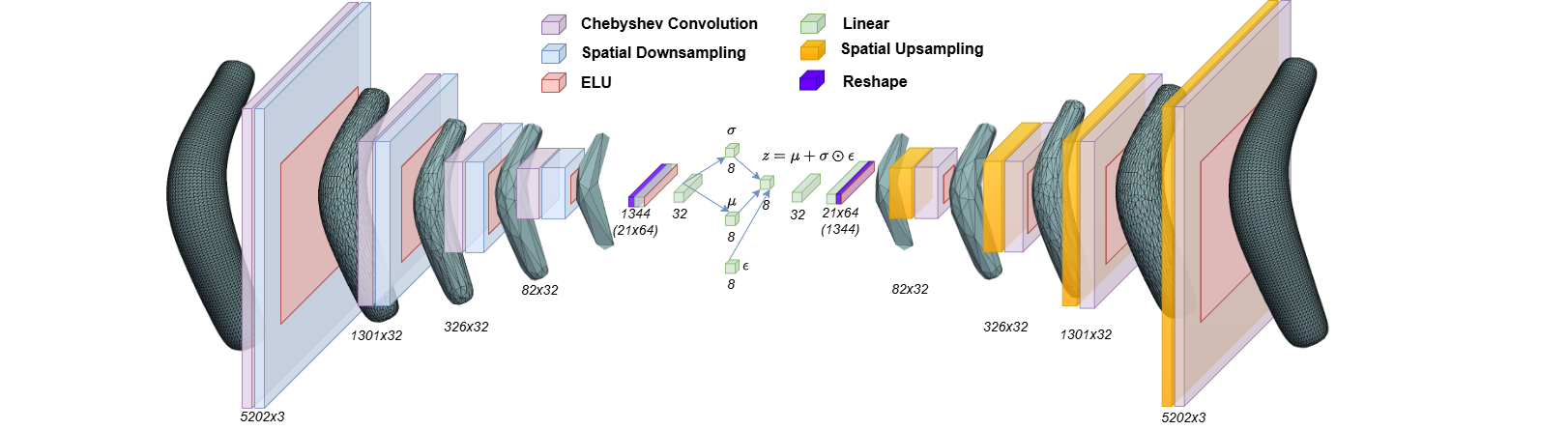} 
    \caption{Architecture of GCN-$\beta$-VAE used in the generation experiments. Meshes are downsampled and upsampled by a factor of 4 in each Spatial pooling layer (\Cref{sec:spatial-pooling}). Spectral graph convolutions, implemented using Chebyshev polynomials (\Cref{sec:graph-convolution}), are applied to the mesh data, transforming the 3 input channels through successive convolutional layers with output feature dimensions of $32 \rightarrow 32 \rightarrow 32 \rightarrow 64$. Additionally, linear layers are applied in the hidden layers preceding and following the latent space, as well as within the latent space itself. An Exponential Linear Unit (ELU) serves as the activation function throughout the network}
\label{fig:vae} 
\end{figure}

Variational Autoencoders (VAEs) represent a class of autoencoders specifically designed to learn efficient representations or codings of input data \cite{kingma2022autoencoding}. In contrast to traditional autoencoders that produce deterministic encodings, VAEs adopt a probabilistic framework. Rather than mapping an input directly to a single point in the latent space, the encoder of a VAE maps the input to the parameters of a probability distribution $p_\theta$  parameterized by $\theta$. A sample is then drawn from this distribution to obtain the latent representation, which is subsequently passed to the decoder. The decoder attempts to reconstruct the original input from this latent code, thereby enabling a generative modeling approach grounded in the principles of variational inference.

In practice, we aim to fit the following latent variable model to the data 
\[
p_\theta(x)=\int p_\theta(x|z)p_\theta(z)dz, 
\]
via maximum likelihood estimation (MLE) $\max \mathbb E_{p_{\mathrm{data}}(x)}[\log p_\theta(x)]$. Here,  $p_\theta(z)$ is the assumed prior probability distribution on the latent space, before data are taken into account, and $p_\theta(x|z)$ is the conditional probability of the data as a function of the latent space variables. Variational inference provides a variational lower bound for MLE and thus replaces MLE with the equivalent task of approximating the true posterior distribution $p_\theta(z|x)$ with a tractable distribution $q_\phi(z|x)$, parametrized by a neural network architecture forming the encoder module of the VAE (left module in \Cref{fig:vae}), just like the encoder $q_\phi(z|x)$, the conditional probability $p_\theta(x|z)$ is parametrized by a neural network model forming the decoder module of the VAE architecture (right module in \Cref{fig:vae}). To evaluate the statistical distance between the real and approximate posterior, the Kullback--Leibler divergence is employed,  
\[
D_{\mathrm{KL}}(q_\phi(z|x)||p_\theta(z|x)) = \int q_\phi(z|x)\log\frac{q_\phi(z|x)}{p_\theta(z|x)}dz.
\]
 Expanding this expression using Bayes' rule gives:

\[
D_{\mathrm{KL}}(q_\phi(z|x)\,\|\,p_\theta(z|x)) =
- \mathbb{E}_{z\sim q_\phi(z|x)}[\log p_\theta(x|z)] +
D_{\mathrm{KL}}(q_\phi(z|x)\,\|\,p_\theta(z)) +
\log p_\theta(x).
\]

This decomposition reveals two important terms that form the core of the variational approximation to MLE. The first term, \(\mathbb{E}_{z\sim q_\phi(z|x)}[\log p_\theta(x|z)]\), is the likelihood of the
observed data given the latent variables which is maximized to ensure the model accurately reproduces the input data from the latent space. The second term, \(D_{\mathrm{KL}}(q_\phi(z|x)\,\|\,p_\theta(z))\), is the KL divergence between the approximate posterior $q_\phi(z|x)$ and the prior $p_\phi(z)$, which is the assumed probability distribution of the latent space. This KL term acts like a regularizer, promoting a consistent latent space by penalizing deviations from the prior.
By rearranging terms, the resulting VAE loss function is expressed as:

\begin{equation}\label{eq:loss}
\begin{aligned}
L_{VAE}(\theta,\phi)
&= \mathbb{E}_{p_{\text{data}}(x)} \left[  
D_{\mathrm{KL}}(q_\phi(z|x)\,\|\,p_\theta(z|x)) - \log p_\theta(x)
\right] \\\\
&= \mathbb{E}_{p_{\text{data}}(x)} \left[  
-\mathbb{E}_{z\sim q_\phi(z|x)} \left[ \log p_\theta(x|z) \right]
+ D_{\mathrm{KL}} \big( q_\phi(z|x) \| p_\theta(z) \big)
\right].
\end{aligned}
\end{equation}

In the context of Variational Bayesian methods, the VAE loss corresponds to what is known as the evidence lower bound (ELBO). The KL divergence is always non-negative, and therefore $-L_{VAE}(\theta,\phi)$ is the lower bound of $\log p_\theta(x)$:

\[
-L_{VAE}(\theta,\phi)=\mathbb{E}_{p_{\text{data}}(x)}[\log p_\theta(x)-D_{\mathrm{KL}}(q_\phi(z|x)\,\|\,p_\theta(z|x))]\leq\mathbb{E}_{p_{\text{data}}(x)}[\log p_\theta(x)].
\]

Therefore, minimizing the loss function means maximizing the lower bound probability of generating realistic data.

\subsubsection{Sampling in the latent space}
The expectation term in the loss function invokes generating samples from $z\sim q_{\phi}(z|x)$. 
Sampling is inherently a stochastic process, which prevents straightforward gradient backpropagation. To enable end-to-end training, the reparameterization trick is employed. The encoder $q_\phi (z|x)$ is modeled to output a pair of parameters $q_\phi:x\longrightarrow (\mu, \sigma)$ for a given input $x$, representing the mean and standard deviation of a multivariate Gaussian distribution. The approximate posterior is defined as \( z \sim q_{\phi}(z|x) = \mathcal{N}(\mu, \sigma^2 I) \), while the prior distribution is set to a standard multivariate Gaussian \( z \sim p_{\theta}(z) = \mathcal{N}(0, I) \).

The reparameterization trick enables sampling by expressing \( z \) as:
\begin{align}
z = \mu + \sigma \odot \epsilon, \quad \text{with} \quad \epsilon \sim \mathcal{N}(0, I),
\label{eq:rep_tr}
\end{align}
where $\odot$ denotes elementwise multiplication. This formulation allows the model to learn the parameters $\mu$ and $\sigma$ through backpropagation, while maintaining stochasticity via the auxiliary noise variable $\epsilon$. The reparameterization trick is implemented by the green linear layers in \Cref{fig:vae}, which operate within the latent space of the VAE. By Substituting the Gaussian prior \( p_{\theta}(z) = \mathcal{N}(0, I) \) and the approximate posterior \( q_{\phi}(z \mid x) = \mathcal{N}(\mu, \sigma^2 I) \) into the KL divergence term yields the final expression for the VAE loss.
\[
L_{\text{VAE}}(\theta,\phi) 
= -\frac{1}{2} \sum_{i=1}^d \left( 1 + \log(\sigma_i^2) - \mu_i^2 - \sigma_i^2 \right) 
- \mathbb{E}_{z \sim q_\phi (z|x)} \left[\log p_\theta (x|z)\right].
\]

The proposed ${\beta}$-VAE \cite{higgins2017betavae} is a modification of the Variational Autoencoder discussed above, with a special emphasis on discovering disentangled latent factors. Following the same arguments as for the VAE, the objective is to maximize the probability of generating real data while keeping the distance between the real and estimated posterior distributions small. To achieve this, an additional hyperparameter $\beta$ is introduced as a weighting factor on the KL divergence term. Proper tuning of $\beta$ favors a more orthogonal and disentangled latent space representation:

\begin{equation}
\begin{aligned}
L_{\beta\text{-VAE}}(\theta,\phi) = 
&-\frac{\beta}{2} \sum_{i=1}^d \left( 1 + \log(\sigma_i^2) - \mu_i^2 - \sigma_i^2 \right) 
- \mathbb{E}_{z \sim q_\phi (z|x)} [\log p_\theta (x|z)].
\end{aligned}
\label{eq:general_beta_loss_function}
\end{equation}

Increasing the value of \( \beta \) encourages the model to learn such efficient and disentangled latent representations. However, if \( \beta \) is set too high, it may introduce a trade-off, where improved disentanglement comes at the cost of degraded reconstruction quality.

\subsubsection{Combining regularization and reconstruction loss}\label{sec:rec-loss}

The chosen training loss combines the regularization KL-based term from the $\beta$-VAE, as described above, with a reconstruction component that encourages the generated outputs to match the geometries present in the training data. The reconstruction component \( \mathbb{E}_{z \sim q_\phi (z|x)} \log p_\theta (x|z) \) is formulated as a weighted combination of multiple loss terms, each capturing different geometric aspects of the 3D mesh structure. This composite design allows for flexibility in adapting the loss to the specific characteristics of the data and the reconstruction task. By exploiting the inherent connectivity of mesh graphs, higher-order loss terms are defined over local neighborhoods of vertices, enabling more effective regularization of complex 3D shapes.
\cite{wang2018pixel2meshgenerating3dmesh,choi2021pose2meshgraphconvolutionalnetwork}. 
Specifically, the followings components are incorporated: 
\begin{itemize}[leftmargin=*]
\item{Vertex Coordinate Loss:} Minimizes the L1 distance between the predicted 3D mesh coordinates $M$ and the ground truth coordinates $M^*$, providing a robust constraint that penalizes local deviations in vertex positions:
\[
L_{\text{vertex}} = \|M - M^*\|_1,
\]
where $M^*$ denotes the ground truth mesh coordinates, $ \| \cdot \|_1 $ represents the L1 norm.
\item{Chamfer Loss:} Computes how far each point in one set is from the closest point in the other set, which ensures that the reconstructed shape closely matches the overall structure and distribution of the target geometry:
\[
L_{\text{chamfer}} = \sum_{x \in M} \min_{y \in M^*} \| x - y \|_2^2
+  \sum_{y \in M^*}  \min_{x \in M} \| y - x \|_2^2,
\]
where the term \( \| \cdot \|_2 \) denotes the L2 norm, used to calculate the Euclidean distance between two vertices.

\item{Normal Loss:} Enforces the consistency between the normal vectors of the predicted mesh and the corresponding ground truth normals, thereby promoting smooth surfaces and enhanced local detail. The surface normal loss, \(L_{\text{normal}}\), is defined as follows:
\[
L_{\text{normal}} = \sum_{f} \sum_{\{i,j\} \subset f} \left\langle \frac{x_i - x_j}{\|x_i - x_j\|_2}, n^{*}_f \right\rangle,
\]
where \(f\) represents a triangle face on the mesh, and \(n^{*}_f\) is the unit normal vector of the corresponding ground truth face. The terms \(x_i\) and \(x_j\) refer to the \(i\)-th and \(j\)-th vertices of the triangle face \(f\).
This loss enforces the edge between a vertex and its neighbours to be perpendicular to the surface normal from the ground truth.
\item{Edge Loss:} Enforces the consistency of edge lengths between the predicted mesh and the ground truth mesh. This loss is particularly effective for recovering smoothness in regions with dense vertices, as well as penalizing flying vertices, which can cause long edges. The edge loss \( L_{\text{edge}} \) is defined as follows:
\[
L_{\text{edge}} = \sum_{f} \sum_{\{i,j\} \subset f} \left| \|x_i - x_j\|_2 - \|x^{*}_i - x^{*}_j\|_2 \right|,
\]
where \( f \) represents a triangle face on the mesh, \( x_i \) and \( x_j \) are the vertices of the predicted mesh, and \( x^{*}_i \) and \( x^{*}_j \) denote the corresponding vertices of the ground truth mesh. 
\end{itemize}

The final loss is then written as:
\begin{equation}
L(\theta,\phi) = -\frac{\beta}{2} \sum_{i=1}^d \left( 1 + \log(\sigma_i^2) - \mu_i^2 - \sigma_i^2 \right)  
+ \alpha_{\text{vertex}} L_{\text{vertex}} 
+ \alpha_{\text{chamfer}} L_{\text{chamfer}} 
+ \alpha_{\text{edge}} L_{\text{edge}} 
+ \alpha_{\text{normal}} L_{\text{normal}}.
\label{eq:training_loss}
\end{equation}

Here, the coefficients \( \{ \alpha_{\text{label}} \} \) are hyperparameters that regulate the trade-off between the contributions of the individual loss terms. During training, these are set as \( \alpha_{\text{chamfer}} = \alpha_{\text{vertex}} = 1 \) and \( \alpha_{\text{edge}} = \alpha_{\text{normal}} = 0.1 \) for all configurations of latent size and \( \beta \).

\subsubsection{Spectral Graph Convolution}\label{sec:graph-convolution}
A 3D aneurism mesh is represented as an undirected graph $M = (\mathcal{V}, \mathcal{E}, \mathcal{A})  $, where  $\mathcal{V}=\{v_i\}^N_{i=1}$ is the set of \( N =|\mathcal{V}| \) edges vertices, $\mathcal{E}=\{e_j\}^E_{j=1}$ the set of  \( E =|\mathcal{E}| \) edges, and  $\mathcal{A}=(\mathbf{a}_{ij})_{N\times N}$ the adjacency matrix, where $\mathbf{a}_{ij}=0$ if $(i,j)\notin\mathcal{E}$ and $\mathbf{a}_{ij}\neq 0$ if $(i,j)\in\mathcal{E}$. One can also compute the symmetric normalized graph Laplacian as $\mathcal{L} = \mathcal{I}_N - (\mathcal{D}^{+})^{1/2}\mathcal{A}(\mathcal{D}^{+})^{1/2}$ where $\mathcal{D}^{+}$ is the Moore-Penrose inverse of the diagonal matrix $\mathcal{D}=diag( \sum_j\mathbf{a}_{ij})$ and $\mathcal{I}_N$ is the identity matrix of size $N$. 
The normalized Laplacian in its spectral basis has the form $\mathcal{L}=\mathcal{U}\Lambda\mathcal{U}^T$ where $\mathcal{U}\in\mathbb{R}^{N\times N}$ is the matrix of the orthonormal eigenvectors of $\mathcal{L}$ and $\Lambda$ is the matrix of the eigenvalues of $\mathcal{L}$. A signal on the nodes (vertices) of the mesh graph is defined as $\mathcal{S}\in\mathbb{R}^{N \times F}$ where the number of features $F$, before applying any of the convolutional filter operations, is equal to $3$, which is equal to the number of dimensions in which the mesh exists. The value $\mathcal{S}^j$ is the value of the $j^{th}$ feature element of $\mathcal{S}$ on the nodes. 
The fast spectral filtering method \cite{defferrard2017convolutionalneuralnetworksgraphs}, previously used in \cite{Ranjan_2018_ECCV}, is employed. An input feature on a node $\mathcal{S}_{F_{in}}$ is transformed into $\mathcal{S}_{F_{out}}$ using a recursive filter kernel $f$ based on Chebyshev polynomials of order $K$, as follows:
\begin{align*}
f_{\theta}(\mathcal{L})=\sum_{k=0}^{K-1}\theta_k\mathcal{T}_k(\Bar{\mathcal{L}}) ,
\end{align*}
where $\Bar{\mathcal{L}}$ is the shifted and rescaled Laplacian $\Bar{\mathcal{L}}=2\mathcal{L}/\lambda_{max}-I_N$, normalized by its largest eigenvalue $\lambda_{max}$ and shifted by the identity matrix;   $\theta\in\mathbb{R}^K =(\theta_0,\dots,\theta_{K-1})$ is a vector of learnable coefficients; and  $\mathcal{T}_k$ denotes the $k$-th Chebyshev polynomial of the first kind, defined recursively as $ \mathcal{T}_k(x) = 2x\mathcal{T}_{k-1}(x)-\mathcal{T}_{k-2}(x)$ with $\mathcal{T}_1 = x$ and $\mathcal{T}_0 = 1$.
The output signal would be written as:
\begin{align*}
\mathcal{S}^{j}_{F_{out}}=\sum_{i=1}^{F_{in}}f_{\theta_{i,j}}(\mathcal{L})S^{i}_{F_{in}} ,
\end{align*}
where \(F_{\mathrm{in}}\) denotes the number of input features per node. The index \(j\) runs over the output features being computed by the filter, the \(j\)-th output feature channel \(S_{F_{\mathrm{out}}}^{j}\) is formed by summing over all \(F_{\mathrm{in}}\) input channels. Each term in the sum consists of the Chebyshev-based filter \(f_{\theta_{i,j}}(\mathcal{L})\), which transforms the \(i\)-th input channel to the \(j\)-th output channel, multiplied by the \(i\)-th input feature \(S_{F_{\mathrm{in}}}^{i}\). This spectral convolution operation, based on Chebyshev polynomials, is implemented as the purple Chebyshev convolution layers in the architecture diagram \Cref{fig:vae}
.

\subsubsection{Spatial Pooling}\label{sec:spatial-pooling}
Spatial pooling is a down-sampling operation commonly used to combine small-scale (local) and large-scale (global) information by reducing the spatial resolution of data, typically in image-based neural networks. In essence, pooling summarizes local information to create progressively simplified representations. However, for mesh data, spatial pooling becomes more complex because meshes do not have a regular grid structure like images, and thus, simplification must respect geometric and topological constraints.
Spatial pooling on meshes is implemented using a simplification algorithm based on iterative vertex pair contraction \cite{10.1145/258734.258849}. Specifically, a pair contraction, represented as $(v_1, v_2) \to \bar{v}$, merges two vertices $v_1$ and $v_2$ to a new position $\bar{v}$, connects all their incident edges to $v_1$, and removes the vertex $v_2$. Subsequently, any edges or faces that become degenerate are removed. Because contractions are highly localized operations, applying them iteratively generates a series of simplified models $M_N, M_{N-1}, \dots, M_g$, starting from the initial model $M_N$ down to a final approximation $M_f$.

The in-network down-sampling method from \cite{Ranjan_2018_ECCV} is implemented on a mesh initially containing $N$ vertices, using transformation matrices $Q_d \in {0, 1}^{n \times N}$ for down-sampling and $Q_u \in \mathbb{R}^{N \times n}$ for up-sampling, where $N > n$.  In the architecture diagram \Cref{fig:vae}, the matrix-based down-sampling and up-sampling operations performed via $Q_d$ and $Q_u$, respectively are implemented as the azure-colored downsampling layers and the yellow upsampling layers, which appear symmetrically on the encpder-decoder structure. The down-sampling process iteratively merges vertex pairs while preserving surface detail through quadric error matrices. After down-sampling, the resulting vertex set $V_d$ forms a subset of the original set $V$, such that $V_d \subset V$. The elements of $Q_d(p, q)$ indicate whether the $q$-th vertex is retained ($Q_d(p, q) = 1$) or discarded ($Q_d(p, q) = 0$ for all $p$).
Since perfect down-sampling and up-sampling cannot be achieved for arbitrary surfaces, the up-sampling matrix is designed concurrently with the down-sampling step. The vertices retained during down-sampling undergo transformations and are similarly retained during up-sampling, where $Q_u(q, p) = 1$ if $Q_d(p, q) = 1$. Vertices that are discarded during the down-sampling stage are mapped back onto the simplified mesh surface using barycentric coordinates. The discarded vertex $v_q$ is projected onto the nearest triangle formed by vertices in $V_d$, and its position is calculated using the barycentric formula $v_{\text{ep}} = w_i v_i + w_j v_j + w_k v_k$, where $v_i, v_j, v_k \in V_d$ and the weights satisfy $w_i + w_j + w_k = 1$. These weights are then used to update the up-sampling matrix as $Q_u(q, i) = w_i$, $Q_u(q, j) = w_j$, and $Q_u(q, k) = w_k$, while setting $Q_u(q, l) = 0$ for all other vertices. The final up-sampled mesh $V_u$ is obtained through sparse matrix multiplication, $V_u = Q_u V_d$.

\subsection{ProcAug Augmentation}\label{sec:procau}
In many computer vision tasks, data augmentation has proven to be an essential technique for improving model robustness and generalization. Augmentation pipelines rely on elementary transformations, such as random rotations, random crops, flips, and  jittering. Recent works have moved toward more sophisticated and automated augmentation strategies, for example, AutoAugment \cite{Cubuk_2019_CVPR}  and RandAugment \cite{cubuk2019randaugmentpracticalautomateddata}, where a search algorithm or a simplified policy is employed to discover optimal augmentations.

\begin{figure}[t]
    \centering
    \begin{minipage}{.45\textwidth}
    \includegraphics[trim={31cm 3cm 33cm 5cm},clip,width=.49\linewidth]{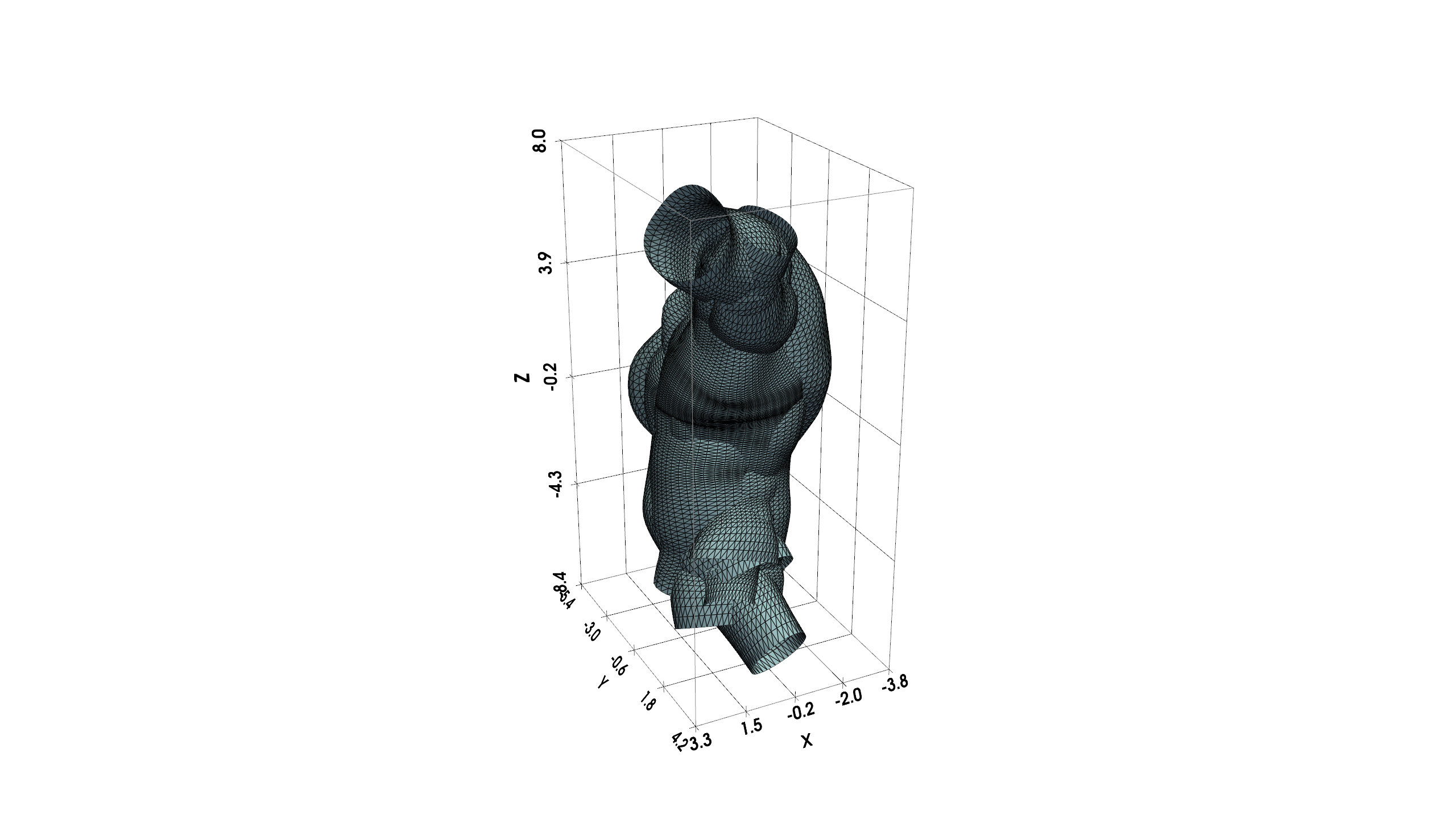}
    \includegraphics[trim={32cm 3cm 32cm 5cm},clip,width=.49\linewidth]{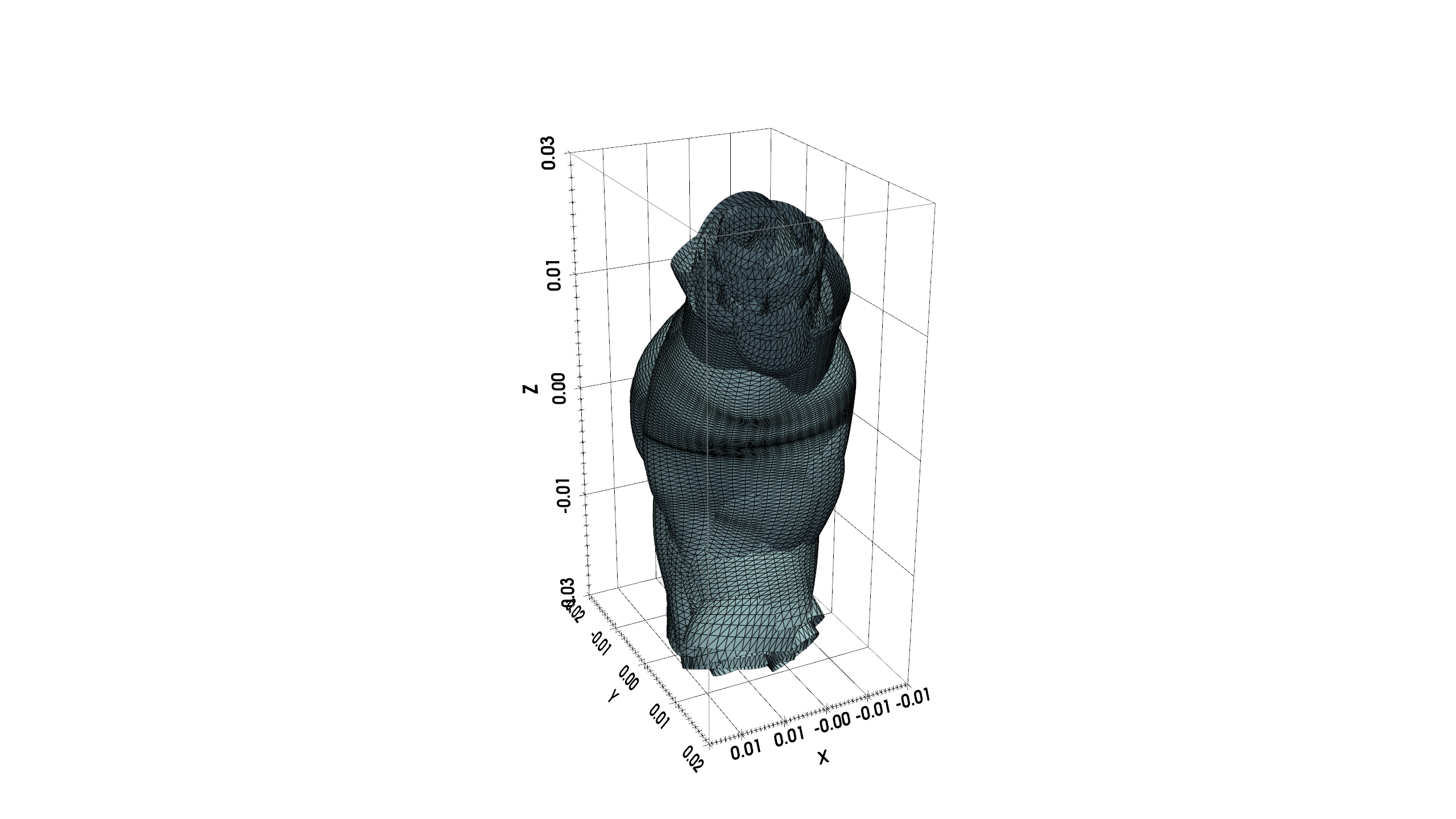}
    \end{minipage}
    \hspace{.5cm}
    \begin{minipage}{.4\textwidth}
        \includegraphics[trim={8cm 5cm 1cm 1cm},clip,width=\linewidth]{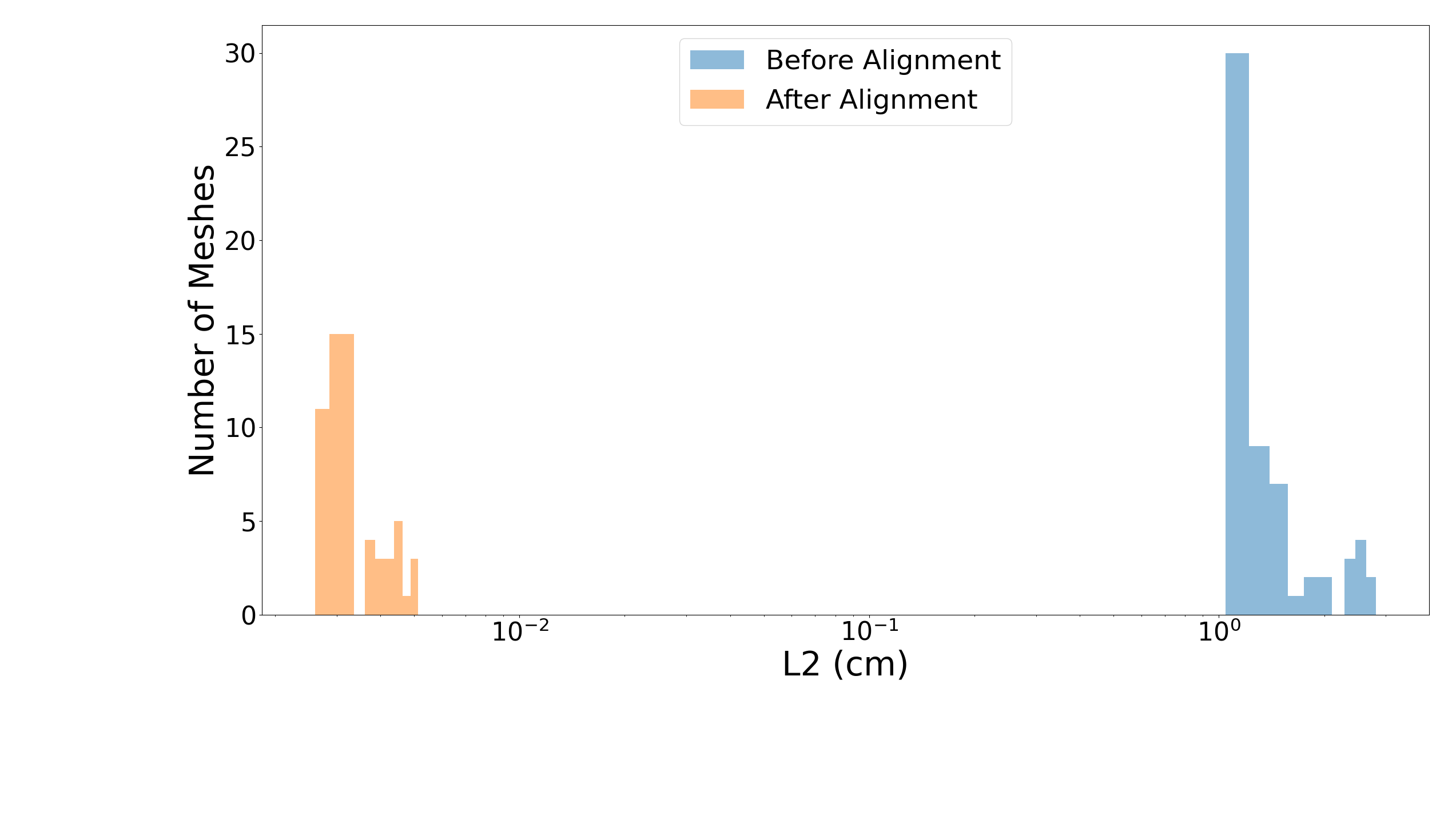}\\
    \includegraphics[trim={10.8cm 6.9cm 1cm 1cm},clip,width=\linewidth]{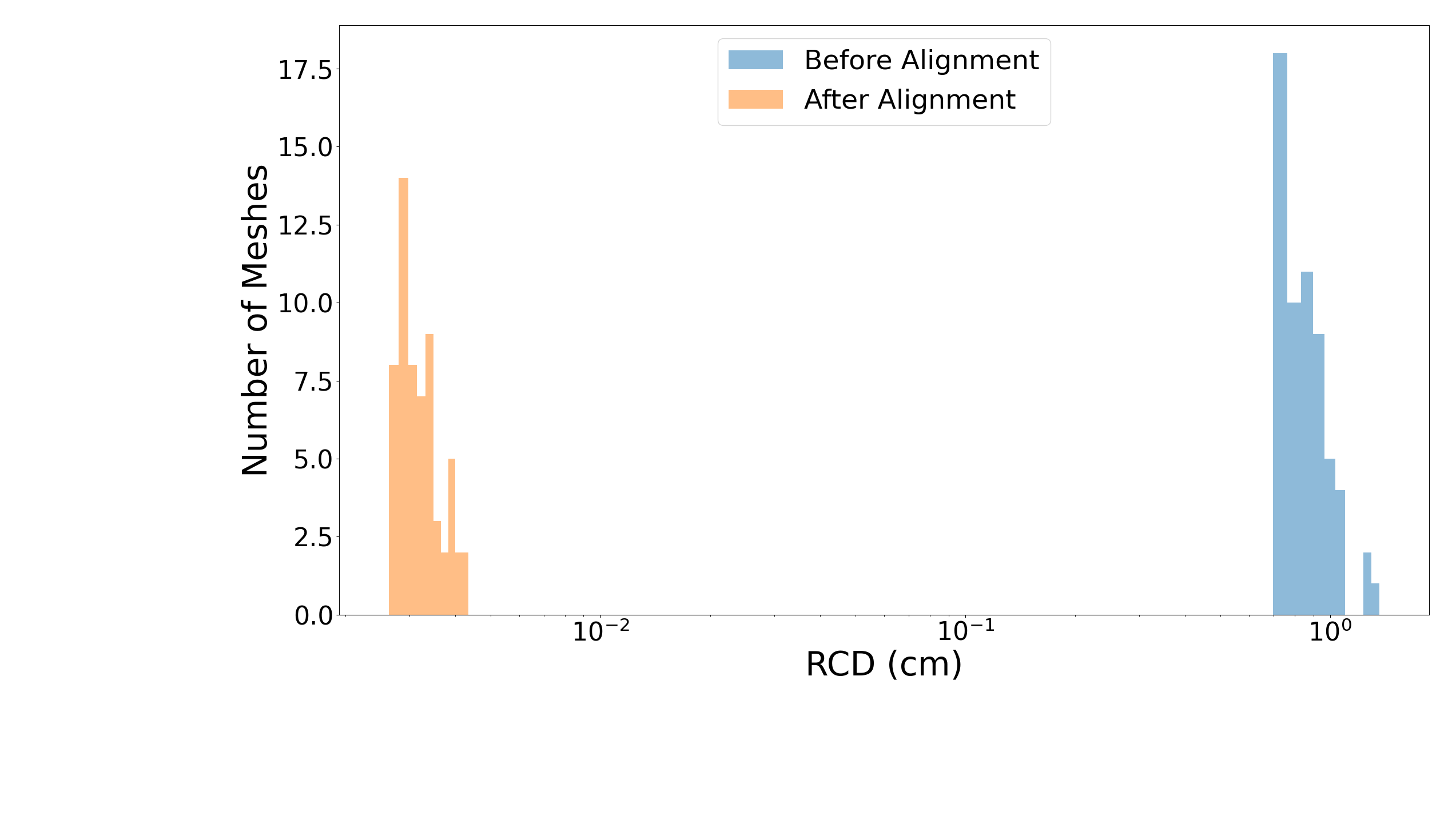}
    \end{minipage}
\caption{Original real meshes (left) Corresponding meshes after Procrustes alignment (right). The alignment removes differences in scale, rotation, and translation, bringing the meshes into a common reference frame.
    \textbf{(b)} Histograms comparison of reconstruction errors before and after Procrustes alignment.  
The plots show the distribution of Chamfer Distance (bottom) and L2 error (top) for meshes before alignment and after alignment.  
After Procrustes alignment, both error distributions become more concentrated around zero, indicating a reduction in variability.}
    \label{fig:preproc_meshes}
\end{figure}

Procrustes analysis \cite{Gower_1975,Bookstein_1992} has gained attention in shape analysis due to its ability to align and compare data by reducing the effects of translations, scales, and rotations. More specifically, it seeks a transformation that best superimposes one set of points onto another, effectively capturing the intrinsic shape while normalizing nuisance factors. This property makes Procrustes analysis a powerful tool for quantifying and transferring geometric transformations across datasets,  as illustrated in \Cref{fig:preproc_meshes}, this alignment significantly reduces variability in shape representations by minimizing reconstruction errors such as Chamfer Distance and L2 error distributions..

Procrustes analysis and data augmentation can be understood as two sides of the same coin in that they deal with the same transformations, translations, rotations, and scalings, but in opposite ways. Procrustes analysis identifies and removes these variations to align data points in a common coordinate system, effectively standardizing the shapes so that their intrinsic structure is more directly comparable. While this process reduces the variability of the dataset,  it can also remove some precious morphological details that may be essential for capturing the true range of anatomical differences in AAA. Data augmentation, on the other hand, deliberately applies these transformations (and others, such as jitter or translations) to existing data points, thereby increasing variability in a controlled way. The aim of augmentation is not to eliminate transformations but to introduce them, expanding the training set and making models more robust to real-world variations. Thus, while Procrustes analysis seeks to unify or harmonize data by subtracting differences, data augmentation enlarges the feature space by creating new instances that share essential properties of the original data, ultimately illustrating how each approach works with the same underlying transformations but pursues fundamentally opposite goals.

At a high level, the idea is to extract real-valued transformations (e.g., rotation angles, scaling factors) from data alignment. These transformations can then be randomly sampled and applied to incoming batches during training. This approach blends the flexibility of RandAugment, where specific augmentation types can be randomly selected, with the geometrically grounded transformations from Procrustes analysis. The result is a method that can preserve the underlying structure of meshes while still introducing valuable variability that encourages the model to learn more robust representations.

Procrustes analysis normalize the AAA mesh, represented by the matrix $M\in\mathbb{R}^{N\times3}$ to have unit norm $M=\frac{M}{\|M\|_F}$ as well as the reference geometry $M_{ref}=\frac{M_{ref}}{\|M_{ref}\|_F}$ . Then, Procrustes analysis solves the orthogonal Procrustes problem by minimizing:
\begin{equation}
    \min_{\Omega} \| M \Omega - M_{ref} \|_F,
\end{equation}
where $\Omega$ is the rotation matrix which most closely maps $M$ to $M_{ref}$.
This problem admits an explicit solution \cite{Gower_1975}. Using the Singular Value Decomposition (SVD), matrices \(U\), \(\Sigma\), and \(V^T\) are obtained as follows: \(U, \Sigma, V^T = \text{SVD}(M^T M_{ref})\), where \(U\) and \(V\) are orthonormal and \(\Sigma\) is diagonal. Thus the optimal rotation matrix \( \Omega^* \) is given by:
\begin{equation}
    \Omega^* = U V^T \, .
\end{equation}

Both the information on the scaling, as well as the rotation can be used as data augmentation parameters, in fact in 3D, a rotation can be parameterized by three Euler angles \cite{bernardes2022quaternion} (e.g., \(\psi, \xi, \gamma\)) around the \(x\), \(y\), and \(z\) axes, respectively. Suppose we observe or estimate that each shape \(M_i\) is rotated by angles within some range w.r.t. the reference shape:
\[
\psi_{\min} \;\leq\; \psi_i \;\leq\; \psi_{\max}, 
\quad
\xi_{\min} \;\leq\; \xi_i \;\leq\; \xi_{\max},
\quad
\gamma_{\min} \;\leq\; \gamma_i \;\leq\; \gamma_{\max}.
\]
Each value of Euler angles can then be sampled and utilized to rotate the training geometry.
\[
\psi_{rand} \;\sim\; \mathcal{U}(\psi_{\min},\,\psi_{\max}), 
\quad
\xi_{rand} \;\sim\; \mathcal{U}(\xi_{\min},\,\xi_{\max}), 
\quad
\gamma_{rand} \;\sim\; \mathcal{U}(\gamma_{\min},\,\gamma_{\max}),
\]
where \(\mathcal{U}\) denotes a uniform distribution. The resulting 3D rotation matrix, \(\Omega_{\text{rand}}\), is constructed via the standard Euler angle to rotation matrix composition. For instance,selecting the the \(Z\!Y\!X\) Euler angles ordering:
\[
\Omega_{\text{rand}} 
\;=\;
\Omega_z(\gamma_{rand}) \, \Omega_y(\xi_{rand}) \, \Omega_x(\psi_{rand}),
\]
where each \(\Omega_x(\psi), \Omega_y(\xi), \Omega_z(\gamma)\) is a \(3 \times 3\) rotation matrix about its respective axis.
Finally, the augmented norm is defined:
\[M_{aug}=M_i\Omega_{\text{rand}}^{T}.\]
Similarly, a range of values is defined based on the norms:
\[s_{\min} = \|M\|_{\min} \;\leq\; \|M_i\|  \;\leq\; \|M\|_{\max} = s_{\max},\]
where $\|M\|_{\max}$ and $\|M\|_{\min}$ are the largest and smallest norm values among the training set, and sample them during training
\[
s_{\text{rand}} \;\sim\; \mathcal{U}(s_{\min},\,s_{\max}),
\]
the augmented data is then written as:
\[M_{aug}=M_i\frac{s_{rand}}{\|M_i\|}.\]

This approach allows to systematically generate augmented training samples by varying both scale and orientation, enhancing the robustness of the model. Moreover, inspired by RandAugment \cite{Cubuk_2019_CVPR}, a parameter-free augmentation procedure is adopted to simplify policy search while maintaining diversity in the augmented samples. Instead of learning augmentation policies and probabilities for each transformation, each transformation is applied with a uniform probability of $\frac{1}{K}$, where $K = 3$, including identity, scaling, and rotation.

\section{Results}\label{sec:results}
The following section evaluates the performance of the model and outlines the construction of a meaningful latent space suitable for generative experiments. The analysis begins with the impact of the augmentation strategy, ProcAug, in \Cref{subsec:procperf}. From \Cref{sec:rec_out} to \Cref{subsec:hlc}, key properties of the latent space are systematically investigated. Specifically, \Cref{sec:rec_out} assesses the model’s out-of-sample generalization capability by comparison with a baseline PCA approach. \Cref{sec:dis_lat} examines the disentanglement properties of the learned latent space, focusing on the statistical independence of its representations. \Cref{subsec:hlc} introduces the concept of hierarchical latent mode contribution to identify how different latent dimensions capture distinct modes of variation in the data. Finally, based on insights from these analyses, the best-performing model identified in earlier sections is used for generative in-sample experiments in \Cref{sec:gen}.

Given the frequent shortage of data encountered in the medical field, the ability of machine learning models to generalize effectively from limited datasets is critical. Before evaluating the generative capacity of the model, its generalization capabilities are first discussed, as these capabilities determine the accuracy on unseen data and potential applicability in real-world scenarios. Accordingly, k-fold validation is conducted on all metrics presented in the following section with $k=10$, averaging the values obtained from each split. This out-of-sample analysis importantly guides the subsequent generation of novel and meaningful data.

Specifically, the trained GCN-$\beta$-VAE model is utilized on the in-sample dataset to produce new data. The ultimate goal is to derive a latent representation of AAA
that is compact, disentangled, and capable of accurately reconstructing unseen data, thereby validating the effectiveness of the model’s generative and representational capacities. During reconstruction, the reparameterization trick (\Cref{eq:rep_tr}) simplifies to using only the mean, $\boldsymbol{z}=\boldsymbol{\mu}$.

The quality of reconstruction serves as an essential measure of the model’s ability to capture the underlying geometry and topology of the input data. To quantify reconstruction quality, both the $L_2$ norm and the $L_2$ reconstruction percentage are employed, defined as follows:
\begin{equation}
    E = 
\left(
1 - \frac{\| M - M^* \|_2}{\| M^* \|_2}
\right) \times 100
\quad
\text{where } \|x\|_2 = \sqrt{\sum_i x_i^2},
\label{eq:reconstruction}
\end{equation}

where \( M^* \) represents the set of original data points and $M$ represents the set of reconstructed data points. 
Additionally, the Chamfer distance \cite{9127813,NEURIPS2020_4c5bcfec}, already defined in the loss function \Cref{eq:training_loss}, is employed and treated as an absolute error:
\begin{equation}
\mathrm{Err}{\mathrm{CD}} = \sum{x \in M} \min_{y \in M^*} | x - y |2^2 + \sum{y \in M^*} \min_{x \in M} | y - x |_2^2.
\label{eq:chamfer}
\end{equation}
To enhance interpretability and comparability across metrics, the Root Chamfer Distance (RCD) is also reported, defined as:
\begin{equation}
RCD = \sqrt{\mathrm{Err}_{\mathrm{CD}}},
\label{eq:RCD}
\end{equation}
thus making it more directly comparable to the $L_2$ norm.

To evaluate the stastical independence of the latent representation variables, the KL-Divergence defined previously in \Cref{eq:training_loss} is employed, as well as the determinant of the correlation matrix \( \mathbf{R}\in\mathbb{R}^{d\times d} \) times $100$, denoted as \( \det(\mathbf{R}) \), which is defined as:
\begin{equation}
\mathbf{R} = (R_\mathrm{ij})_{d \times d},
\end{equation}
with components:
\[
R_{ij} =
\begin{cases} 
1 & \text{if } i = j, \\
\frac{C_{ij}}{\sqrt{C_{ii} C_{jj}}} & \text{if } 1 \leq i \neq j \leq d,
\end{cases}
\label{eq:correlation_matrix}
\]
where $d$ is the latent size dimension and \( C_{ij} \) represents the \( (i, j) \)-th component of the covariance matrix \( \mathbf{C} \in \mathbb{R}^{d \times d
} \) defined as: 
\[\mathbf{C}=\mathbb{E}\left[ (\mathbf{X} - {\mathbb{E}[\mathbf{X}]})(\mathbf{X} - {\mathbb{E}[\mathbf{X}]})^T \right].\]
Where, given \(n\) observations of the latent space $\{ \boldsymbol{\mu}^{(i)} \}_{i=1}^{n} \subset \mathbb{R}^d$, one can define, the latent matrix: \(\mathbf{X}\in\mathbb{R}^{n \times d}\) where each row is an observation and each column is a variable of the latent space.
\subsection{ProcAug performance}\label{subsec:procperf}
\begin{table}[t]
    \centering
    \caption{Comparison of the reconstruction performance of VAE (Euclidean norm and RCD) trained with different data-augmentation procedures. The results show the mean over 10 random foldings. 
    }
    \label{tab:l2_chamfer_comparison}
    \begin{tabular}{cccccc}
        \toprule
        \multicolumn{6}{c}{\textbf{$L_2$ (cm)}} \\
        \midrule
        \textbf{Latent Size} & \textbf{ProcAug} & \textbf{Only Scaling} & \textbf{Only Rotation} & \textbf{No ProcAug} & \textbf{\% Diff (ProcAug)} \\
        \midrule
        4  & 0.376 & 0.447 & 0.436 & 0.447 & \textbf{-15.93\%} \\
        8  & 0.251 & 0.258 & 0.272 & 0.275 & \textbf{-8.46\%} \\
        12 & 0.224 & 0.226 & 0.241 & 0.254 & \textbf{-11.96\%} \\
        16 & 0.207 & 0.209 & 0.221 & 0.237 & \textbf{-12.69\%} \\
        20 & 0.195 & 0.197 & 0.212 & 0.222 & \textbf{-12.22\%} \\
        24 & 0.184 & 0.194 & 0.196 & 0.227 & \textbf{-18.90\%} \\
        \toprule
        \multicolumn{6}{c}{\textbf{RCD (cm)}} \\
        \midrule
        4  & 0.373 & 0.427 & 0.421 & 0.431 & \textbf{-13.52\%} \\
        8  & 0.258 & 0.267 & 0.285 & 0.277 & \textbf{-6.70\%} \\
        12 & 0.237 & 0.240 & 0.248 & 0.260 & \textbf{-8.99\%} \\
        16 & 0.224 & 0.234 & 0.234 & 0.254 & \textbf{-11.88\%} \\
        20 & 0.218 & 0.222 & 0.229 & 0.245 & \textbf{-10.90\%} \\
        24 & 0.209 & 0.220 & 0.220 & 0.248 & \textbf{-15.78\%} \\
        \bottomrule
    \end{tabular}
\end{table}
\begin{figure}[!htb]
    \centering
    \includegraphics[trim={4cm 0cm 5cm 0cm},clip,width=0.7\textwidth]{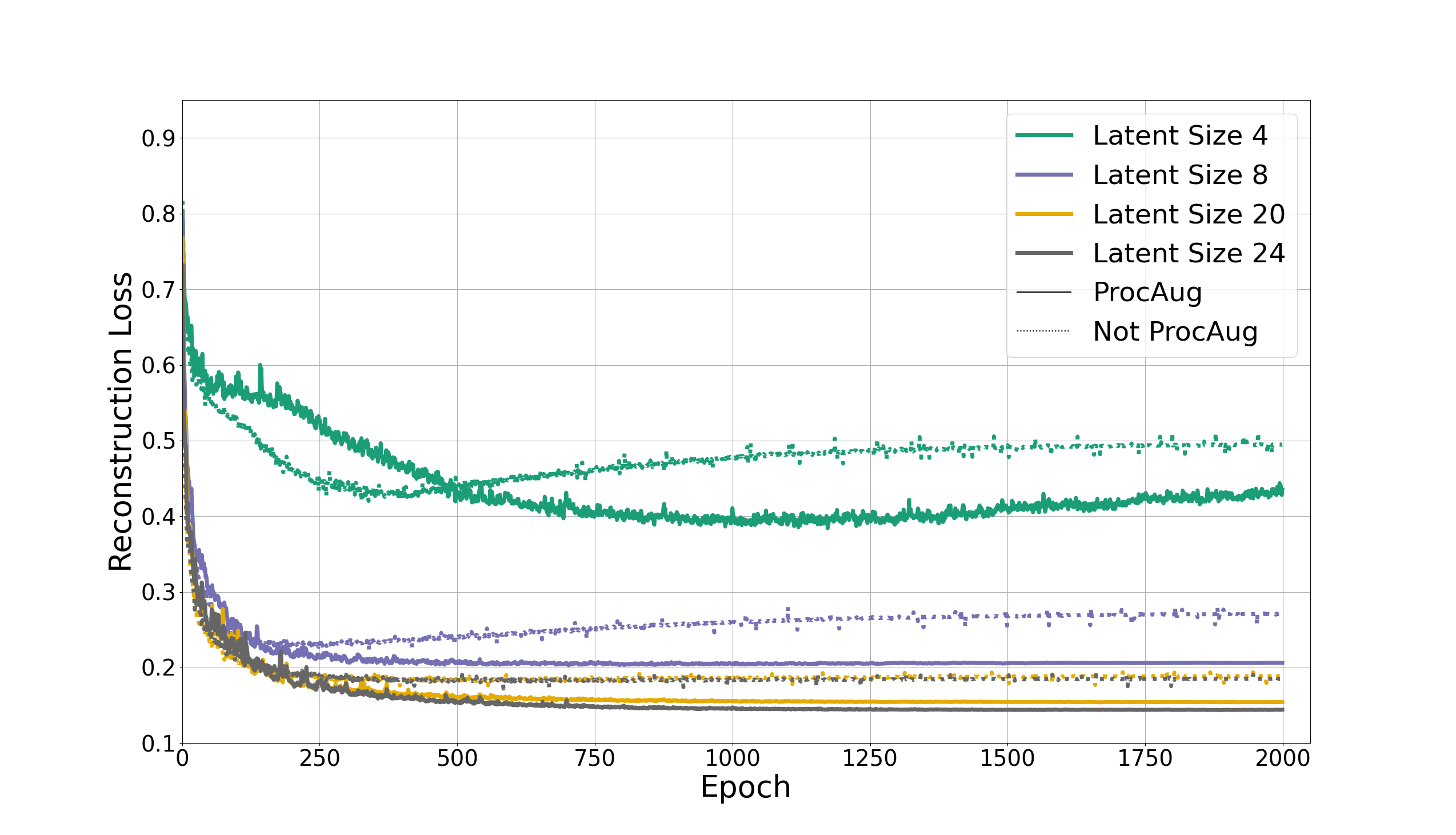}
    \caption{Test Reconstruction error vs. epochs for different latent sizes, comparing ProcAug and NoProcAug. ProcAug consistently achieves lower reconstruction error and convergence across various latent sizes. Moreover, NoProcAug exhibits spikes in the error curve; in contrast, ProcAug, especially for higher latent sizes, results in smoother curves, indicating a more stable reconstruction performance. The results are k-folded 10 times.}
    \label{fig:recon_vs_epoch}
\end{figure}
The impact of ProcAug is evaluated in an ablation study, in which individual components such as scaling and rotation are selectively removed to assess their contribution, using two metrics: Euclidean Distance ($L_2$) and RCD (\Cref{eq:RCD}). \Cref{tab:l2_chamfer_comparison}  summarizes these results on the test dataset, averaged over 10-fold cross-validation. The results indicate that ProcAug consistently improves reconstruction accuracy, achieving the lowest errors across all latent sizes. While applying only scaling or only rotation individually also leads to improvements over the baseline (No ProcAug), combining both transformations within ProcAug yields the most significant performance gains. Specifically, ProcAug achieves a relative reduction in MSE ranging from $8.46\%$ to $18.90\%$ and in RCD from $6.70\%$ to $15.78\%$, compared to the baseline. This demonstrates that the synergistic effect of jointly applying rotation and scaling enhances the robustness and accuracy of the learned representations more than either transformation alone.

Furthermore, \Cref{fig:recon_vs_epoch} illustrates the evolution of a test reconstruction term (k-folded 10 times) in the loss function \Cref{eq:training_loss} over the training epochs for different latent sizes, comparing ProcAug and NoProcAug. The figure reinforces the findings from \Cref{tab:l2_chamfer_comparison}, showing that ProcAug not only achieves lower final reconstruction error but also accelerates convergence across different latent sizes.

\subsection{Out-of-sample Reconstruction}\label{sec:rec_out}
Similarly to PCA, autoencoder-based methods project data onto a lower-dimensional latent space and then provide the inverse transformation. While the first finds eigenvectors of the covariance matrix to maximize variance in the projected data, the latter uses a nonlinear neural network to learn a latent representation by minimizing a loss function.

\begin{figure}[!htb]
    \centering
    \begin{subfigure}[b]{0.475\textwidth}
        \centering
        \includegraphics[ width=\textwidth]{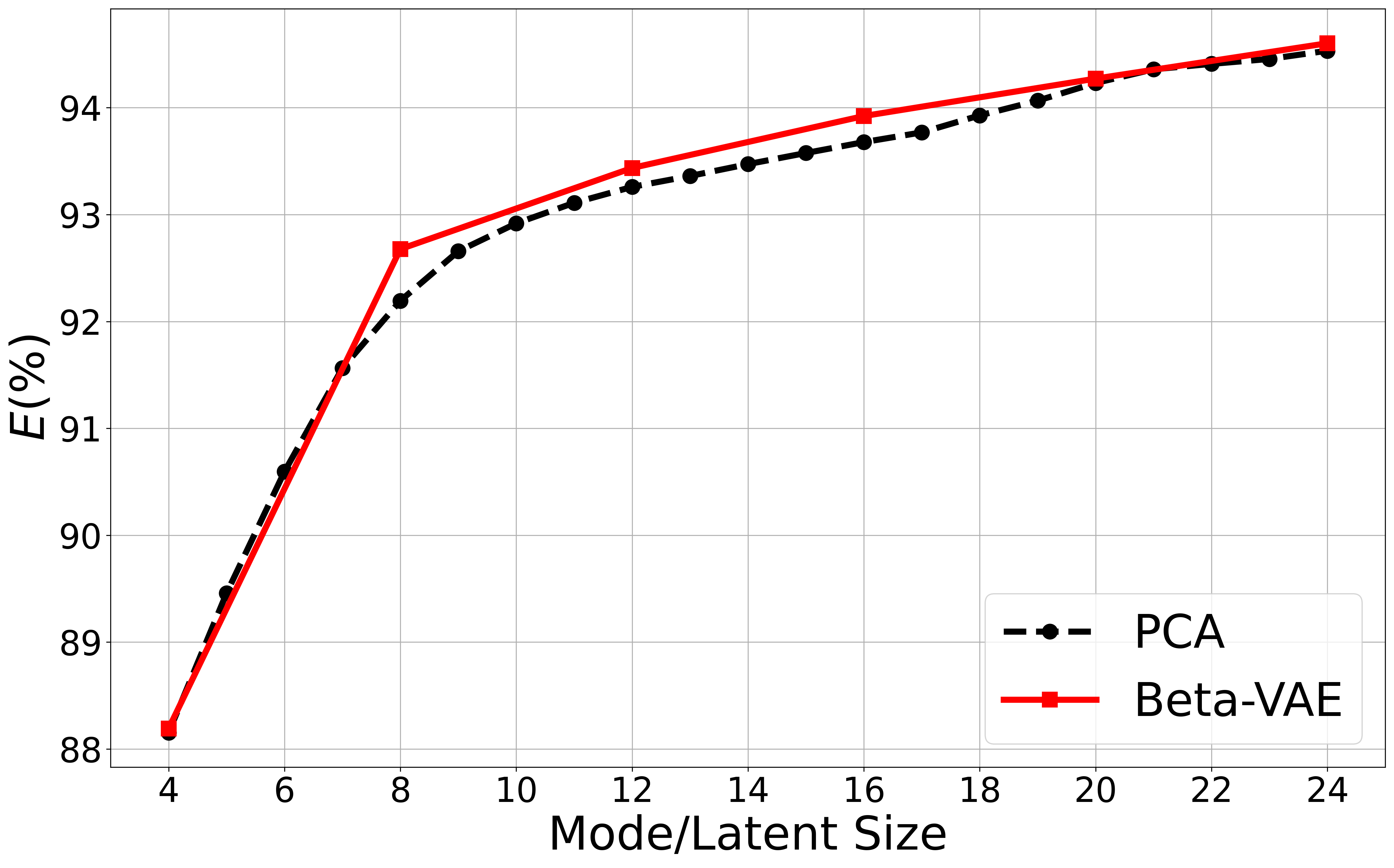} 
        \label{fig:rec_pca_vae_all}
    \end{subfigure}
    \hfill
    \begin{subfigure}[b]{0.475\textwidth}
        \centering
        \includegraphics[width=\textwidth]{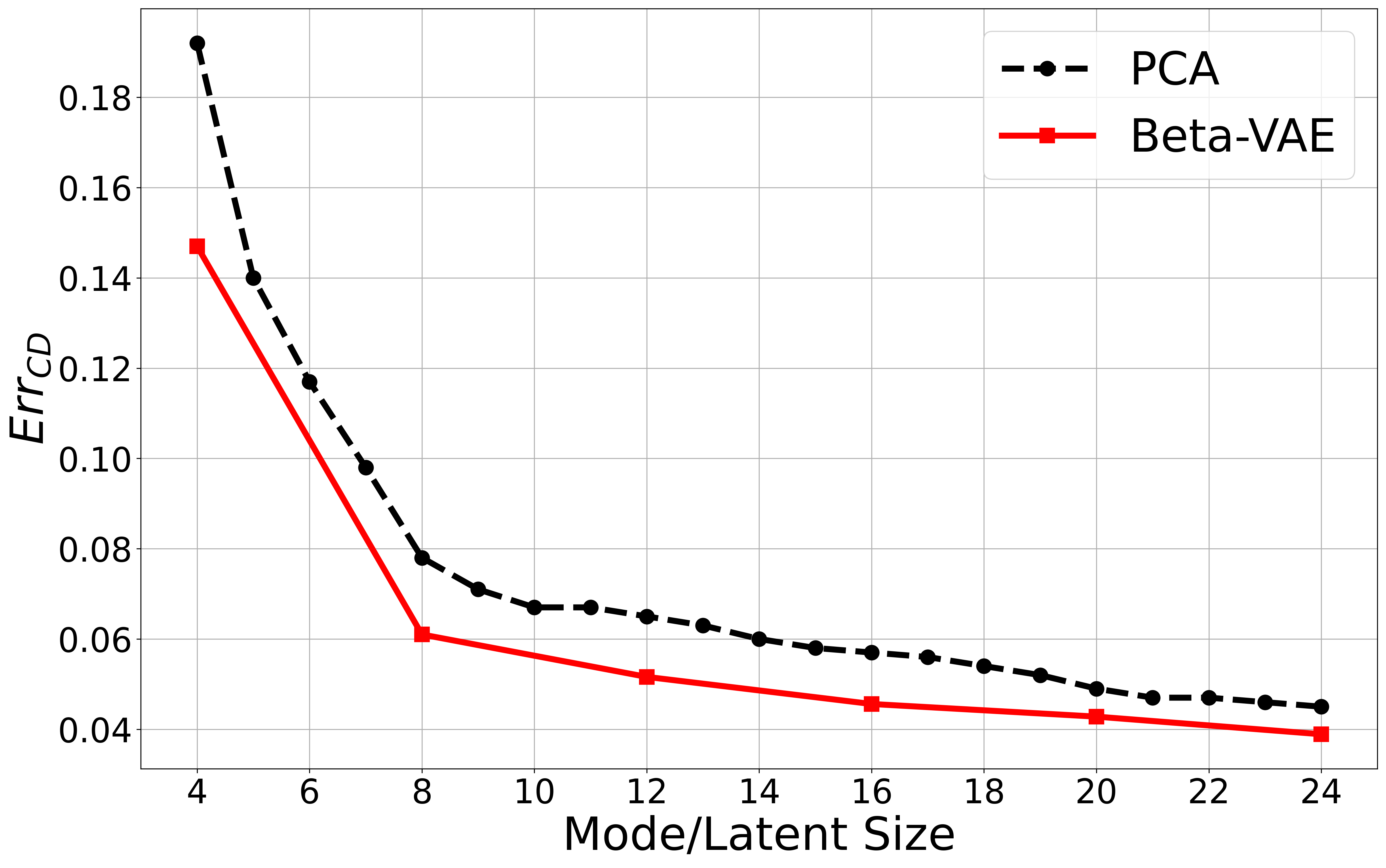}
        \label{fig:figure2}
    \end{subfigure}
    \caption{Behaviour of our trained model with different latent size dimensions vs PCA for $L_2$ reconstruction  \eqref{eq:reconstruction}  (left) and Chamfer error \eqref{eq:chamfer} (right). }
    \label{fig:cd_pca_vae_all}
\label{fig:cd_pca_vae}
\end{figure}

Moreover, PCA learns an orthogonal basis using training data, projecting the test set onto it and then uses the transpose principal components matrix to map back the reduced representation to the original feature space. On the other hand, autoencoder-based model maps the test set onto the feature space using a decoder (not necessarily symmetric to the encoder) block that performs upsampling. To accurately evaluate parallelism, PCA is compared against different $\beta$-VAE models, each trained with varying latent dimension sizes ranging from $d = 4, 8, \dots, 24$. The reconstruction results, considering $\beta=1.0e-3$, are depicted in \Cref{fig:cd_pca_vae}. While the pointwise $L_2$ reconstruction error is comparable to PCA, this metric alone does not fully represent reconstructed shape quality. Minimizing $L_2$ does not necessarily guarantee good shape preservation, particularly for anatomical structures. In fact, two shapes with similar pointwise distances can still significantly differ in overall geometry and topology \cite{wu2021densityawarechamferdistancecomprehensive,montanaro2022rethinkingcompositionalitypointclouds}. The GCN-$\beta$-VAE notably outperforms PCA regarding Chamfer error, a geometry-aware metric, indicating superior shape preservation. This distinction is crucial, as anatomical and functional interpretations rely significantly on accurate geometry. 

\subsubsection{Latent Analysis and Disentangled Representation}\label{sec:dis_lat}
\begin{figure}[!htb]
    \centering
    \begin{subfigure}[t]{0.33\textwidth}
        \centering
        \includegraphics[width=\textwidth]{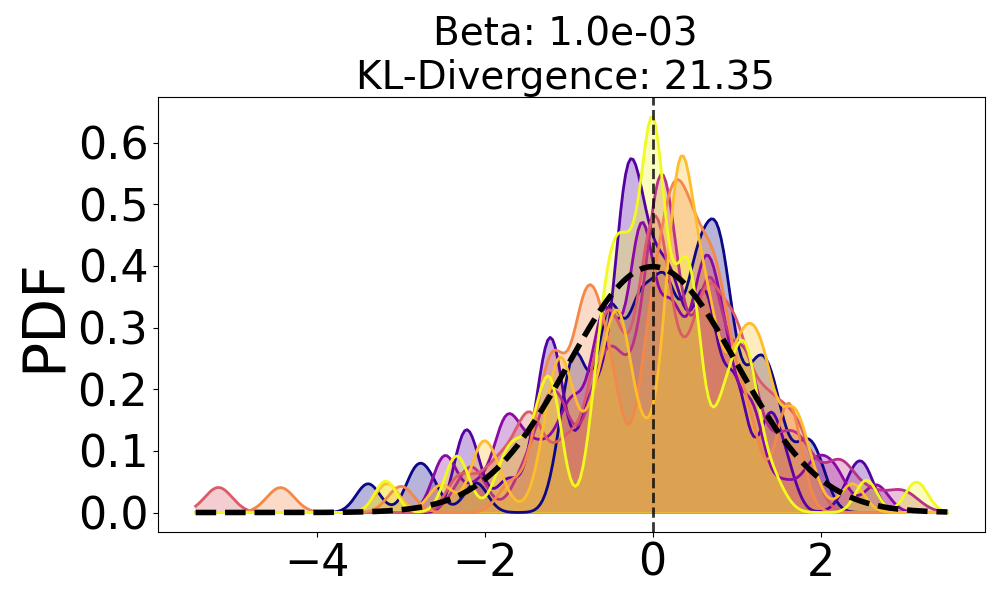}
    \end{subfigure}%
    \begin{subfigure}[t]{0.33\textwidth}
        \centering
        \includegraphics[width=\textwidth]{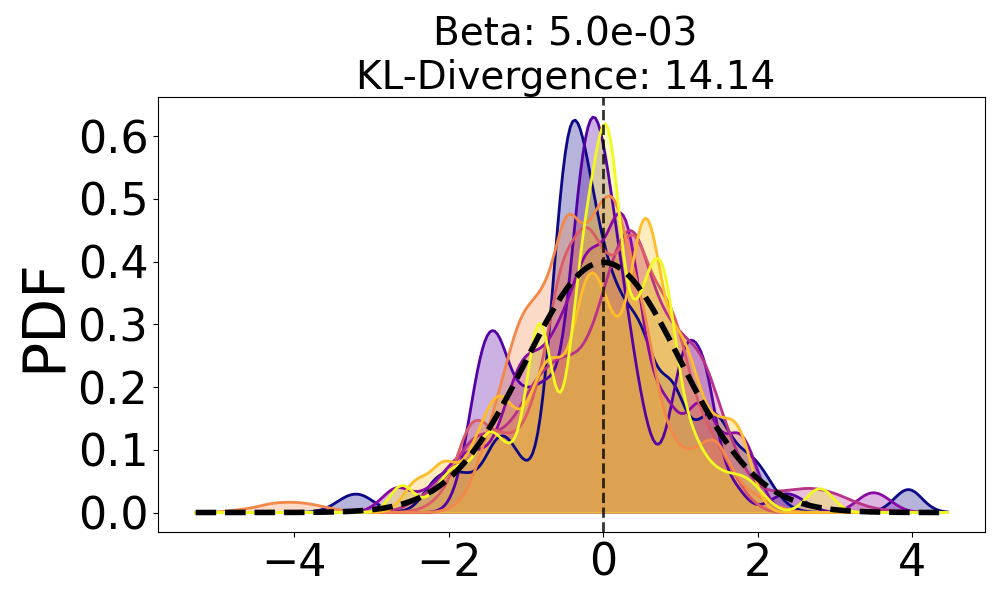}

    \end{subfigure}%
    \begin{subfigure}[t]{0.33\textwidth}
        \centering
        \includegraphics[width=\textwidth]{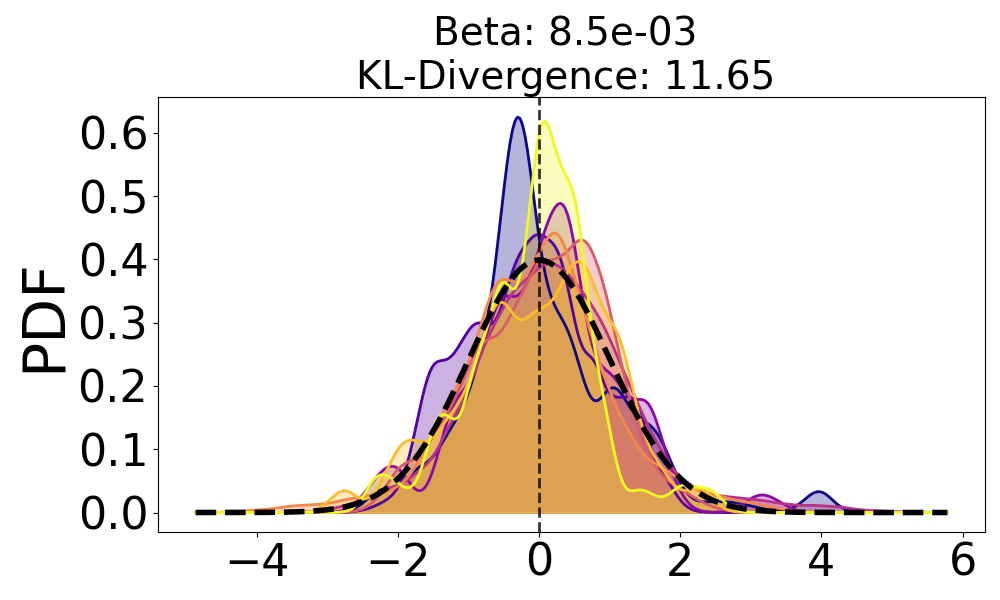}
        
        \label{fig:fig3}
    \end{subfigure}
    
    \caption{Probability Density Functions for each variable in the latent space (colored continuous curves) and Standard Gaussian (dotted curve). As beta increases (left to right), the KL-divergence between the estimated posterior and the prior decreases.}
\label{fig:distribution}
\end{figure}

Unsupervised $\beta$-VAE relies on identifying independent latent representations through the influence of the Lagrangian multiplier $\beta$ on the KL-divergence constraint defined in \Cref{eq:general_beta_loss_function}. By forcing the estimated posterior distribution $q_\phi(z|x)$ to resemble the prior distribution $p_\theta(z) \sim \mathcal{N}(0, I)$, the latent representation is encouraged to approximate a standard multivariate Gaussian characterized by zero mean and identity covariance. This behavior is demonstrated in \Cref{fig:distribution}, where distributions of latent variables progressively approach a standard Gaussian, attaining unit variance and reduced correlation with other variables as $\beta$ increases.

\begin{figure}[!htb]
    \centering
    \begin{subfigure}[t]{0.325\textwidth}
        \centering
        \includegraphics[trim={12cm 0cm 2cm 0cm}, clip, width=\textwidth]{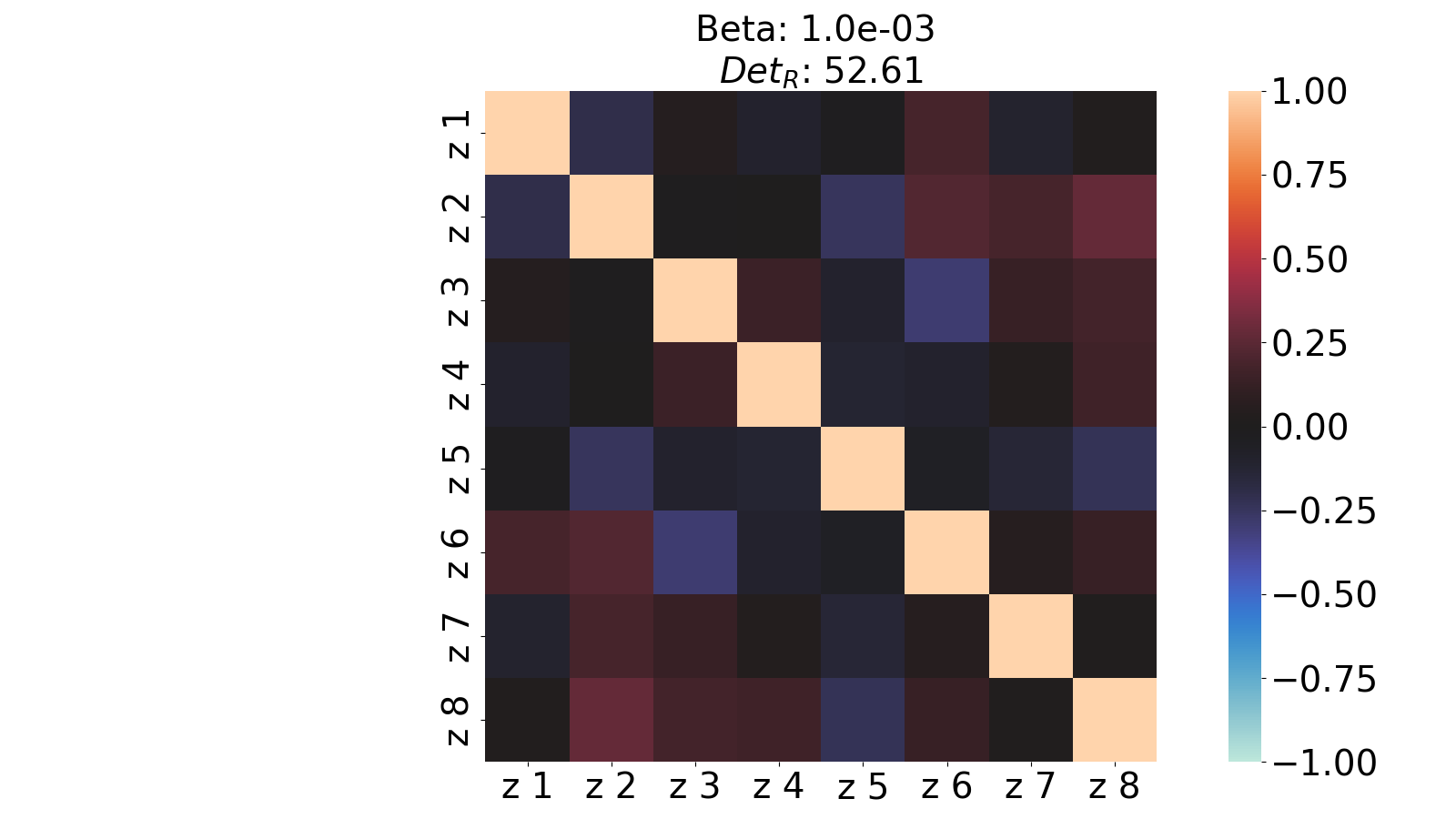}
    \end{subfigure}
    \hfill
    \begin{subfigure}[t]{0.325\textwidth}
        \centering
        \includegraphics[ trim={12cm 0cm 2cm 0cm}, clip, width=\textwidth]{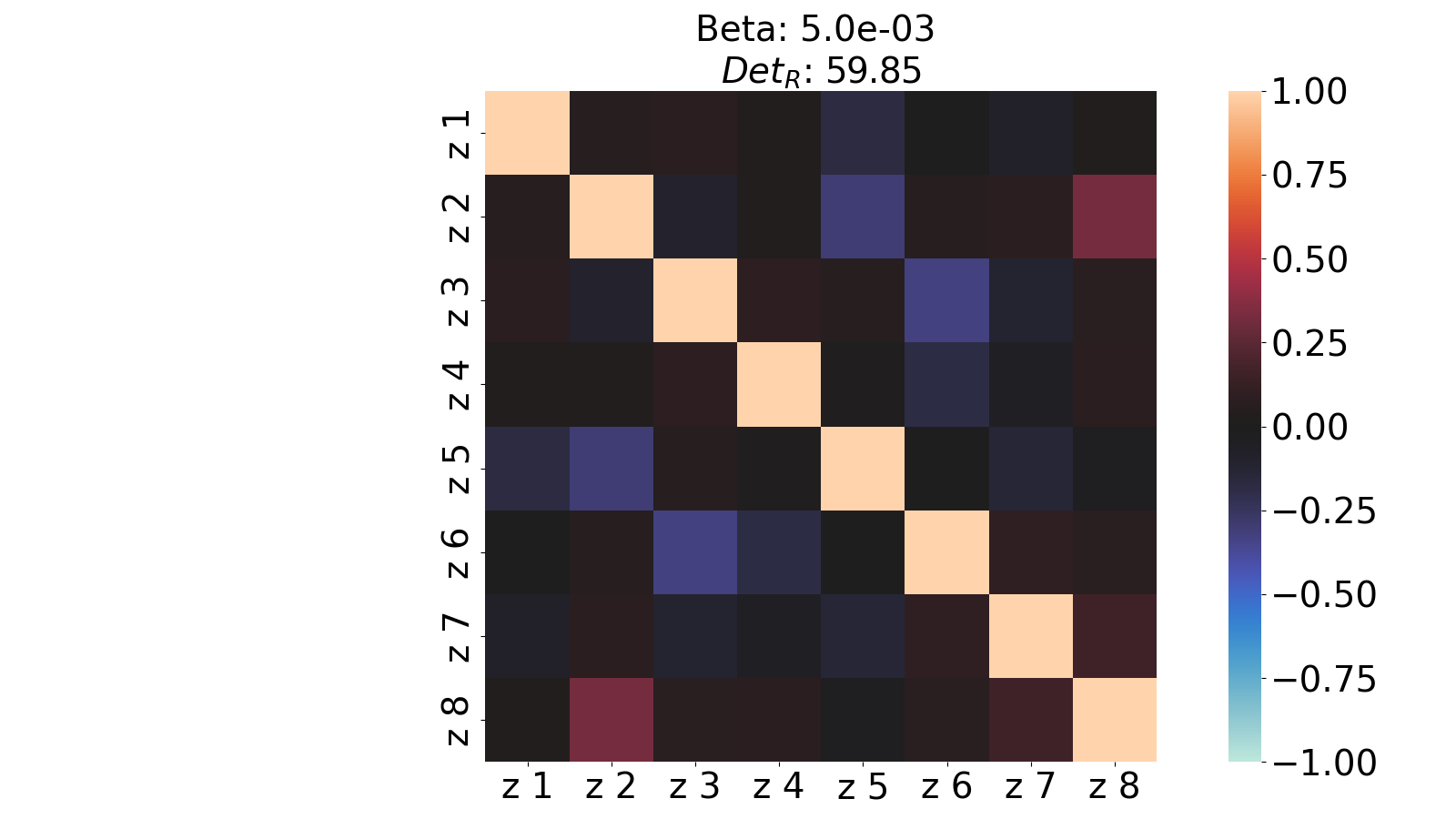}
    \end{subfigure}
    \hfill
    \begin{subfigure}[t]{0.325\textwidth}
        \centering
        \includegraphics[trim={12cm 0cm 2cm 0cm}, clip, width=\textwidth]{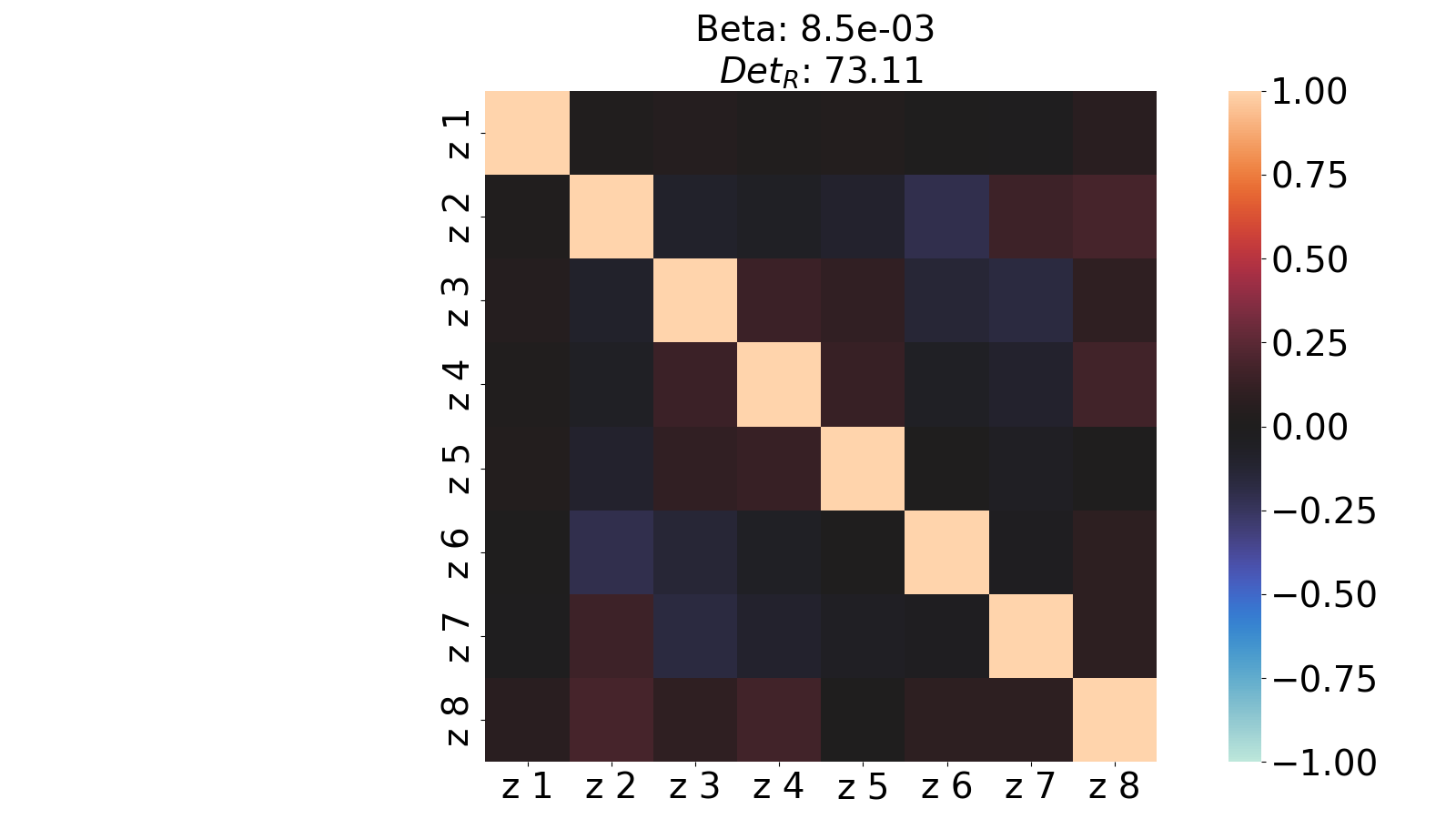}
    \end{subfigure}
\caption{Correlation matrices of the latent variables for different values of beta. The colorbar indicates the value of each element of the matrix. }
\label{fig:Det for betas}
\end{figure}

This trend indicates that the regularization induced by $\beta$ impacts the compactness of the latent space. Specifically, given a fixed latent dimension, only a subset of latent variables is necessary to adequately represent the entire dataset. This concept, termed inner latent compactness, is further explored in \Cref{subsec:hlc}. To evaluate the independence among latent variables across different $\beta$ values, the correlation matrix of these variables is analyzed in \Cref{fig:Det for betas}. Increasing $\beta$ results in a higher determinant of the correlation matrix, pushing it closer to the identity matrix. This effect, recognized for promoting more interpretable latent variables \cite{burgess2018understandingdisentanglingbetavae,kim2019disentanglingfactorising}, slightly compromises reconstruction accuracy.

\begin{figure}[!htb]
    \centering
    \begin{subfigure}[t]{0.48\textwidth}
        \centering
        \includegraphics[ trim={0cm 0cm 0cm 0cm}, clip,width=\textwidth]{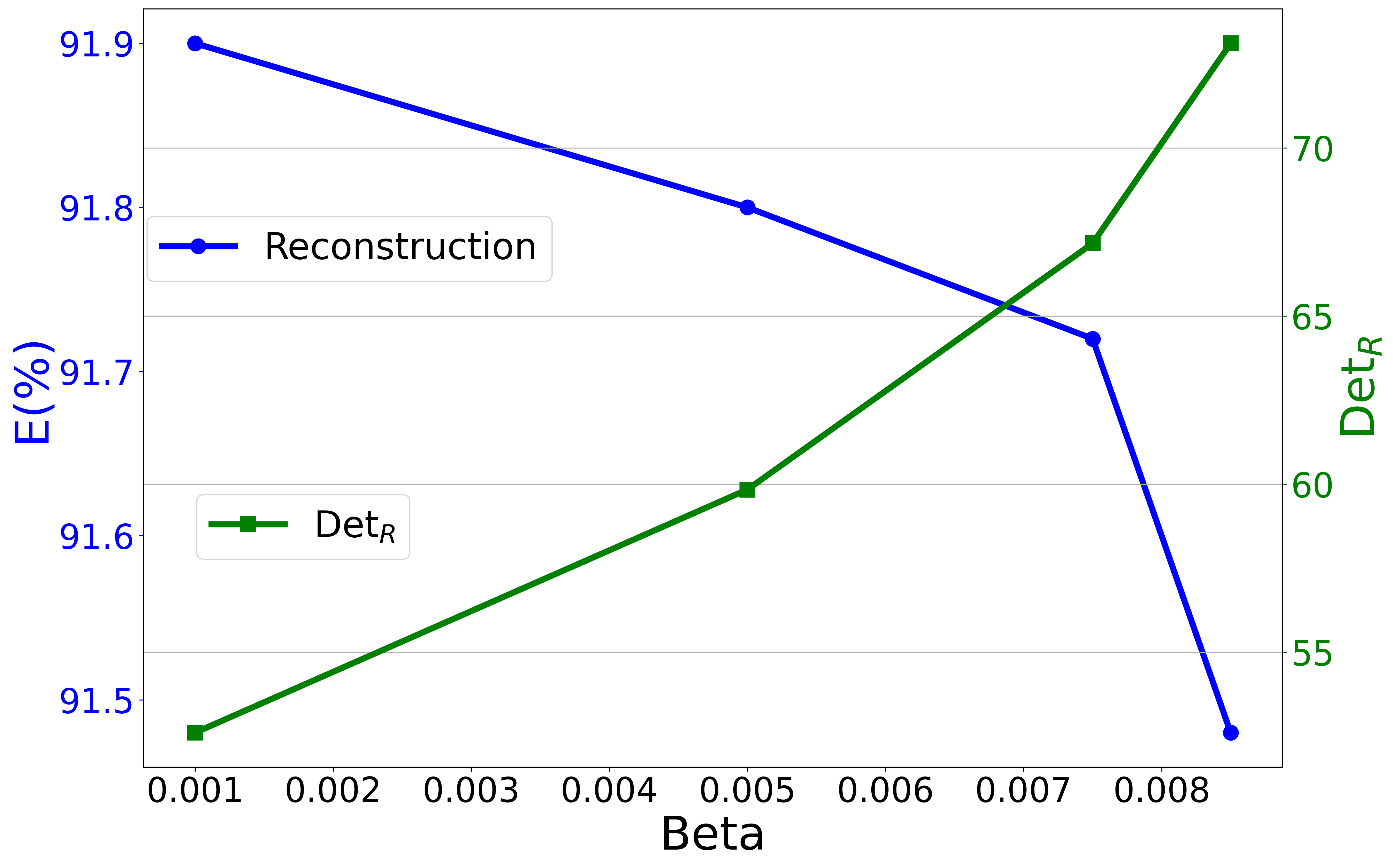} 
    \end{subfigure}
    \hfill
    \begin{subfigure}[t]{0.48\textwidth}
        \centering
        \includegraphics[ trim={0cm 0cm 0cm 0cm}, clip,width=\textwidth]{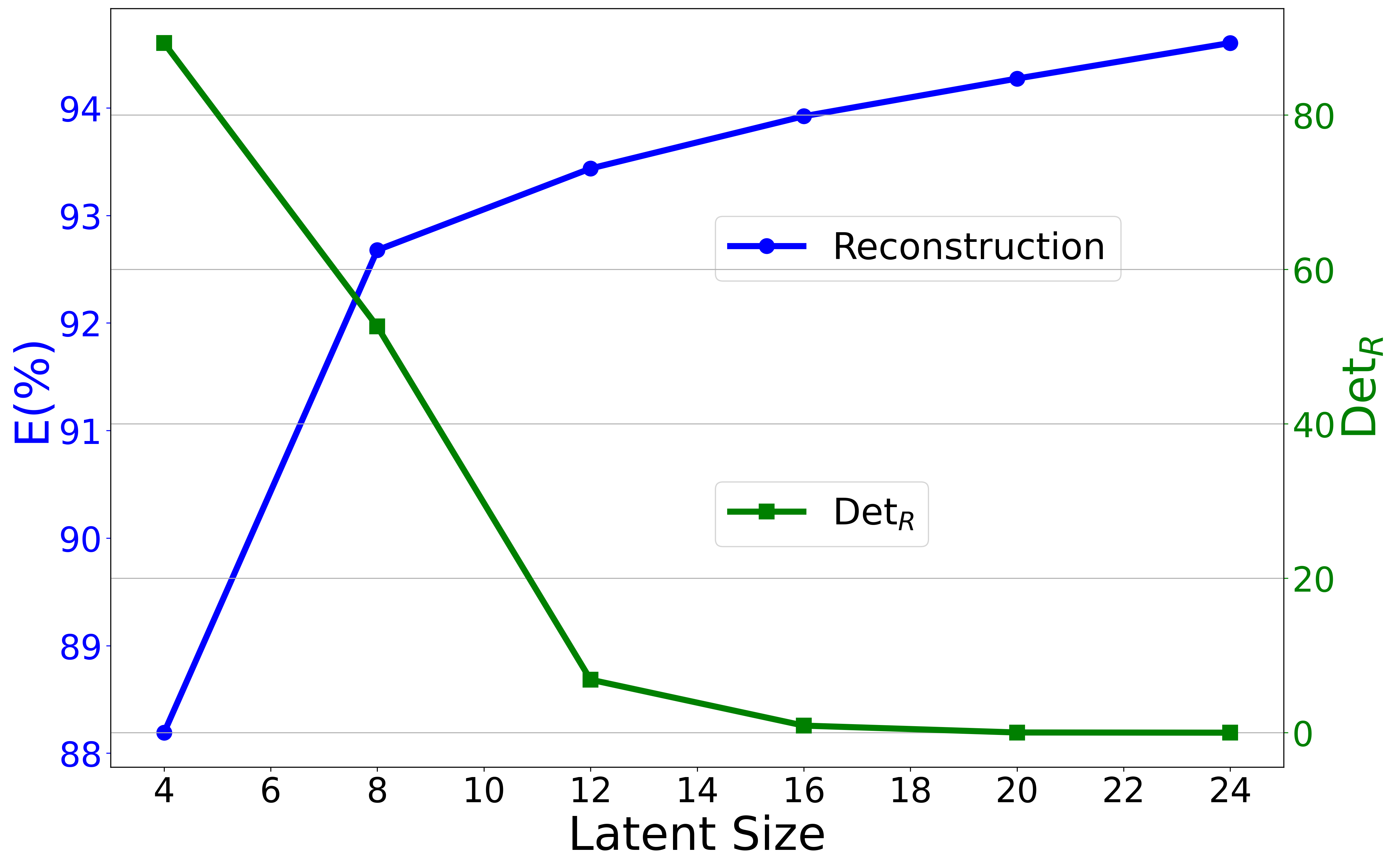}
    \end{subfigure}
\caption{Behavior of the reconstruction percentage (blue curve) and $\det(\mathbf{R})$ (green curve) as a function of beta with a latent dimension of $8$ (left) and as a function of the dimension of the latent space with a $\beta$ of $1.0e-3$ (right).}
\label{fig:rec_de_lat_bet}
\end{figure}

Reconstruction accuracy and correlation matrix determinants across varying $\beta$ values and latent dimensions are presented in \Cref{fig:rec_de_lat_bet}. When increasing $\beta$ from $1.0e-03$ to $8.5e-03$, the determinant of the correlation matrix improves from $\det(\mathbf{R})=52.61$ to $\det(\mathbf{R})=73.11$, while reconstruction percentage marginally declines from $E = 91.90\%$ to $E = 91.48\%$. Nevertheless, this performance consistently surpasses PCA, which achieves a reconstruction percentage of $E = 91.19\%$.


The behavior of the reconstruction percentage and $\det(\mathbf{R})$ is also shown as a function of the latent size. As the latent size increases, reconstruction accuracy improves significantly, ranging from $E = 88.19\%$ at latent size $d = 4$ to $E = 94.61\%$ at $d = 24$. This outcome aligns with expectations, as a larger latent space enables the network to encode more information, reducing loss during the encoding process. However, the increased capacity also introduces the risk of encoding semantically irrelevant details or "noise" rather than focusing on independent statistical factors. This effect is mirrored in the determinant $\det(\mathbf{R})$, which decreases from $89.33$ at $d = 4$ to $0.92$ at $d = 16$. The decline in determinant values for higher-dimensional latent spaces reflects the model’s greater freedom to encode detailed input characteristics, potentially resulting in correlated noise or redundancies instead of purely salient, independent features.

\subsubsection{Hierarchical Inner Latent Contribution}\label{subsec:hlc}

In \Cref{sec:dis_lat}, it is shown that both PCA and a VAE with increasing latent dimensionality can capture similar amounts of variance. Unlike PCA, which naturally ranks its principal components by explained variance, a VAE does not impose an inherent order on its latent dimensions. This distinction is especially relevant in the context of $\beta$-VAEs, where higher $\beta$ values promote a more structured and disentangled latent space by encouraging the posterior to approximate the isotropic prior $\mathcal{N}(0,I)$. In this setting, a systematic approach is required to identify the most informative latent modes for data reconstruction and meaningful variation capture.

The method proposed in \cite{EIVAZI2022117038} addresses this need by introducing a mode-ranking algorithm that yields physically relevant cumulative modes. The approach assumes alignment between the model’s learned reference geometry and the latent space structure, a condition that holds particularly well in $\beta$-VAEs, where latent representations cluster around a central reference vector ${ z_i = 0 }_{i=1}^{d}$

\begin{figure}[!htb]
\centering
\includegraphics[width=0.48\textwidth]{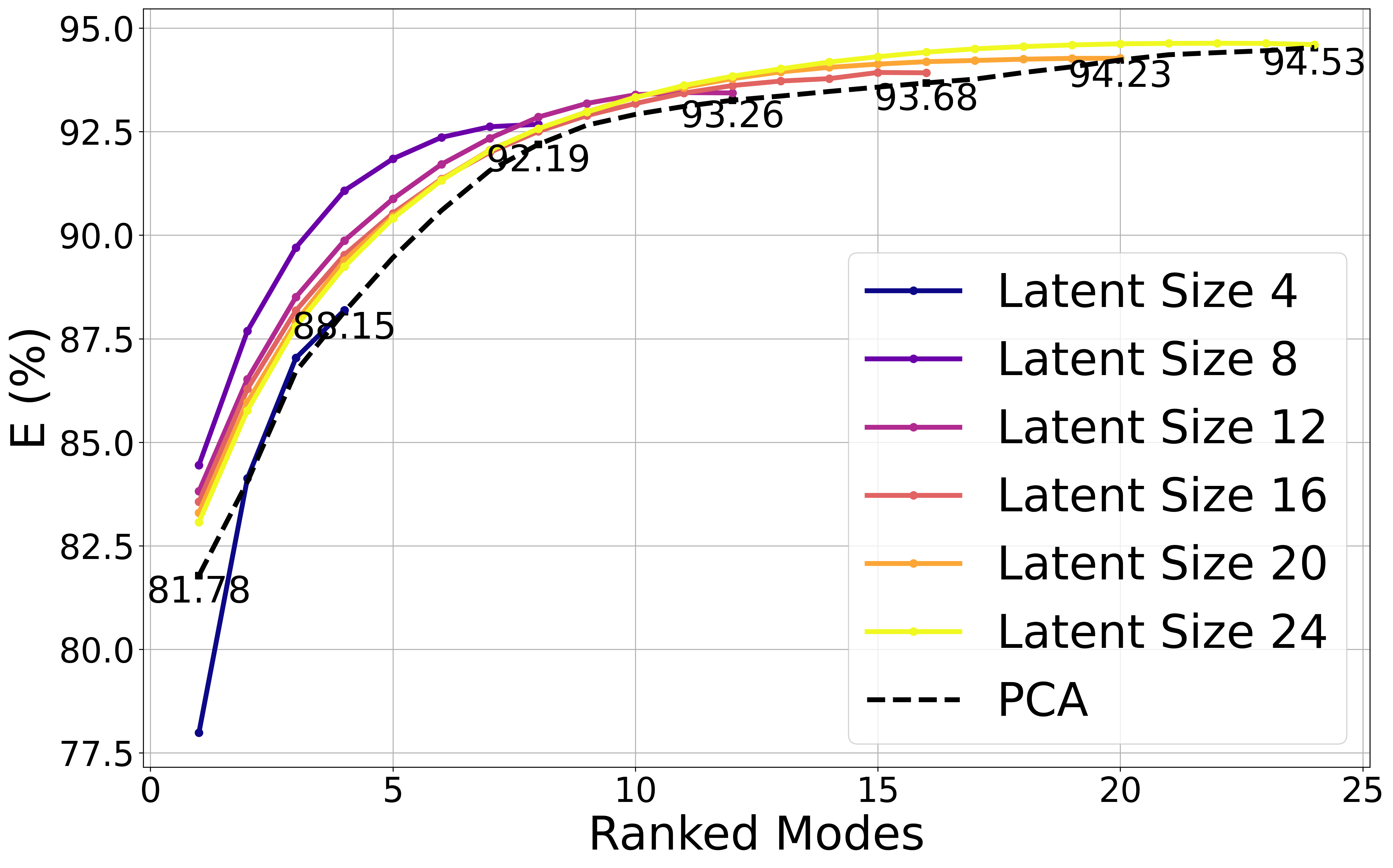}
\hfill
\includegraphics[width=0.48\textwidth]{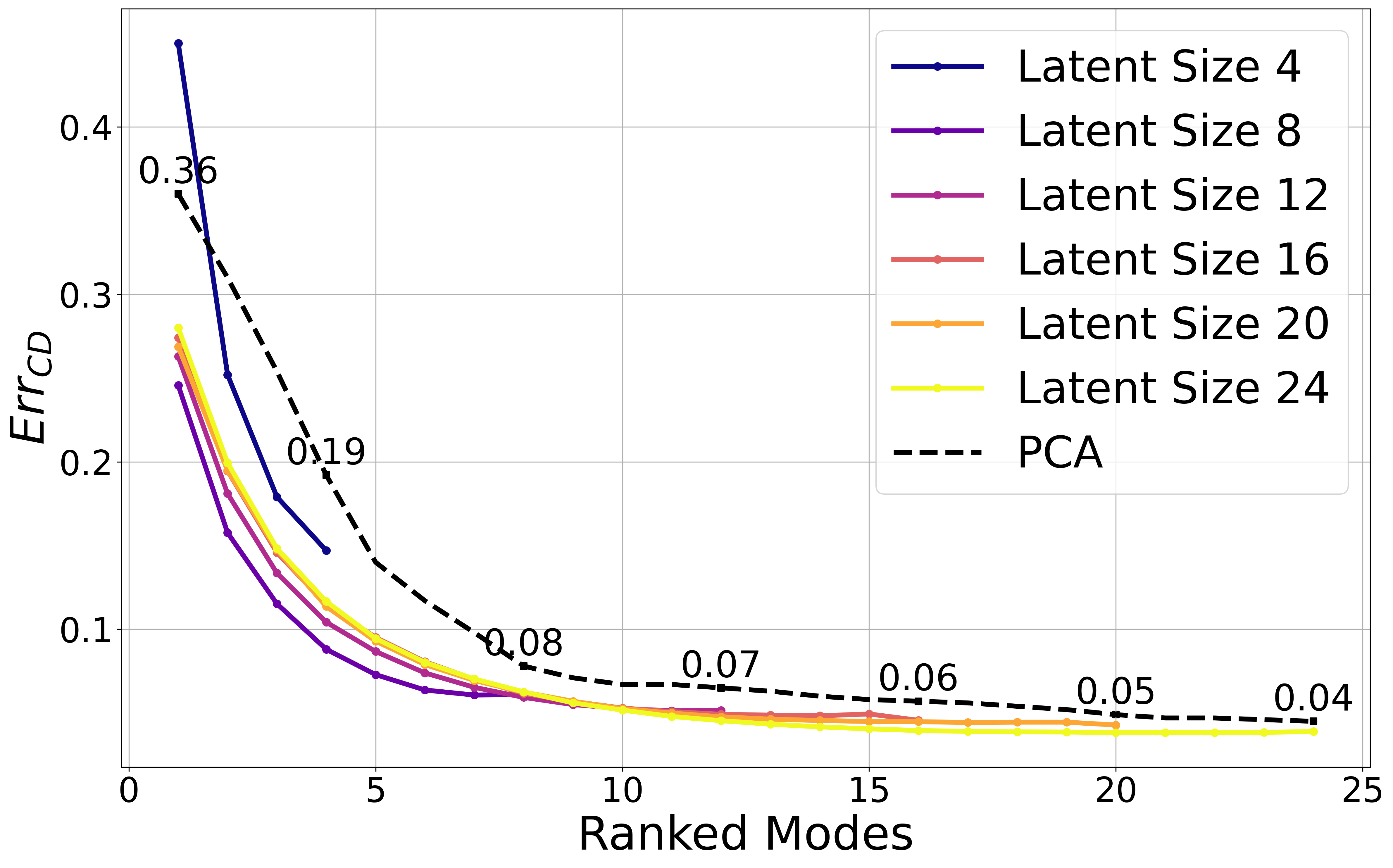}
\caption{Progression of cumulative reconstruction percentage (left) and Chamfer error (right) for latent dimensions ranging from $d=4$ to $d=24$. The curves resemble the typical pattern of cumulative explained variance in PCA. However, aside from the smallest latent space ($d=4$), the GCN-$\beta$-VAE consistently outperforms PCA, achieving higher reconstruction accuracy and lower Chamfer error for each number of modes considered. These results emphasize the effectiveness of the learned mode ranking in improving reconstruction quality.}
\label{fig:all_cum}
\end{figure}

This ranking method is employed in this work (\Cref{alg:ranking_graph_bvae}) to rank latent modes by reconstruction percentage $E(\%)$. A trained GCN-$\beta$-VAE encodes the high-dimensional input data $M$ into its latent representation via the encoder $\mathcal{E} : M \mapsto \boldsymbol{\mu}, \boldsymbol{\sigma}$, using only the mean $\boldsymbol{z} = \hat{\boldsymbol{\mu}}$. To evaluate each mode’s contribution, a modified latent vector is created by retaining only the $i$th component of $\boldsymbol{z}$ and zeroing the rest. This modified vector $\hat{\boldsymbol{z}}_i$ is decoded to produce a partial reconstruction $\hat{M}_i$, and the corresponding energy percentage $E_i(\%)$ quantifies the mode’s influence. The mode yielding the highest energy is selected first. In subsequent steps, previously selected modes are preserved, and each remaining candidate is tested in combination to identify the next highest contributing mode. This iterative process continues until all modes are ranked, producing an ordered list of latent dimensions prioritized by their effectiveness in reconstructing the original data $M$.
\begin{algorithm}[t]
\caption{Ranking GCN-$\beta$-VAE modes}
\label{alg:ranking_graph_bvae}
\begin{algorithmic}[1]
    \Require Trained encoder $\mathcal{E}$ and decoder $\mathcal{D}$ of the CNN-$\beta$VAE; data $\mathbf{x}$
    \Ensure $\mathbf{J}$, the vector containing the ranked mode indices
    \State Initialize $\mathbf{J}$ as an empty vector
    \State $\mathbf{d} \gets$ (vector containing indices of all modes)
    \State $\mu\gets\mu , \sigma \gets \mathcal{E}(\mathbf{M^*})$
    \State $z \gets \mu$\Comment{Use the mean as the representative latent vector}
    \For{$j \in \mathbf{d}$}
        \State Initialize $\mathbf{E}$ as empty
        \For{$i \in \mathbf{d}[\sim \mathbf{J}]$}
            \State $\mathbf{I} \gets [\,i,\, \mathbf{J}\,]$ \Comment{Concatenate $i$ with already-selected modes}
            \State $\hat{\mathbf{z}}_i \gets \mathbf{z}$, then zero out all latent variables \emph{not in} $\mathbf{I}$
            \State $\hat M_i \gets \mathcal{D}\bigl(\hat{\mathbf{z}}_i\bigr)$
            \State Append $E\!\bigl(M, \hat M_i\bigr)$ to $\mathbf{E}$ \Comment{Compute energy/score}
        \EndFor
        \State $\mathbf{J} \gets \mathbf{J} \cup \bigl\{\mathbf{d}[\sim \mathbf{J}][\arg\max(\mathbf{E})]\bigr\}$
    \EndFor
    \State \textbf{Output:} $\mathbf{J}$, the ranked list of modes
\end{algorithmic}
\end{algorithm}
To illustrate how mode ranking affects reconstruction, the cumulative reconstruction percentage and the cumulative Chamfer error are evaluated across different latent space sizes in \Cref{fig:all_cum}. Specifically, dimensions ranging from $d = 4$ to $d = 24$ are analyzed. The resulting curves display the typical pattern of cumulative explained variance observed in PCA, indicating that the latent space captures increasingly more variance as dimensionality grows.

\begin{figure}[!htb]
\centering
\includegraphics[width=0.48\textwidth]{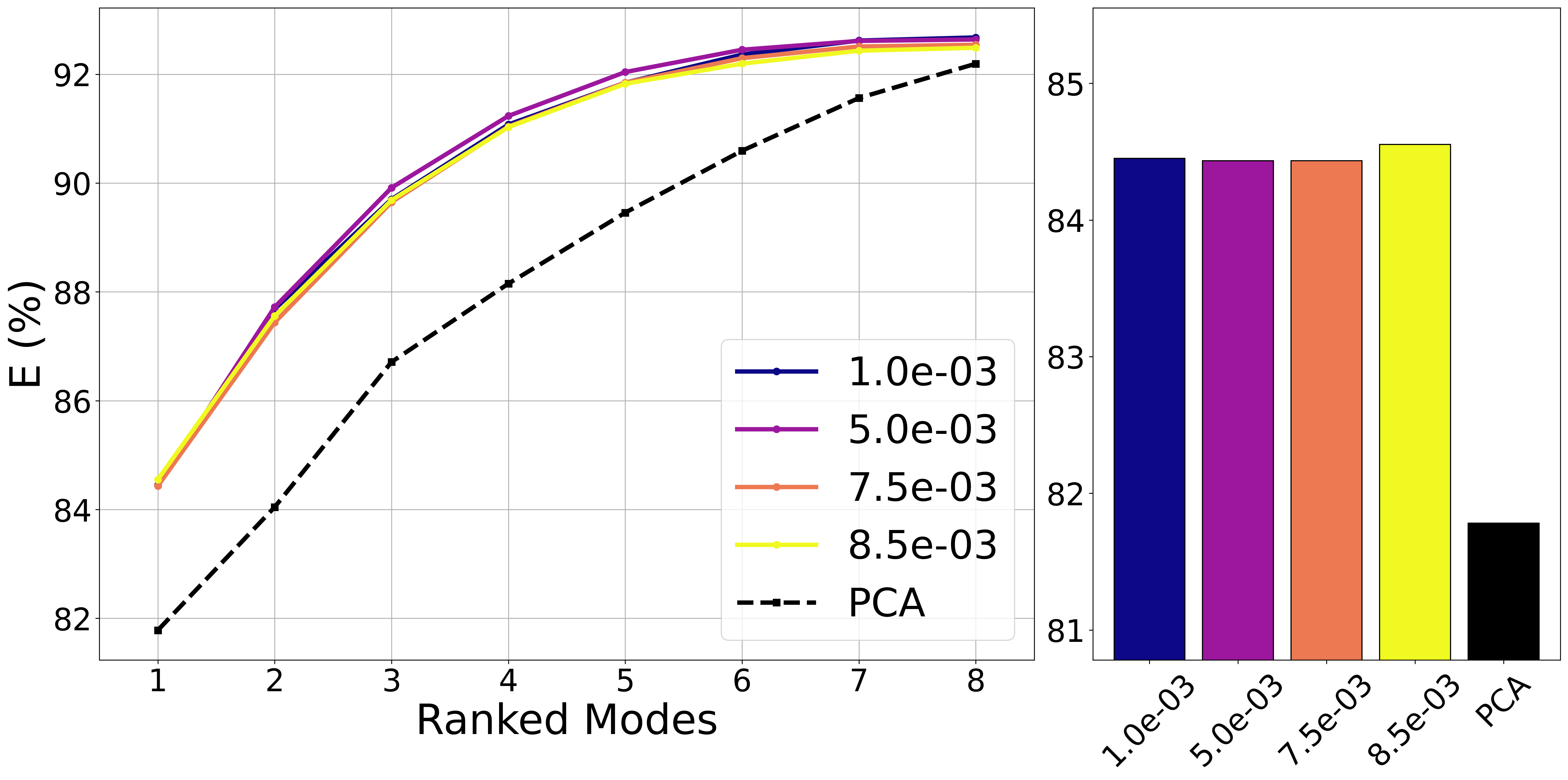}
\hfill
\includegraphics[width=0.48\textwidth]{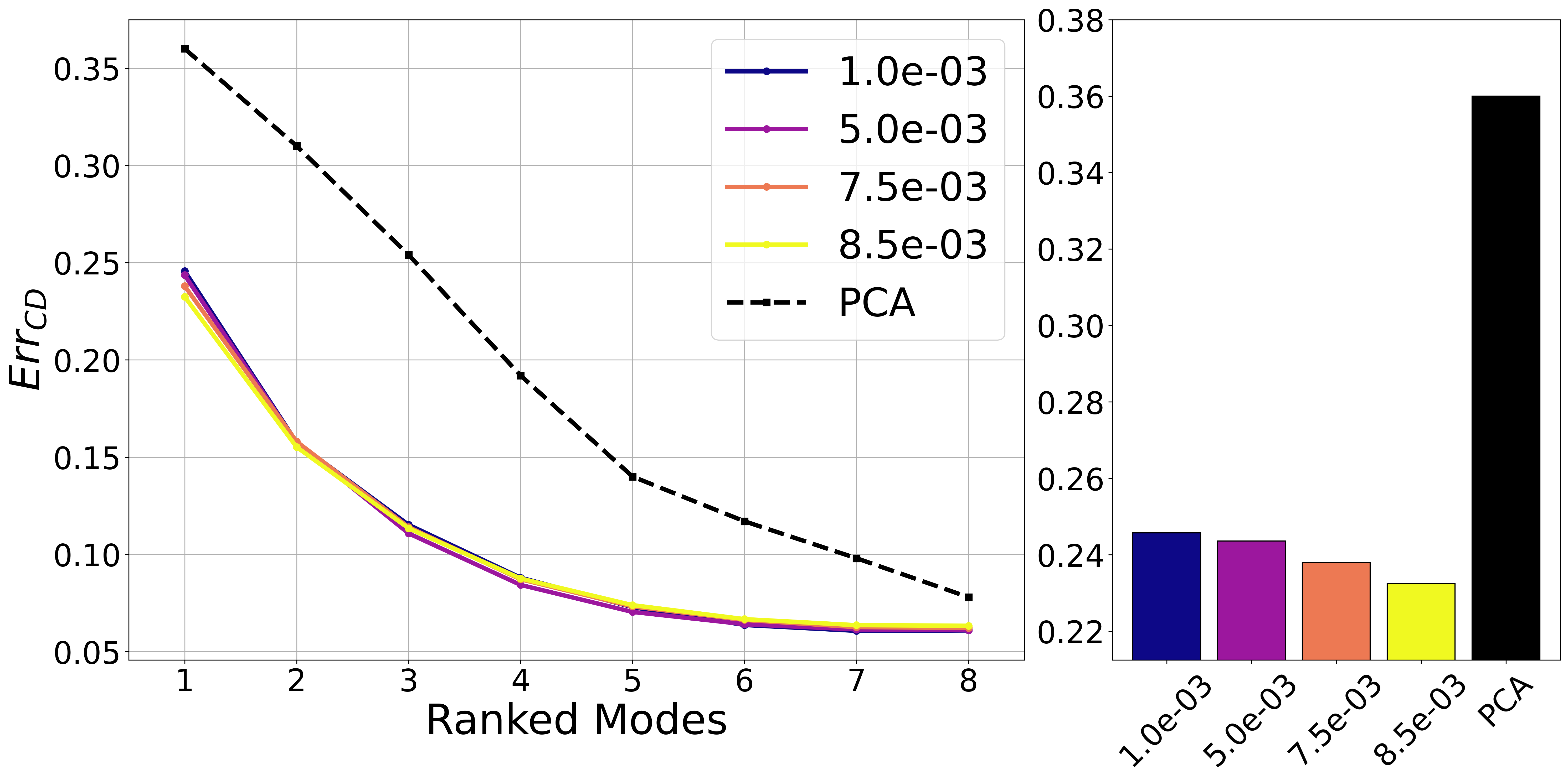} 
\caption{Evolution of latent compactness as $\beta$ increases, shown for both Chamfer error (right) and reconstruction percentage (left). The left panels depict the ranked latent modes as $\beta$ varies from $1.0e-3$ to $8.5e-3$, while the right panels show the value of the first latent mode over the same range. As $\beta$ increases,  the reconstruction percentage $E$ in the first mode slightly rises while the Chamfer error $E_{rrCD}$ decreases, indicating that a stronger KL regularization term promotes a tighter, more structured latent space.}
\label{fig:betas_cum}
\end{figure}
\begin{figure}[!htb]
    \centering
    \begin{subfigure}{0.95\textwidth}
    \centering
        \phantomsubcaption\label{fig:hier_a}
        \tikz\node[inner sep=0pt,label={[anchor=north west]north west:PCA}]{
            \includegraphics[width=0.85\textwidth, trim={0cm 12.5cm 1.85cm 12cm}, clip]{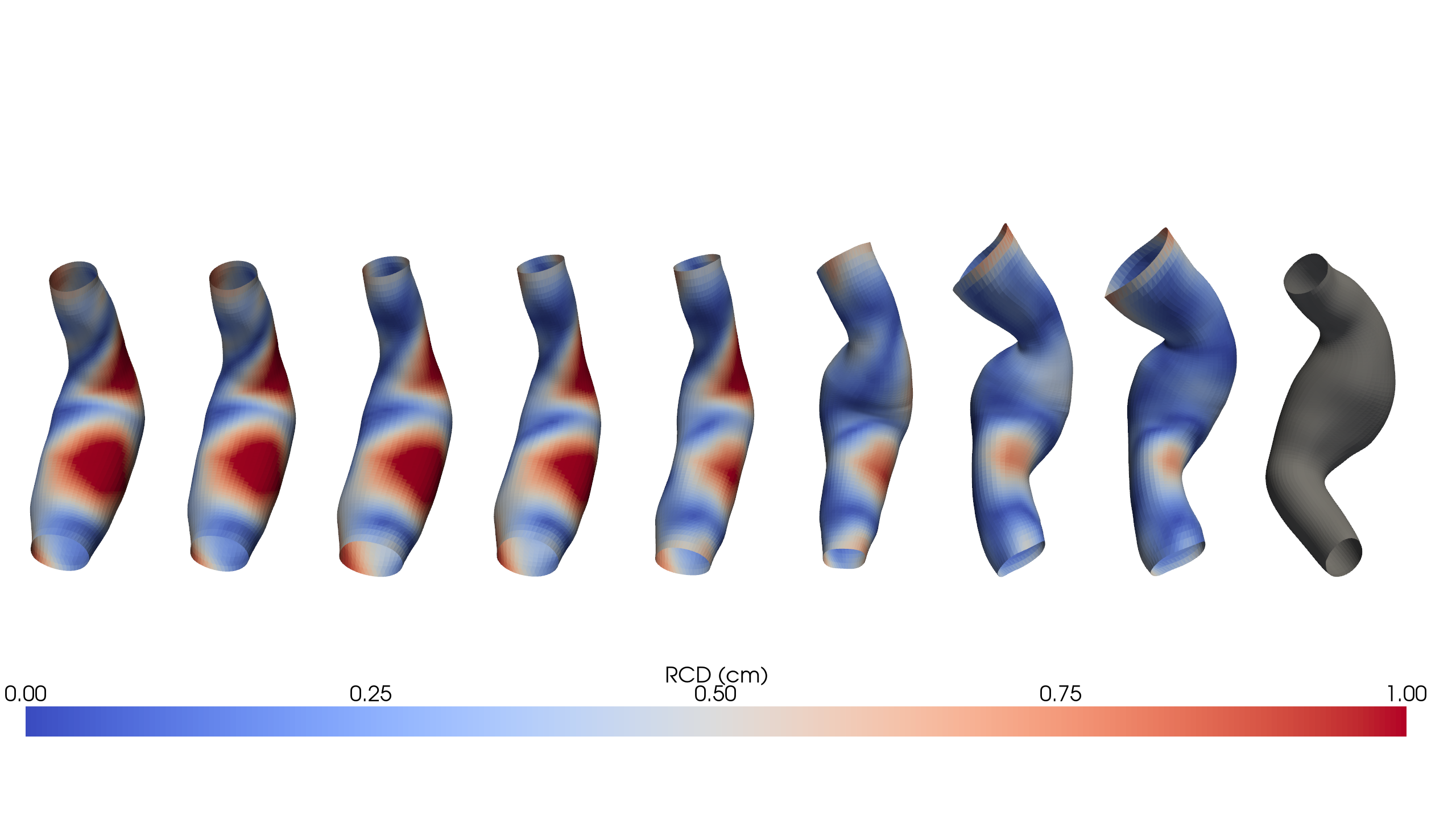}
        };
    \end{subfigure}
    \begin{subfigure}{0.95\textwidth}
    \centering
        \phantomsubcaption\label{fig:hier_b}
        \tikz\node[inner sep=0pt,label={[anchor=north west]north west:$\beta=1.0e{-3}$}]{
            \includegraphics[width=0.85\textwidth, trim={0cm 12.5cm 1.85cm 12cm}, clip]{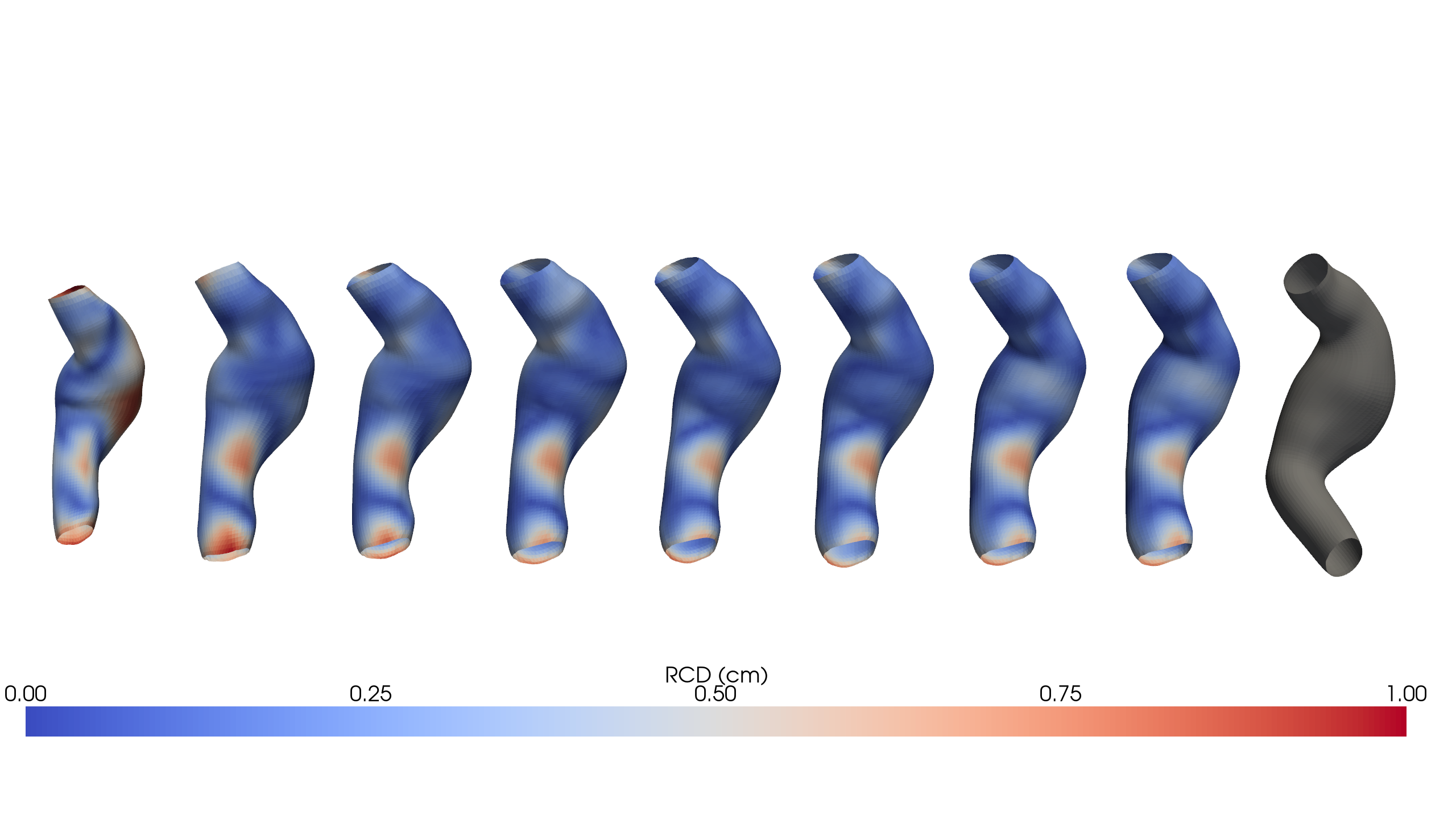}
        };
    \end{subfigure}
    \begin{subfigure}{0.95\textwidth}
    \centering
        \phantomsubcaption\label{fig:hier_c}
        \tikz\node[inner sep=0pt,label={[anchor=north west]north west:$\beta=8.5e{-3}$}] {
            \includegraphics[width=0.85\textwidth, trim={0cm 4.5cm 1.85cm 12cm}, clip]{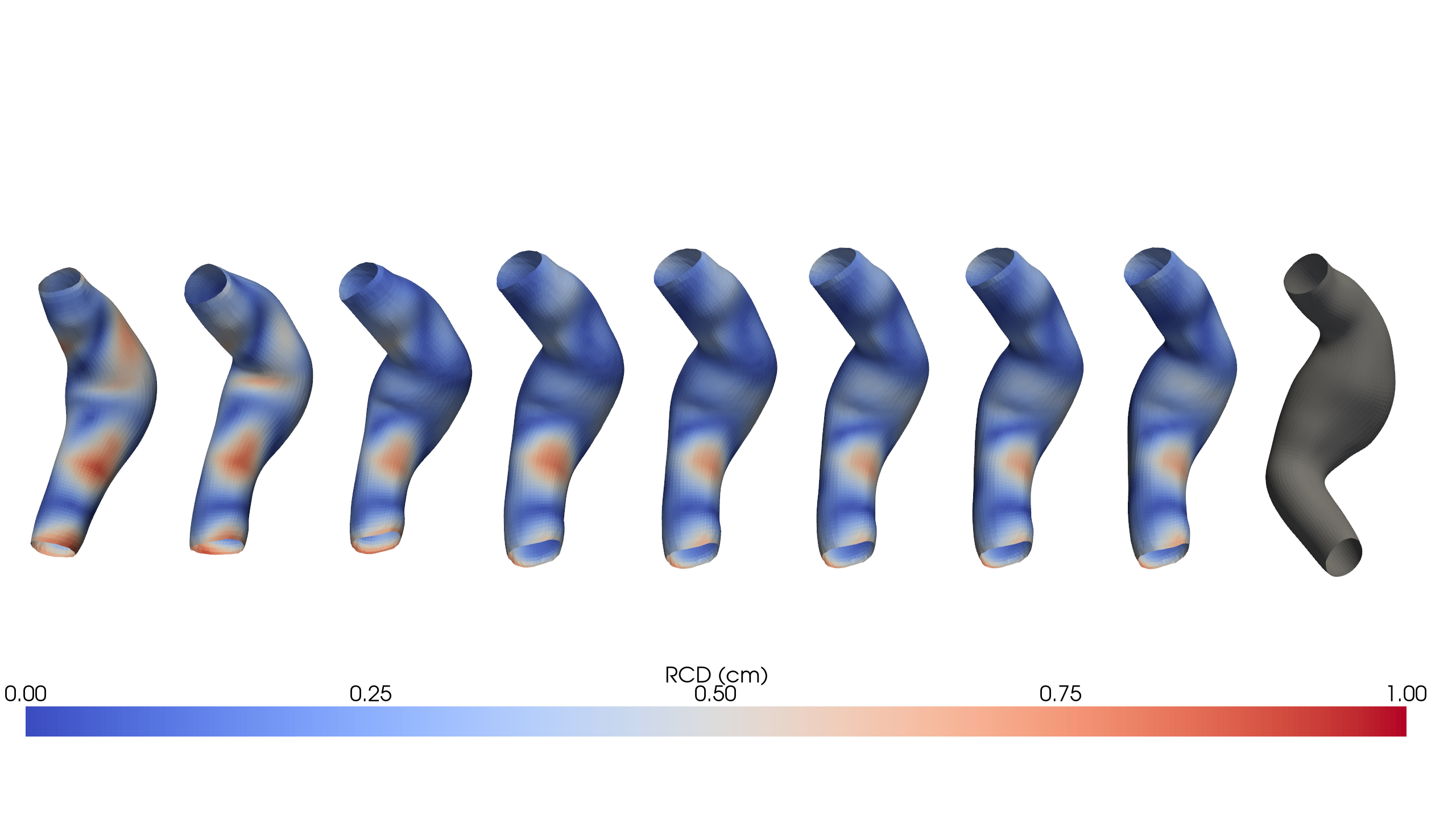}
        };
    \end{subfigure}
    \caption{Ranked mode reconstructions from PCA (top row), GCN-$\beta$-VAE with $\beta = 1.0e-3$ (middle row), and
GCN-$\beta$-VAE with $\beta = 8.5e - 3$ (bottom row). Each column shows the reconstructed shape from a single latent
mode (or principal component), progressing from rank 1 to rank 8. Vertex wise color coding is based RCD(cm). The first modes of the GCN-$\beta$-VAE yield more compact deformations than PCA, which
struggles especially in the higher order modes (far right), most notably along the boundary regions.}
    \label{fig:hier}
\end{figure}

Notably, except for the latent size 4, the curves remain consistently above the reference PCA curve for $E$ and below for what concerns the chamfer error, which signifies that for the same number of modes, the VAE achieves superior reconstruction quality compared to PCA. This discrepancy is most pronounced in the first mode, namely the vector in which only the element at position $j$ is nonzero; that is, $\{ z_i = 0 \}_{i=1, i \neq j}^{d} \cup \{ z_j\neq 0 \}$, indicating that a latent code near the reference geometry can achieve substantial reconstruction accuracy and shape fidelity compared to PCA. This observation suggests a form of inner compactness within the model’s learned latent space.
Furthermore, in \Cref{fig:betas_cum}, it is examined how this compactness evolves as the weight $\beta$ on the KL-Divergence term increases. As expected, as $\beta$ increases, the reconstruction percentage $E$ rises, while the Chamfer error $E_{rrCD}$ decreases. Increasing $\beta$ enforces a tighter alignment between the posterior and the standard Gaussian prior, effectively compressing and regularizing the latent space. These results reinforce the notion that stronger regularization via $\beta$ improves compactness and structure within the latent space. Finally, \Cref{fig:hier} Figure 11 illustrates the reconstruction error, represented by the RCD values encoded in the mesh colors for one particular sample of the test set. While PCA achieves a reduction in the average integral distance as the number of modes increases, it fails to recover fine morphological details of the geometry, even when more modes are used. In contrast, the VAE not only preserves local geometric features with fewer modes but also progressively enhances the local accuracy of the reconstruction as additional modes are incorporated. With $\beta = 1\text{e}{-3}$, the first mode alone yields an average error of 0.41\,cm, which rapidly decreases to around 0.20\,cm by the seventh mode. By contrast, PCA starts at 0.42\,cm and remains above 0.3\,cm until the sixth mode, ultimately dropping to 0.23\,cm by the eighth mode. Increasing $\beta$ to $8.5\text{e}{-3}$ improves the VAE’s initial mode (0.36\,cm). Reconstruction via PCA exhibits pronounced artefacts, particularly in the neck and boundary regions, with these issues persisting even in the cumulative higher-order modes. In contrast, both GCN-$\beta$-VAEs provide more spatially compact and accurate reconstructions in early modes, with the first latent direction yielding a deformation significantly closer to the ground truth than PCA. This effect becomes more pronounced with increasing $\beta$, supporting the notion that stronger regularization induces greater compactness in the latent representation. Altogether, these findings demonstrate that heavier regularization (larger $\beta$) fosters a more compact and practical latent-space representation than PCA and lower-$\beta$ settings.

\subsection{In-sample Generative Experiments}\label{sec:gen}
This section outlines the experimental procedure used to generate AAA shapes with the GCN-$\beta$-VAE model. In \Cref{sec:rec_out}, the focus was on evaluating the generalization capability of the GCN-$\beta$-VAE on unseen data. To this aim, both the regularization weight $\beta$ and the latent dimension $d$ are tuned to promote a latent space characterized by maximally independent factors. This independence is assessed through the analysis of the correlation matrix $\mathbf{R}$ of the latent variables, and the model configuration that maximizes $\det(\mathbf{R})$ indicating minimal correlation among latent dimensions is selected, provided that reconstruction accuracy remains adequate.
With the optimal configuration identified as $\beta = 8.5 \times 10^{-3}$ and latent dimension $d = 8$, this model is subsequently applied to in-sample generative tasks. Specifically, in-sample generation is performed by either smoothly interpolating between two latent codes from the training dataset or by sampling from a Gaussian distribution centered around the latent code of a selected training sample. The interpolation procedure facilitates the generation of intermediate structural transitions, whereas Gaussian perturbations enable the modeling of localized variability surrounding instances.
These experiments demonstrate the learned latent space capacity to synthesize new samples closely aligned with the training distribution, reinforcing the GCN-$\beta$-VAE's effectiveness in capturing and encoding meaningful latent structures within the training set. For generative purposes, only the mean of the latent distribution is used, such that $z_i = \mu_i$ instead of $z_i = \mu_i + \sigma_i \odot \epsilon$. The stochastic term is omitted as its primary function is to impose regularization during training rather than to influence the generation process.
\subsubsection{Extrapolation Through Gaussian Noise}
\label{sec:ext}
\begin{figure}[!htb]
    \centering
    \begin{adjustbox}{width=0.85\textwidth}
        \begin{NiceTabular}{cc}
        \Block[borders={right,tikz={densely dashed, thick}}]{3-1}{
            \begin{tikzpicture}
                \node[inner sep=0pt] (img) at (0,0) {
                    \includegraphics[trim={12cm 1cm 12cm 1cm},clip,width=0.15\textwidth]{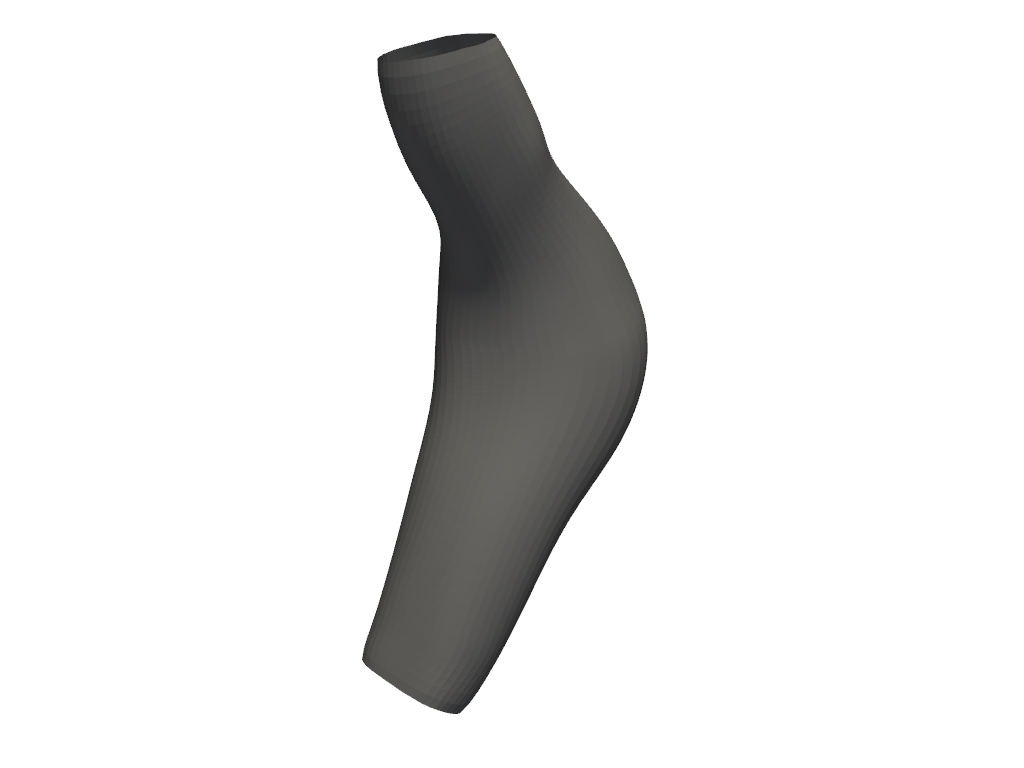}
                };
                \phantomsubcaption\label{fig:2a}
               
                \node[anchor=south west, yshift=-1cm] at (img.south west) {\centering \Large Ground Truth};
            \end{tikzpicture}
        }
        &
        \begin{tikzpicture}
            \node[inner sep=0pt] (img) {\includegraphics[trim={3cm 12cm 3cm 7.6cm},clip,width=0.85\textwidth]{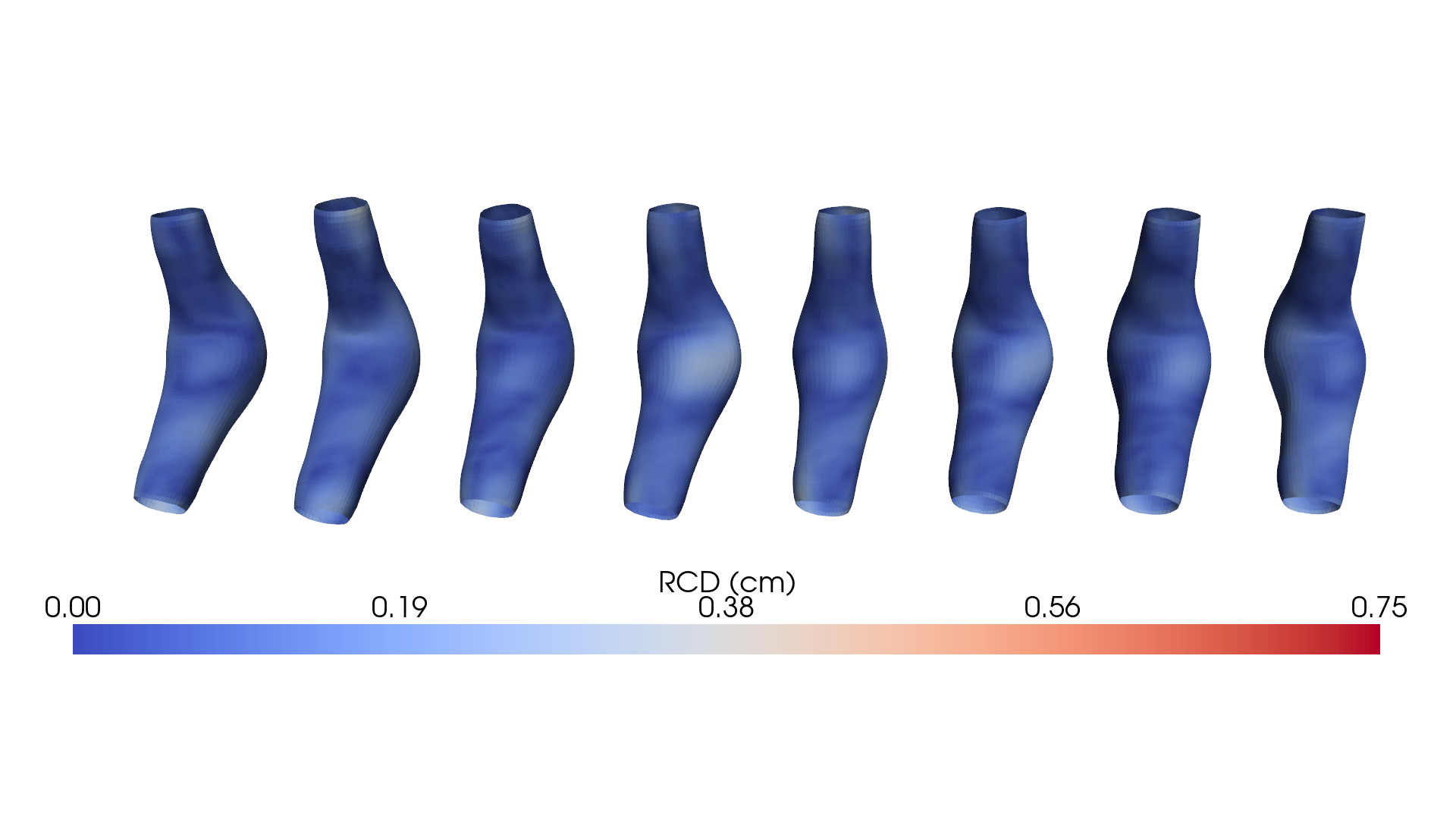}};
            \phantomsubcaption\label{fig:2b}
            \node[anchor=north west] at (img.north west) {\Large\textbf{$S=0.25$}};
        \end{tikzpicture}
        \\
        &
        \begin{tikzpicture}
            \node[inner sep=0pt] (img) {\includegraphics[trim={3cm 12cm 3cm 7cm},clip,width=0.85\textwidth]{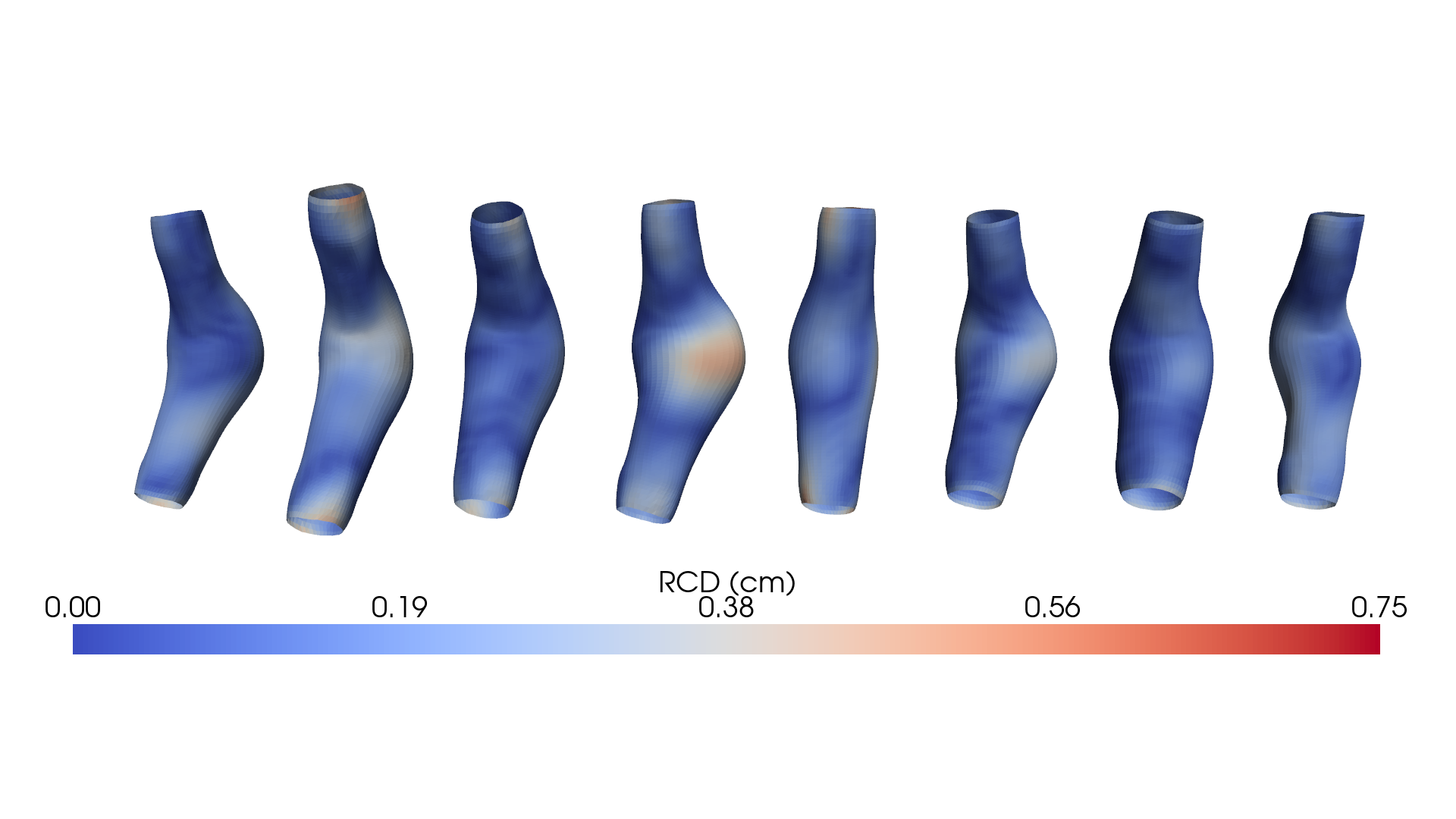}};
            \phantomsubcaption\label{fig:2c}
            \node[anchor=north west] at (img.north west) {\Large\textbf{$S=0.5$}};
        \end{tikzpicture}
        \\
        &
        \begin{tikzpicture}
            \node[inner sep=0pt] (img) {\includegraphics[trim={3cm 7cm 3cm 7cm},clip,width=0.85\textwidth]{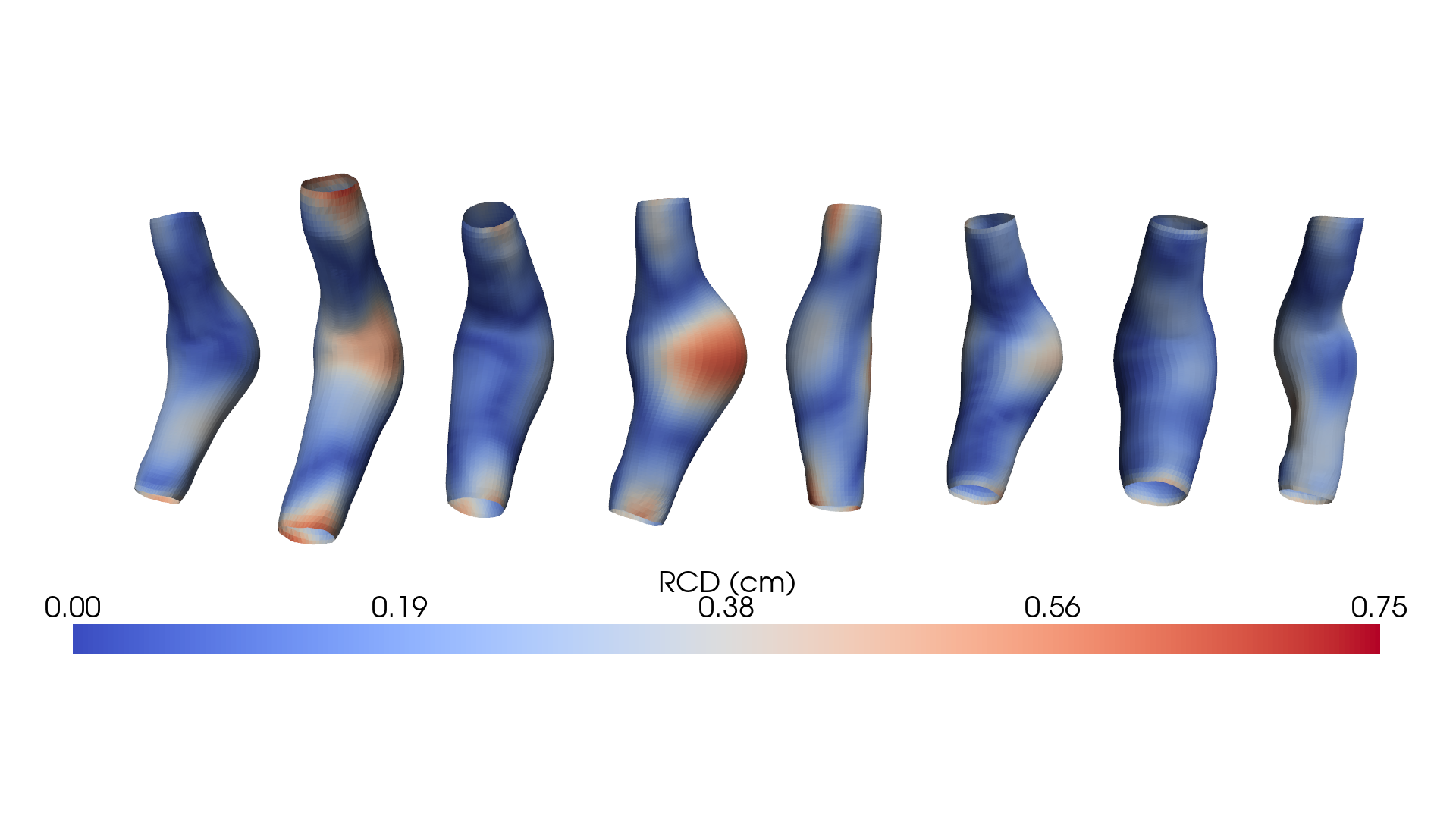}};
            \phantomsubcaption\label{fig:2d}
            \node[anchor=north west] at (img.north west) {\Large\textbf{$S=0.75$}};
        \end{tikzpicture}
        \\
        \end{NiceTabular}
    \end{adjustbox}

    \caption{Ground-truth aneurysm mesh (leftmost column) and RCD visualizations for eight extrapolated meshes (right columns) generated with increasing levels of latent perturbation magnitude $S$. Rows correspond to $S = 0.25$, $S = 0.5$, and $S = 0.75$, from top to bottom. The RCD values are visualized as color maps over the surface of each extrapolated mesh, indicating the degree of geometric deviation from the ground truth. As $S$ increases, the deviations become more pronounced, with notable examples including aneurysm sac expansion (especially evident in columns 4 and 6) and axial elongation (highlighted in column 2). Despite these changes, the global anatomical integrity of the abdominal aortic aneurysm (AAA) is preserved across all levels of perturbation.}
\label{fig:extr}
\end{figure}
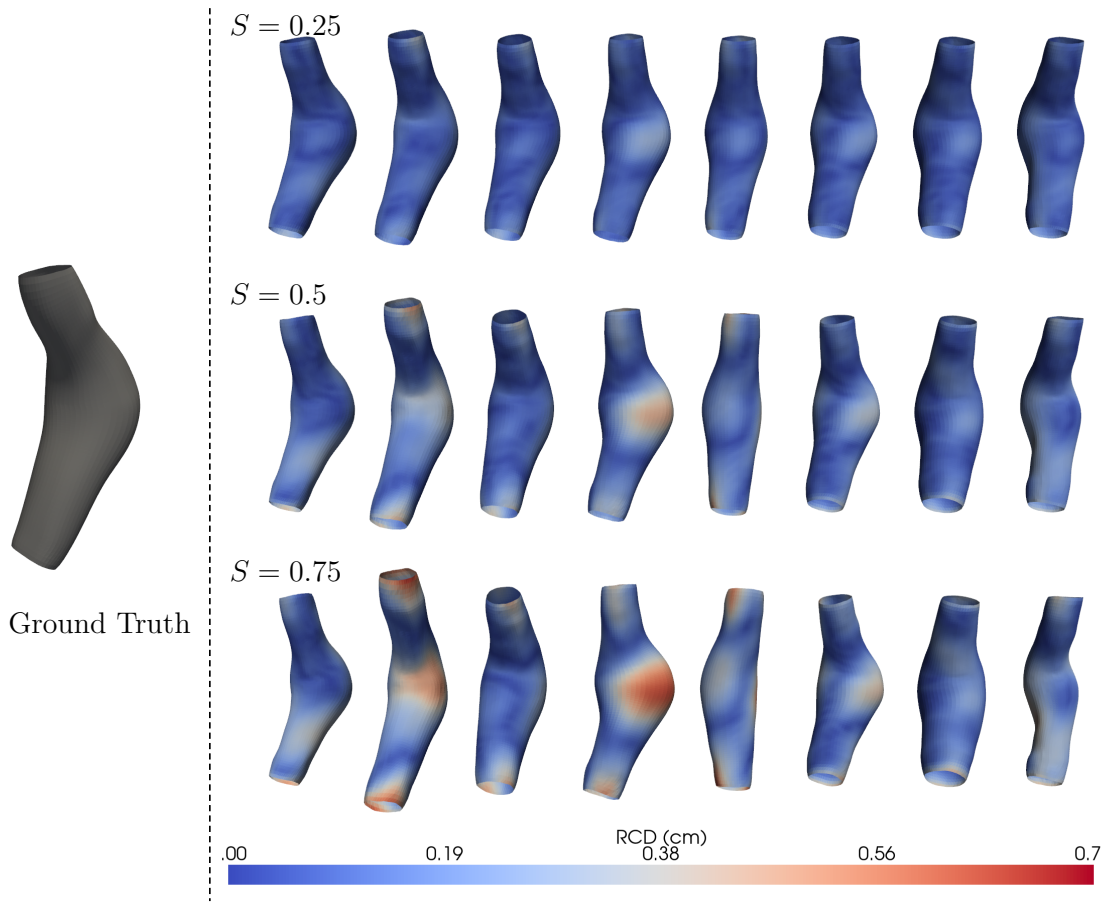

An AAA mesh $M$ is first embedded into a latent space by an encoder that maps it to a mean latent vector $\mu \in \mathbb{R}^d$. To explore shape variability while preserving anatomical realism, new mesh instances $\tilde{M}$ are generated by extrapolating from the latent mean using a controlled Gaussian perturbation:
\[
\mathbf{z} = \mu + S \cdot\Sigma, \quad \text{where} \quad \Sigma \sim \mathcal{N}(0,I),
\]
where $S \in \mathbb{R}^+$ modulates the noise amplitude. Larger values of $S$ produce more pronounced deviations from the ground truth. The resulting extrapolated meshes $\tilde{M}$ simulate plausible anatomical variations while maintaining topological consistency with the original AAA.

As visualized in \Cref{fig:extr}, the leftmost column shows the ground-truth mesh $M$, and the subsequent eight columns display extrapolated meshes $\tilde{M}_i$ with varying shapes. Each row corresponds to a different noise magnitude ($S = 0.25$, $S = 0.5$, and $S = 0.75$). The RCD is computed between each generated mesh and the ground truth and is shown as a heatmap over the mesh surfaces. As $S$ increases, the extrapolated meshes exhibit greater geometric deviations. Notably, the fourth and sixth columns display a marked increase in aneurysm sac size, indicative of radial expansion, while the second column shows elongation along the vessel axis. These shape variations are consistent with clinically relevant deformations, underscoring the method's capacity to generate realistic anatomical diversity through controlled latent space perturbations.

\subsubsection{Latent Interpolation}
\label{sec:int}
\begin{figure}[!htb]
    \centering
    \begin{adjustbox}{width=0.85\textwidth}
        \begin{NiceTabular}{cc}
            \Block[borders={right,tikz={densely dashed, thick}}]{1-1}{}
            \includegraphics[trim={5cm 1cm 8cm 3cm}, clip, width=0.2\textwidth]{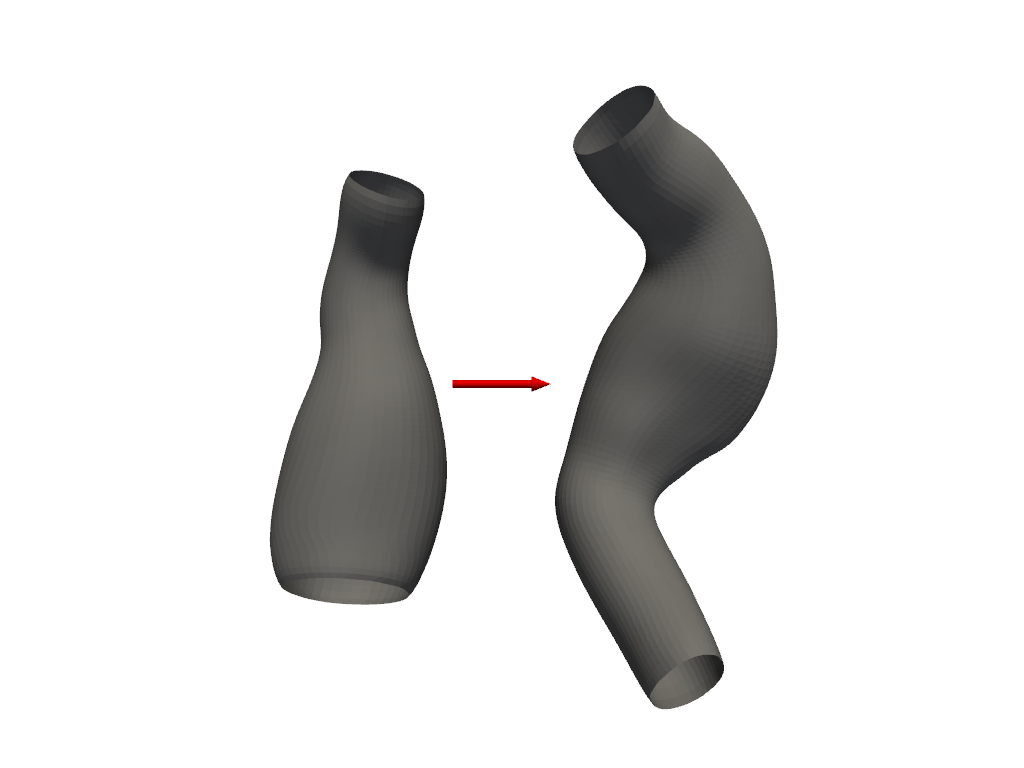} &
            \includegraphics[trim={3cm 11cm 3cm 12cm}, clip, width=0.8\textwidth]{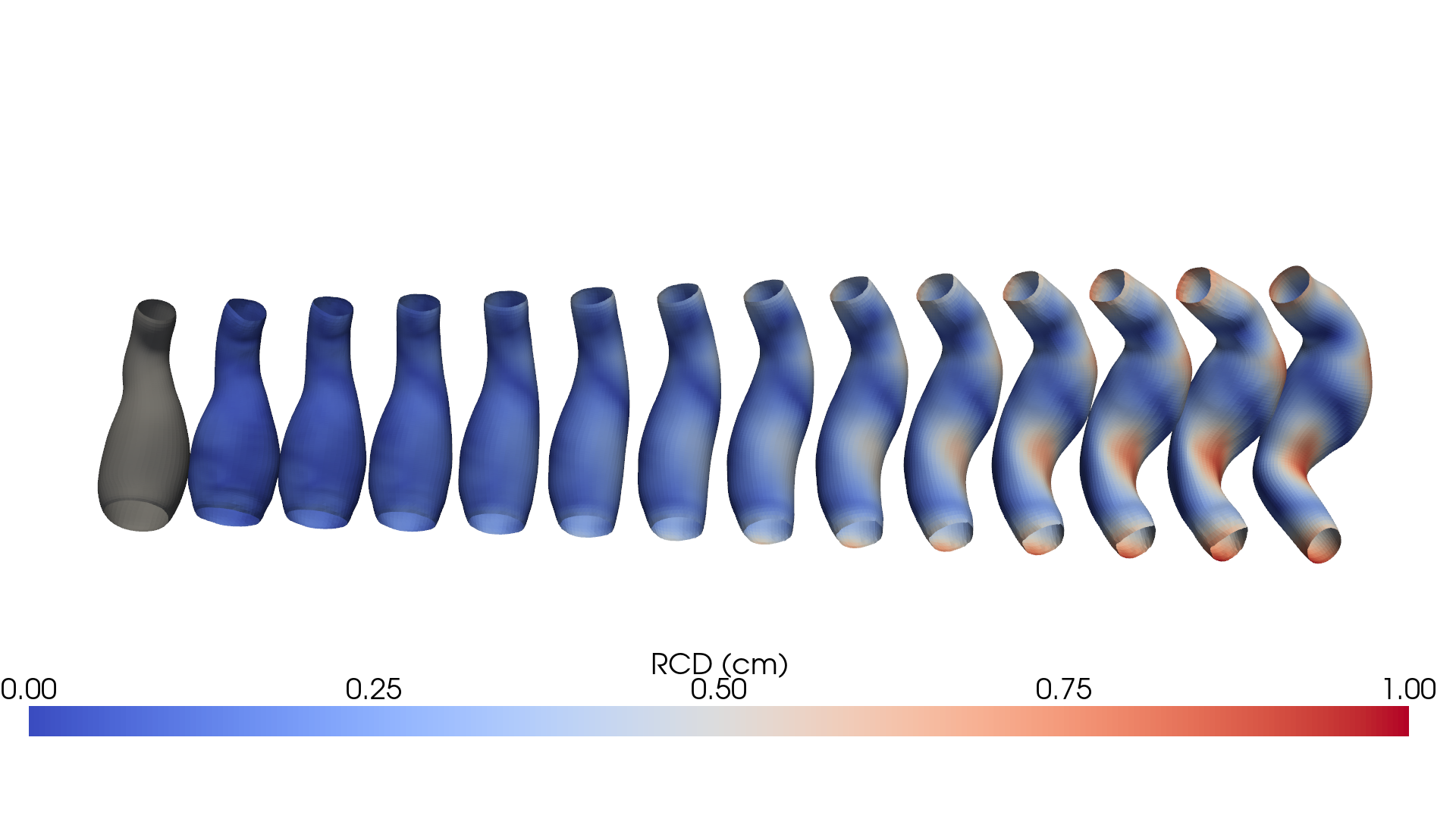} \\
            \Block[borders={right,tikz={densely dashed, thick}}]{1-1}{}
            \includegraphics[trim={5cm 1cm 8cm 3cm}, clip, width=0.2\textwidth]{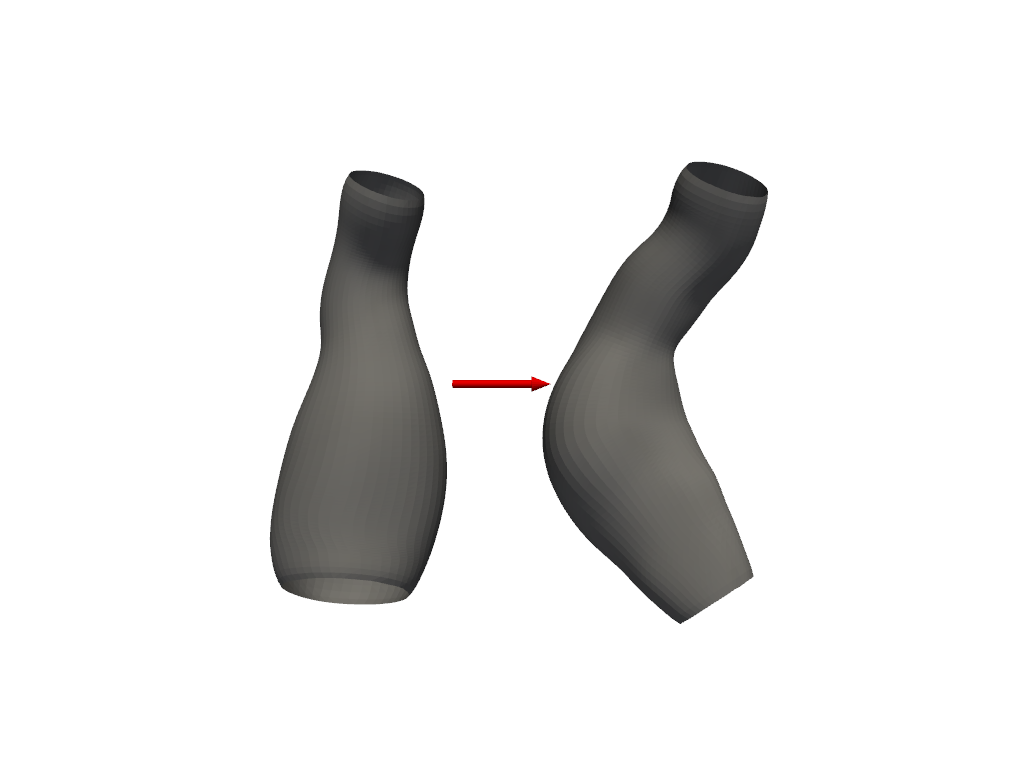} &
            \raisebox{0.5cm}{\includegraphics[trim={3cm 12cm 3cm 12cm}, clip, width=0.8\textwidth]{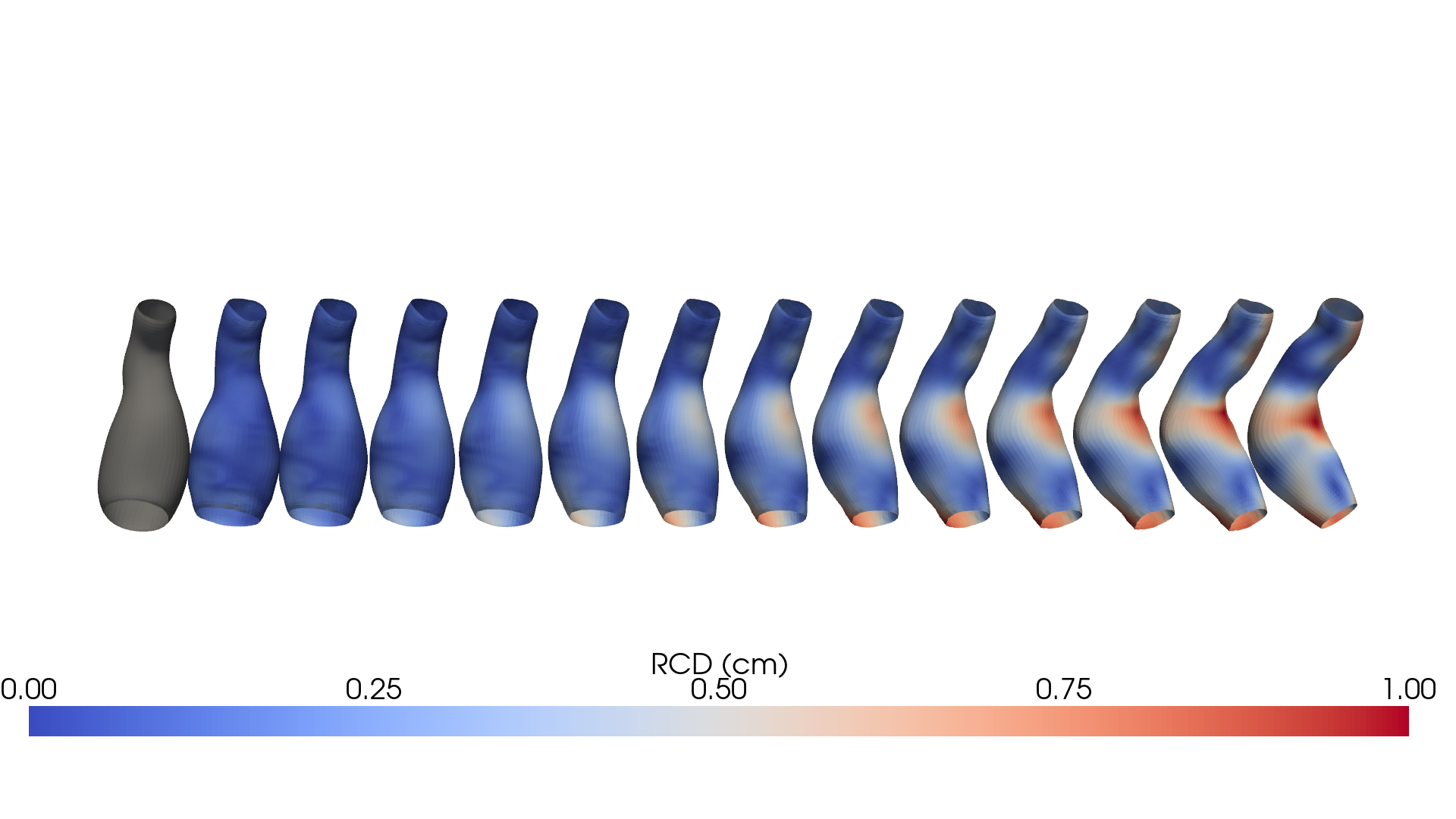}}\\
            \Block[borders={right,tikz={densely dashed, thick}}]{1-1}{}
            \raisebox{1.5cm}{\includegraphics[trim={5cm 1cm 8cm 3cm}, clip, width=0.2\textwidth]{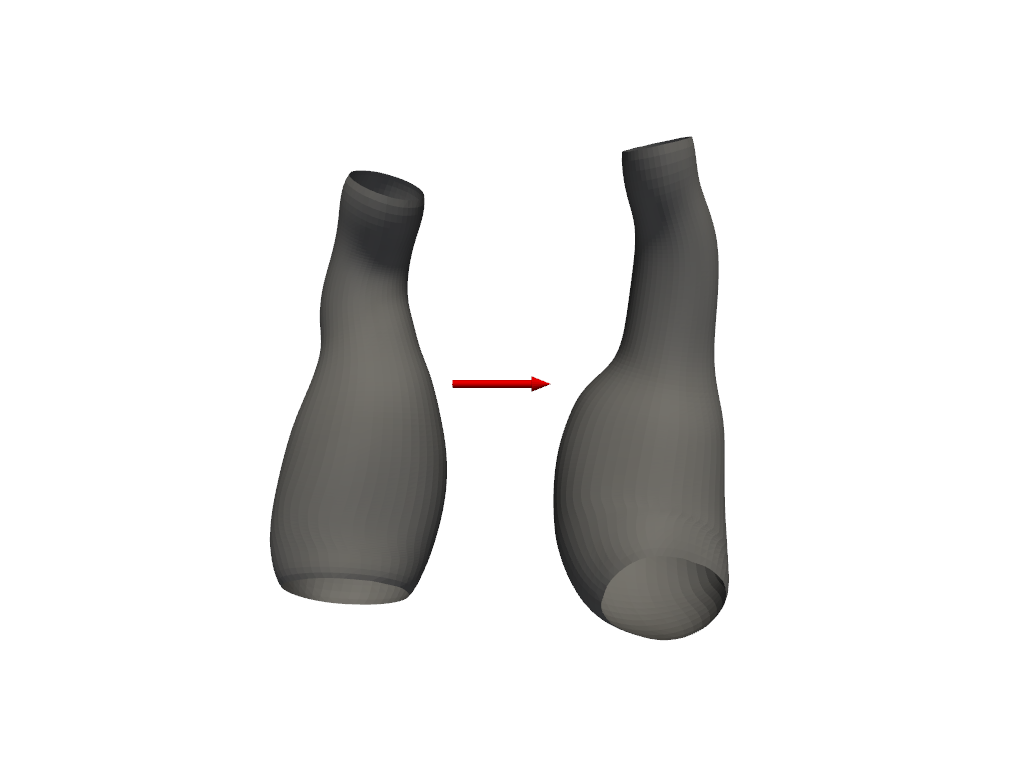}} &
            \includegraphics[trim={3cm 4cm 3cm 12.5cm}, clip, width=0.8\textwidth]{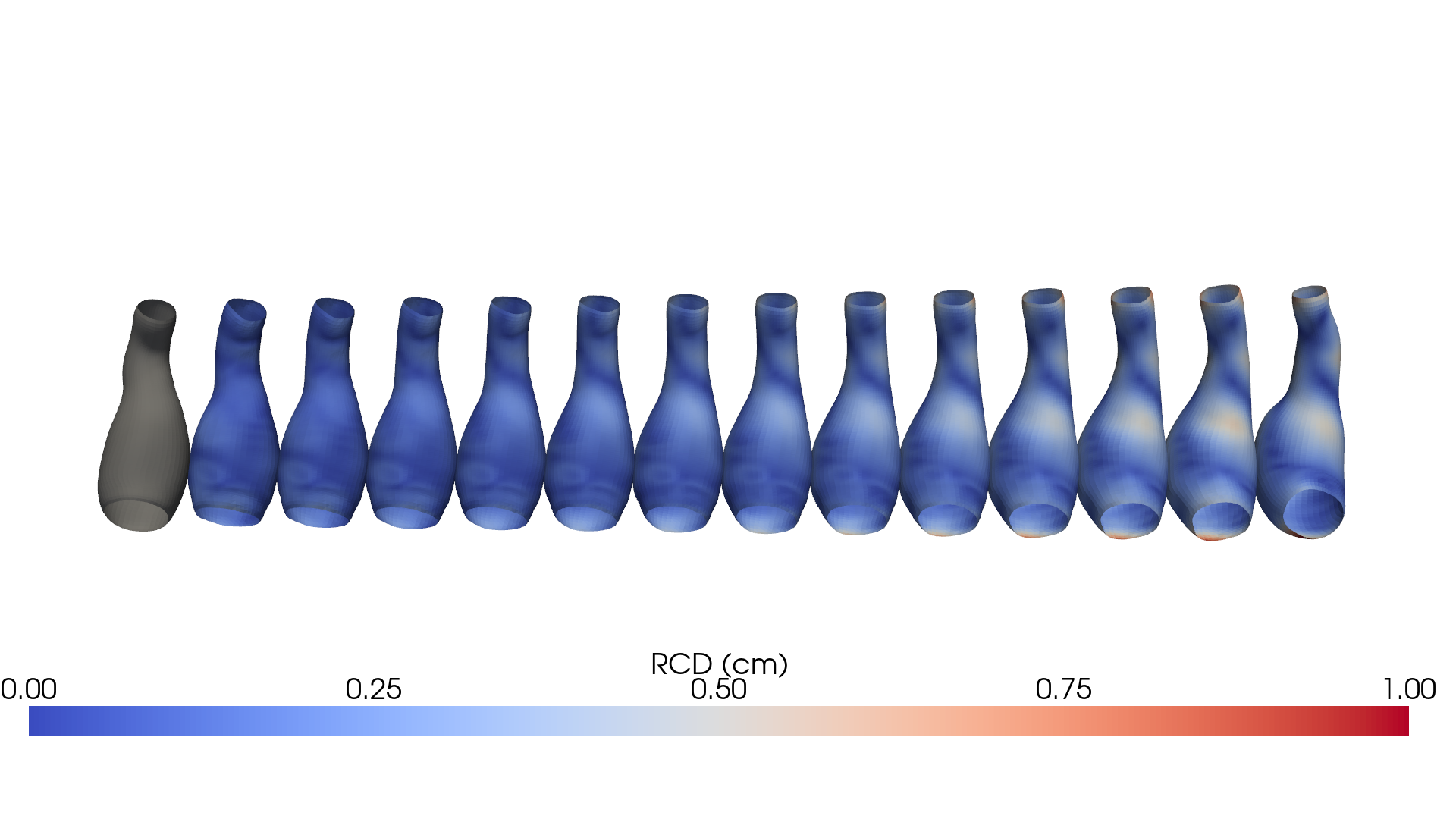}\\
        \end{NiceTabular}
    \end{adjustbox}
    \caption{Interpolation of latent codes. The left column shows the ground truth aneurysm meshes, while the right column presents the intermediate shapes generated by interpolating between latent representations. The color map encodes the root Chamfer distance (RCD), illustrating how the geometry transitions smoothly between the original and interpolated states.}
\label{fig:interpolation}
\end{figure}

In image data, interpolation is straightforward due to the inherent grid structure: pixels are spatially ordered and aligned, allowing for direct blending of intensity values. However 3D point clouds, such as those representing AAAs, lack this spatial ordering \cite{mao2019interpolatedconvolutionalnetworks3d}. Each point cloud is a set of points in $\mathbb{R}^3$ with no inherent order, making direct interpolation between two clouds ill-posed due to the absence of pointwise correspondence.

To overcome this issue, a learned latent space is exploited, in which shape representations are continuous and semantically structured. Given two AAA meshes \(M_1\) and \(M_2\), these meshes are first encoded into corresponding latent vectors \(\mu_1\) and \(\mu_2\) using the encoder. A sequence of intermediate latent codes is then defined by linear interpolation between these two vectors:
\[
\mu(i) = \mu_1 + \frac{i}{n}(\mu_2 - \mu_1), \quad \text{for} \quad i = 0, 1, \dots, n,
\]
where $n$ is the number of interpolation steps. Each interpolated latent vector $\mu(i)$ is then passed through the decoder to generate a mesh $\tilde{M}_i$:

This process yields a smooth sequence of mesh shapes transitioning from $M$ to each respective target geometry, as shown in \Cref{fig:interpolation}, where the leftmost column displays the ground-truth aneurysm geometry $M$, which is shared across all three interpolation rows as the common source shape, as well as the three target geometries. Each row corresponds to a different target shape, depicted in the rightmost column. The intermediate columns visualize decoded shapes from the interpolated latent vectors. The RCD is used to visualize deviations from the initial shape $M$ through color-coded maps over each interpolated mesh. These visualizations confirm that interpolation in latent space produces anatomically plausible, continuous shape transformations.

\section{Conclusion}

In this work, a GCN-$\beta$-VAE framework capturing complex geometric variations in abdominal aortic aneurysm (AAA) morphology through a compact and disentangled latent representation is presentend. Even when trained on a limited dataset, the model consistently outperforms classical techniques such as PCA on both the $L_2$ and Chamfer metrics, demonstrating robustness and generalization capabilities. By constraining the latent space via the $\beta$-VAE structure, clinically relevant morphological features are retained while redundancy among latent dimensions is minimized. This structured encoding enables smooth and interpretable shape synthesis via latent interpolation and extrapolation. The resulting synthetic AAAs preserve critical anatomical characteristics and exhibit clinically meaningful variations including sac diameter, neck angulation, and overall length. A physically plausible augmentation strategy (Procrustes RandAugment) further enriches the training data without compromising anatomical validity, enhancing performance on unseen cases, reducing reconstruction and correspondence errors by up to one fifth on a dataset of only 60 patient cases. Taken together, these elements transform a data-sparse clinical scenario into a source of virtual cohorts, benefiting device testing, haemodynamic simulation and in-silico trials.

Despite these advances, several limitations remain. The latent disentanglement achieved is partial, and some latent factors may still entangle correlated anatomical changes. Additionally, the proposed method is designed for watertight meshes. Therefore, in cases where the AAA surface mesh is open, a preprocessing step is required to close the mesh. This is addressed by creating geometric lids through radial basis function interpolation, allowing the resulting surfaces to be treated as watertight during training.

Future directions include extension to variable-topology meshes or point-cloud representations to capture broader anatomical diversity, semi-supervised learning using known key geometrical features (e.g., the centerline of the blood vessel), and large-scale validation on external datasets. Moreover, incorporating supervised clinical labels could guide the latent space toward clinically actionable features. Finally, real-time generative tools may be developed for patient-specific surgical planning and in silico trials.

\bibliographystyle{unsrt}
\bibliography{sample}

\end{document}